\documentclass[mathfont=newtx,accepted]{uai2026} 

\usepackage[british]{babel}

\usepackage[round]{natbib} 
\usepackage{mathtools, amssymb, amsmath} 
\usepackage{xurl}
\usepackage{xspace}
\usepackage{textcomp}
\usepackage{subcaption}

\usepackage{booktabs} 
\usepackage{tikz} 

\usepackage{xcolor}

\definecolor{teal}{rgb}{0.0, 0.5, 0.5}
\definecolor{amethyst}{rgb}{0.6, 0.4, 0.8}
\definecolor{thulianpink}{rgb}{0.87, 0.44, 0.63}
\definecolor{tiffanyblue}{rgb}{0.04, 0.73, 0.71}
\definecolor{pear}{rgb}{0.82, 0.89, 0.19}
\definecolor{applegreen}{rgb}{0.55, 0.71, 0.0}
\definecolor{burgundy}{rgb}{0.5, 0.0, 0.13}
\definecolor{persianindigo}{rgb}{0.2, 0.07, 0.48}
\definecolor{mydarkblue}{rgb}{0,0.08,0.45}

\definecolor{sc1}{HTML}{0B3D91}   
\definecolor{sc2}{HTML}{1B6EC2}   
\definecolor{sc3}{HTML}{2E8BC0}   
\definecolor{sc4}{HTML}{56A8D6}   
\definecolor{sc5}{HTML}{7EC8E3}   

\definecolor{sae1}{HTML}{8B2500}  
\definecolor{sae2}{HTML}{C44E20}  
\definecolor{sae3}{HTML}{E27729}  
\definecolor{sae4}{HTML}{F0A04B}  
\definecolor{fd1}{HTML}{4B0082}   
\definecolor{fd2}{HTML}{6B24A8}   
\definecolor{fd3}{HTML}{8B4FC7}   
\definecolor{fd4}{HTML}{A87DD8}   
\definecolor{rf1}{HTML}{1A6B1A}   
\definecolor{rf2}{HTML}{2E8B2E}   
\definecolor{rf3}{HTML}{50B050}   
\definecolor{rf4}{HTML}{72D272}   
\definecolor{basegry}{HTML}{555555}

\newcommand{\ind}{\mathbf{1}} 

\usepackage{amsmath,amsfonts,bm}

\def\eqref#1{equation~\ref{#1}}

\def\1{\bm{1}}

\def\rva{{\mathbf{a}}}
\def\rvb{{\mathbf{b}}}

\def\rvd{{\mathbf{d}}}

\def\rvh{{\mathbf{h}}}

\def\rvq{{\mathbf{q}}}
\def\rvr{{\mathbf{r}}}

\def\rvx{{\mathbf{x}}}
\def\rvy{{\mathbf{y}}}
\def\rvz{{\mathbf{z}}}

\def\rmA{{\mathbf{A}}}
\def\rmB{{\mathbf{B}}}

\def\rmD{{\mathbf{D}}}

\def\rmI{{\mathbf{I}}}

\def\rmM{{\mathbf{M}}}

\def\rmW{{\mathbf{W}}}

\def\vz{{\bm{z}}}

\DeclareMathAlphabet{\mathsfit}{\encodingdefault}{\sfdefault}{m}{sl}
\SetMathAlphabet{\mathsfit}{bold}{\encodingdefault}{\sfdefault}{bx}{n}

\def\sR{{\mathbb{R}}}

\DeclareMathOperator*{\argmax}{arg\,max}
\DeclareMathOperator*{\argmin}{arg\,min}

\usepackage{graphicx}
\usepackage{booktabs}
\usepackage{comment}
\usepackage{caption}  
\usepackage{subcaption}
\usepackage{tabularx}
\usepackage{xcolor} 
\usepackage{tablefootnote}
\usepackage{pifont}

\usepackage[capitalize,noabbrev]{cleveref} 

\newcommand{\parless}[1]{\noindent\textbf{#1}}

\hypersetup{
  colorlinks=true,
  linkcolor=thulianpink,
  filecolor=pear,
  urlcolor=tiffanyblue,
  citecolor=amethyst,
}

\usepackage{tcolorbox}
\tcbuselibrary{breakable,skins}
\definecolor{lavender}{HTML}{E6E6FA}
\definecolor{watergreen}{HTML}{AFEEEE}
\definecolor{lightpastelpurple}{rgb}{0.69, 0.61, 0.85}
\newtcolorbox{summarybox}{colback=lavender,colframe=lightpastelpurple,boxrule=0.4pt}

\title{
    Stop Probing, Start Coding: Why Linear Probes and Sparse Autoencoders Fail at Compositional Generalisation
}

\author[1]{Vit\'oria Barin-Pacela\textsuperscript{*}}
\author[1]{Shruti Joshi\textsuperscript{*}}
\author[2]{Isabela Camacho}
\author[1,4]{Simon Lacoste-Julien}
\author[3]{David Klindt}
 \affil[ ]{%
      \textsuperscript{*}Equal contribution; authors listed in alphabetical order.
  }
\affil[1]{%
     Mila - Qu\'ebec AI Institute \& DIRO, Universit\'e de Montr\'eal
}
\affil[2]{%
    Santa Clara University
}
\affil[3]{%
    Cold Spring Harbor Laboratory
  }
\affil[4]{%
    Canada CIFAR AI Chair
  }

\begin{document}

\maketitle

\begin{abstract}
The linear representation hypothesis states that neural network activations encode high-level concepts as linear mixtures. However, under superposition, this encoding is a projection from a higher-dimensional concept space into a lower-dimensional activation space, and a linear decision boundary in the concept space need not remain linear after projection. In this setting, classical sparse coding methods with per-sample iterative inference leverage compressed sensing guarantees to recover latent factors. Sparse autoencoders (SAEs), on the other hand, amortise sparse inference into a fixed encoder, introducing a systematic gap. We show this amortisation gap persists across training set sizes, latent dimensions, and sparsity levels, causing SAEs to fail under out-of-distribution (OOD) compositional shifts. Through controlled experiments that decompose the failure, we identify \emph{dictionary learning}---not the inference procedure---as the limiting factor: SAE-learned dictionaries point in substantially wrong directions, and replacing the encoder with per-sample FISTA on the same dictionary does not close the gap. An oracle baseline proves the problem is solvable with a good dictionary at all scales tested. Our results, including experiments with real LLM activations (Pythia-70M, Gemma-2-2B) reframe the SAE failure as a dictionary learning challenge, not an inference problem, and point to scalable dictionary learning as the key open problem for sparse inference under superposition.
\end{abstract}

\section{Introduction}
\label{sec:intro}
Understanding the internal representations of Large Language Models (LLMs) is crucial for their safe and reliable deployment.
The \textit{linear representation hypothesis} (LRH) is a foundational assumption in mechanistic interpretability, stating that a model's activations are linear mixtures of underlying concepts \citep{smolensky_tensor_1990,mikolov2013,arora-etal-2016-latent,jiang2024originslinearrepresentationslarge, park2024linear, smith2024strongfeature}.
It has motivated crucial progress on methods such as linear probing for concept discovery and activation steering \citep{turner2023steering, chalnev2024improving}.

To make this precise, let $\rvz \in \sR^{d_z}$ denote ground-truth latent variables (concepts), and let $\rvy \in \sR^{d_y}$ denote a model's activations.
The LRH asserts that the relationship between concepts and activations is linear: $\rvy \approx \rmW \rvz$ for some matrix $\rmW \in \sR^{d_y \times d_z}$.
The goal of interpretability is to recover $\rvz$ from $\rvy$---i.e., to infer which concepts are active in an activation vector.
When $d_z > d_y$, more concepts are encoded than there are activation dimensions, a regime known as \emph{superposition} \citep{elhage_toy_2022}. The system $\rvy = \rmW \rvz$ is underdetermined:
each observation $\rvy$ is consistent with infinitely many $\rvz$, so
one cannot simply learn a linear unmixing to map activations back to concepts. In other words, linear
\emph{representation}---concepts being linearly encoded in
activations---is not the same as linear \emph{accessibility}---that concepts are recoverable by a linear transformation. However this distinction is routinely overlooked and the LRH is conflated with the much stronger latter claim.

Concept recovery requires additional structure to resolve the
underdetermination, such as by assuming sparsity, i.e., in practice, only $k \ll d_z$
concepts are active in any given input. Compressed sensing is precisely
the framework that characterises when sparse signals can be recovered
from such underdetermined measurements \citep{donoho2003optimally, donoho2006compressed}. Under the sparsity assumption,
classical
results show that $k$-sparse codes can be recovered from
$d_y = \mathcal{O}(k \log(d_z / k))$ suitably random measurements via nonlinear
algorithms (e.g., basis pursuit, iterative thresholding).
By contrast, requiring recovery to be \emph{linear} (e.g., a linear probe) and allowing for an error \(\epsilon>0\), increases the required number of dimensions to
$d_y = \Omega_{\epsilon}\!\left(\frac{k^2}{\log k}
\log\!\left(\frac{d_z}{k}\right)\right)$ \citep{garg2026featureslanguagemodelstore}, with a quadratic complexity in $k$.  Concretely, for $d_z = 10^6$ latent concepts with $k = 100$ active at once, nonlinear recovery requires only $d_y \approx 920$ dimensions while linear recovery requires $d_y \approx 20{,}000$---a typical transformer hidden size of $4096$ comfortably exceeds the former but falls far short of the latter. Thus, whether concepts are recoverable from a fixed set of activations therefore depends not only on whether they are linearly encoded, but on \emph{how} one attempts to decode them. 

\begin{figure*}
\begin{subfigure}[t]{0.39\textwidth}
    \centering
    \includegraphics[width=\linewidth]{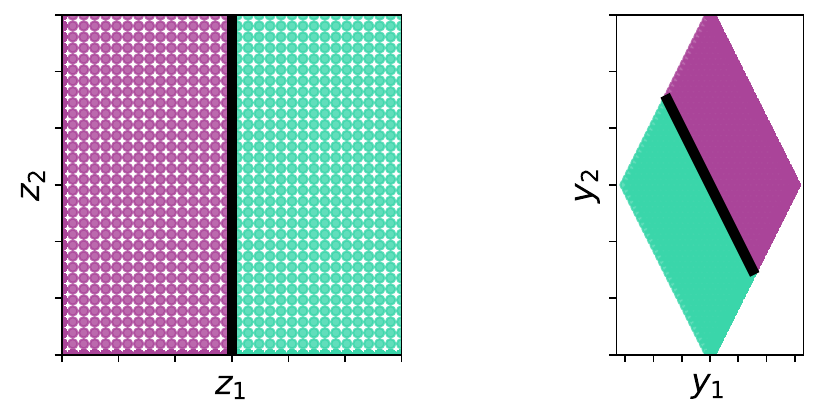}
    \caption{}
    \label{fig:boundary_2d}
\end{subfigure}
\hfill
\begin{subfigure}[t]{0.55\textwidth}
    \centering
    \includegraphics[width=\linewidth]{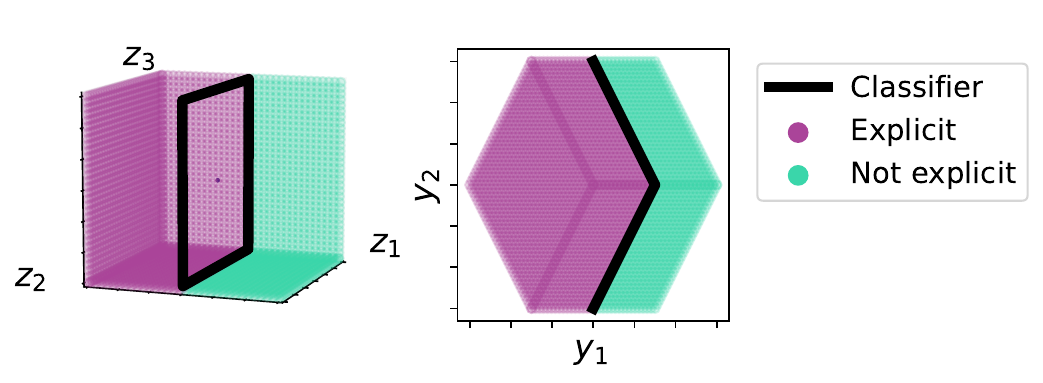}
    \caption{}
    \label{fig:boundary_3d}
\end{subfigure}
\caption[bvhdvb]{
Binary classification with $t = \ind{z_1 > 0.5}$ (green: not explicit, purple:
explicit).
\textbf{(a)}~When $d_z = d_y$, the linear decision boundary in
latent space remains linear after mixing $\rvy = \rmW\rvz$.
\textbf{(b)}~When $d_z > d_y$ (overcompleteness) and \(z\) sparse, we can project down into non-overlapping regions (i.e., compressed sensing is possible), but the decision
boundary becomes nonlinear in activation space, making linear probes
insufficient.
}
\label{fig:overview_boundary}
\end{figure*}

Additionally, with respect to downstream implications, the compression through $\rmW$ necessarily distorts the geometry of the latent space, so that a decision boundary that is linear in $\rvz$-space can become nonlinear in $\rvy$-space (\cref{fig:overview_boundary}). Recovering $\rvz$ from $\rvy$ therefore requires nonlinear methods, but which nonlinear method matters. \textit{Amortised} inference
\citep{gregor2010learning}, as implemented by Sparse Autoencoders
(SAEs) \citep{ng2011sparse,
cunningham2023sparseautoencodershighlyinterpretable}, learns a fixed
encoder $r \colon \sR^{d_y} \to \sR^{d_h}$ that maps an activation to
a sparse code in a single forward pass by training
on a finite distribution of training data. \emph{Per-sample inference}, by
contrast, solves an optimisation problem from scratch for each input,
using only the observation $\rvy$ and the dictionary $\rmW$, with no
dependence on a training distribution. 

\parless{Amortisation gap.} The solution learned via amortised or per-sample inference can be evaluated by how accurately either recovers the true sparse code, e.g.,
whether the correct concepts are identified as active. In-distribution, both approaches
may recover $\rvz$ well, and the discrepancy between them---the
\emph{amortisation gap} \citep{margossian2023amortized, kim2021reducing,
zhang2022generalization, cremer2018inference, schott2021visual,
paiton2020selectivity, oneil2024compute}---can be small. Under
distribution shift, however, the gap widens: when the sparsity pattern
changes (e.g., novel combinations of concepts co-activate), the
amortised encoder's recovery degrades because it was optimised for
training-time statistics, while per-sample methods that optimise
from scratch for each input remain unaffected. This paper studies the amortisation gap in the context of
interpretability under superposition:
\begin{quote}
\emph{Under what conditions does sparse inference recover the true
latent factors from superposed activations, and how does the choice of
inference procedure---amortised, per-sample, or hybrid---affect
robustness under distribution shift?}
\end{quote}

This argument adds nuance to recent studies suggesting that linear probes trained on top of LLM activations are superior to SAEs in out-of-distribution (OOD) binary classification tasks \citep{kantamneni2025sparse}. We argue that the recent OOD failures of SAEs are not an indictment of the superposition hypothesis, but rather a predictable consequence of replacing principled sparse inference with a brittle, amortised encoder. Instead of discarding the powerful framework of sparse coding, we embrace the geometric consequences of superposition and utilize methods equipped to handle the nonlinearity it induces.

\parless{Main contributions.}  In this work, we revisit classical sparse coding to address the central question.
We show that under superposition, even labels that are linearly separable in latent space may become nonlinearly separable in activation space. We demonstrate that this is particularly pronounced in OOD settings, so that a perfect linear probe trained in-distribution will fail OOD (\cref{fig:overview_boundary}).
SAEs perform nonlinear inference, but amortising it into a fixed encoder introduces a systematic amortisation gap (\cref{fig:phase}): the encoder fits the training distribution's co-occurrence structure and fails to generalise to novel combinations of latent factors under OOD composition shift (\cref{fig:toy3d}).
An oracle baseline---per-sample FISTA with the ground-truth dictionary---achieves near-perfect OOD recovery at all scales tested, proving the problem is solvable under compressed sensing theory (\cref{sec:experiments}).  Through controlled experiments that decompose the SAE failure, we identify dictionary learning---not the inference procedure---as the limiting factor. SAE-learned dictionaries point in substantially wrong directions, and replacing the encoder with per-sample FISTA on the same dictionary does not close the gap (\cref{fig:frozen-faceted}). Classical dictionary learning (DL-FISTA) produces better dictionaries at small scale, but both methods fail at dictionary learning when the latent dimension grows large.  These results reframe the SAE failure: the bottleneck is not amortisation of inference but amortisation of dictionary learning, and the path forward requires scalable algorithms for learning dictionaries under the compressed-sensing framework. Finally, we validate feasibility on real LLM activations: on SAEBench sparse probing \citep{karvonen2025saebench,kantamneni2025sparse}, per-sample inference with a learned dictionary is the only unsupervised method to exceed raw-activation probing on Pythia-70M and
outperforms Top-K SAEs on all eight Gemma-2-2B datasets.

\section{Related Work}
Compositional generalisation---the ability to understand and produce novel combinations of learned concepts---remains a fundamental challenge for neural networks \citep{fodor1988connectionism, hupkes2020compositionality}. 
Current approaches in causal representation learning attempt to achieve this through structural constraints, such as \textit{additive} decoders \citep{lachapelle2023additive} (e.g., in SAEs), or specific training objectives like compositional risk minimisation \citep{mahajan2025compositionalriskminimization}.
These works focus on obtaining guarantees for the compositional generalisation of disentangled models, while here, we evaluate the effect of compositional shifts under superposition.

SAEs have recently emerged as a primary tool for decomposing the internal activations of LLMs into, `monosemantic', i.e., interpretable features \citep{cunningham2023sparseautoencodershighlyinterpretable}. Despite their success in interpretability, their out-of-distribution (OOD) robustness is a growing concern. 
Recent evaluations suggest that SAEs trained on general datasets often fail to discover generalisable concepts across different domains or layers \citep{heindrich2025sparseautoencodersgeneralizecase}, underperform compared to simple linear probes \citep{kantamneni2025sparse}, and remain brittle even when scaled \citep{gao2024scalingevaluatingsparseautoencoders}. Interestingly, this brittleness is less pronounced in domain-specific applications, such as medical QA or pathology, where SAEs have shown more stable and biologically relevant feature transfer \citep{oneill2025resurrectingsalmonrethinkingmechanistic, le2024learningbiologicallyrelevantfeatures}. These conflicting results motivate a more principled evaluation of SAEs under distribution shifts \citep{joshi2025identifiablesteeringsparseautoencoding}.

Recent work has highlighted the limitations of SAEs in recovering true latent variables \citep{oneil2024compute, paulo2025sparse}. 
This stands in contrast to the classical sparse coding framework \citep{olshausen1996emergence, ranzato2007sparse}, which utilises iterative optimisation rather than a learned encoder to recover latent variables. This iterative approach provides stronger theoretical guarantees for the unique recovery of latents \citep{hillar2015, lewicki2000learning, gribonval2015sparse}.
In contrast to \citet{oneil2024compute}, who explore different in-distribution (ID) amortisation strategies, we focus our analysis on downstream tasks under OOD compositional shifts. 

Lastly, literature on overcomplete independent component analysis \citep{podosinnikova19a, wang2024identifiabilityovercomplete} explores identifiability where the number of latent variables exceeds the number of observed variables ($d_z > d_y$), though these models typically rely on statistical independence rather than sparsity.

\section{Amortisation vs Pointwise Sparse Inference and Compositional Generalisation}
\label{sec:sc-vs-sae}
We begin by specifying the data-generating process. Consider latent variable vectors $\rvz \in \sR^{n}$ with at most $k$ non-zero entries. A support set $S \subseteq [n]$ is drawn first to index these non-zero components following the process:
\begin{flalign}
    S \sim p_S, \quad \vz \sim p(\vz\mid S)   \nonumber \\
    \mathrm{supp}(p_S) \subseteq \mathcal{S}_k := \{S\subseteq[n]: |S|\le k\}, \nonumber \\
    \mathbb{P}\!\left(\rvz_{S^c}=\mathbf 0 \,\middle|\, S\right) = 1.
    \label{eqn:dgp_z}
\end{flalign}
An unknown generative process $g$ maps latents to data, producing $\rvx := g(\rvz) \in \sR^{d_x}$. Although $\rvz$ is unobserved, we have access to learned representations $\rvy := f(\rvx) \in \sR^{d_y}$ for some fixed (encoding) function $f : \sR^{d_x} \to \sR^{d_y}$ (e.g., an LLM activation map). Since the coordinates of $\rvy$ need not necessarily align with those of $\rvz$, we fit a representation model $r : \sR^{d_y} \to \sR^{d_h}$ estimating $\rvh := r(\rvy)$ and another decoder $q : \sR^{d_h} \to \sR^{d_y}$ s.t. $\hat{\rvy} := q({\rvh})$. Ideally, $\rvh$ serves as a proxy for $\rvz$ and when $d_z$ is known, we can set $d_h=d_z$, while $d_y < d_z$. We refer to individual coordinates of $\rvh$ (one-dimensional subspaces of $\sR^{d_h}$) as \emph{features}, and we seek features that correspond (approximately) to the latent coordinates $\rvz$.

Typically, $\rvh$ is enforced to be sparse, in line with \cref{eqn:dgp_z}.  In practice, SAEs \citep{cunningham2023sparseautoencodershighlyinterpretable} are commonly used to learn the autoencoder $q \circ r$, where typically the decoder $q$ is assumed to be linear and the encoder $r$ is a single-linear layer followed by an activation function such as ReLU \citep{cunningham2023sparseautoencodershighlyinterpretable}, JumpReLU \citep{rajamanoharan2024jumpingaheadimprovingreconstruction}, or TopK \citep{gao2024scalingevaluatingsparseautoencoders, costa2025flat}. The assumption of a linear decoder is motivated by the LRH stating that the composed map $f \circ g$ is linear in the underlying latent variables s.t.,
\begin{align}
\rvy \;=\; f(g(\rvz)) \;\approx\; \rmA \rvz + \rvb, \qquad \rvx\sim p_{\rvx},
\label{eqn:lrh}
\end{align}
for some matrix $\rmA\in\sR^{d_y\times d_z}$ and offset $\rvb\in\sR^{d_y}$, where $p_{\rvx}$ denotes the induced data distribution under the generative process $\rvx=g(\rvz)$. A long line of work provides evidence for this hypothesis (c.f. \citet{rumelhart1973model, hinton1986learning, mikolov2013, ravfogel-etal-2020-null, klindt2025superposition}). More recently, theoretical work justifies why linear properties could arise in these models (c.f. \citet{arora-etal-2016-latent, jiang2024originslinearrepresentationslarge, roeder2021linear, marconato2024all, reizinger2024cross}). 

\parless{Ensuring injectivity under overcompleteness.} SAEs are often trained with an overcomplete feature dimension $d_h > d_y$, so that the decoder operates as a linear dictionary $\rmW\in\sR^{d_h\times d_z}$. When $d_h>d_y$, the decoder dictionary $\rmW \in \sR^{d_y \times d_h}$ has more columns than rows, i.e. it is projecting from a high to a low-dimensional space, like projecting a 3D object onto its 2D shadow. This compression means there must exist nonzero codes $\rvh$ for which $\rmW\rvh=0$ (i.e. $\ker(\rmW)\neq\{\mathbf 0\}$ since $\dim\ker(\rmW) = d_h-\mathrm{rank}(\rmW) \ge d_h-d_y > 0$). Thus, information is inevitably lost, so multiple codes $\rvh$ can produce the same activation $\rvy$, resulting in its infinitely many possible reconstructions. 

This non-uniqueness has a deeper consequence in that there is nothing to identify since identifying a ground truth solution would imply presupposing its uniqueness. But, the model $\rvy=\rmA \rvz$ is inherently not identifiably since, for any invertible (\(d_z\) by \(d_z\)) matrix $\rmM$, we can write $\rvy=(\rmA \rmM) (\rmM^{-1} \rvz) = \rmA' \rvz'$, defining a different dictionary and latent variable that result in the same observed variable $\rvy$. 
Thus, interpreting $\rvh=r(\rvy)$ as recovering latents requires an additional selection principle that prefers one solution among many consistent codes.

To obtain a unique solution, one typically restricts the code to be sparse, mirroring the latent sparsity assumption in \cref{eqn:dgp_z}. Concretely, we restrict $\rvh\in\sR^{d_h}$ to lie in a union of $k$-dimensional subspaces \footnote{equivalently requiring that $\mathrm{supp}(\rvh)\in \mathcal S_k^{(d_h)}:=\{S\subseteq[d_h]:|S|\le k\}$.}:
\[
\Sigma_k^{(d_h)} \;:=\; \{\rvh\in\sR^{d_h} : \|\rvh\|_0 \le k\},
\]
This sparsity can be enforced softly (e.g., via an $\ell_1$ penalty) or as a hard constraint through constrained optimisation \citep{ramirez2025position, gallegoPosada2025cooper}, or as implemented in TopK SAEs, which retain only the $k$ largest-magnitude coordinates of $\rvh$ per input. Algebraically, sparsity replaces the unconstrained feasibility set $\{\rvh\in\sR^{d_h}:\rmW\rvh=\rvy\}$ (typically infinite) with the constrained set $\{\rvh\in\Sigma_k^{(d_h)}:\rmW\rvh=\rvy\}$, which can be a singleton under suitable conditions on $\rmW$. In this sense, $r(\rvy)$ is meaningful only insofar as it implements a consistent sparse selection of dictionary atoms among the many codes that reconstruct the same $\rvy$.

\parless{Identifiability of sparse codes $\rvh$.} Restricting the codes to be sparse is not sufficient to guarantee uniqueness---there may still exist distinct $\rvh,\rvh'\in\Sigma_k^{(d_h)}$ with $\rmW\rvh=\rmW\rvh'$. We need a property of the dictionary $\rmW$ ensuring that it does not collapse sparse codes onto each other so that they can be identifiable. A standard sufficient condition is the restricted isometry property (RIP) \citep{candes2006, donoho_compressed_2006, candes2008introduction}: $\rmW$ satisfies RIP of order $s$ with constant $\delta_s$ if $(1-\delta_s)\|\rvh\|_2^2 \;\le\; \|\rmW\rvh\|_2^2 \;\le\; (1+\delta_s)\|\rvh\|_2^2$ \hspace{1mm} $ \forall \|\rvh\|_0\le s$.
Intuitively, $\rmW$ approximately preserves the geometry of the sparse codes $\rvh$, neither inflating nor collapsing them beyond a tolerance. When we consider $\rvh \in \Sigma_k^{d_h}$, we want to ensure that two distinct $k$-sparse codes $\rvh$ and $\rvh'$ are distinguishable, i.e., their difference is not invisible to $\rmW$. Since, their difference can have at most $2k$ non-zeros, this is equivalent to asking that $\rmW$ maps no nonzero $2k$-sparse vector to zero, since if $\rmW\rvh = \rmW\rvh'$, then $\rmW(\rvh - \rvh') = 0 $. RIP at order $s=2k$ ensures exactly this. \footnote{
A classical alternative is $\mathrm{spark}(\rmW)>2k$ (spark of a matrix is the smallest number of its columns that are linearly dependent), no linear combination of $2k$ or fewer columns can sum to zero, i.e., $\ker(\rmW)\cap\Sigma_{2k}^{(d_h)}=\{\mathbf 0\}$ and hence $\rmW$ is injective on $\Sigma_k^{(d_h)}$ \citep{donoho2003optimally, gribonval2004sparse}. But $\mathrm{spark}$ is typically intractable to compute.} 
So, under RIP, it is impossible for two different sparse vectors to map to the same output since their difference cannot be in the null space of $\rmW$, ensuring injectivity. With high probability, RIP is fulfilled for random Gaussian matrices projecting down into $d_y \geq \mathcal{O}(k \ln (\frac{d_h}{k}) )$ dimensions \citep{candes2008introduction}.  


It is important to distinguish two levels of identifiability that arise in this setting. The first is \emph{code-level identifiability}, i.e., given a fixed and known dictionary $\rmW$, when can we uniquely recover the sparse code $\rvh$ from an observation $\rvy$. This is precisely what RIP guarantees---it ensures that per-sample inference methods such as basis pursuit \citep{chen2001atomic}, or ISTA \citep{daubechies2004iterative}/FISTA \citep{beck2009fast} converge to the unique $k$-sparse solution consistent with $\rvy$. The second is \emph{dictionary identifiability}: can we  recover the dictionary $\rmW$ itself from observations $\rvy$? Classical results show that $\rmW$ is identifiable when the data is sufficiently sparse and diverse  \citep{hillar2015, gribonval2015sparse}, which is similar to a more flexible sufficient support variability condition on data for learning the dictionary \citep{joshi2025identifiablesteeringsparseautoencoding}. These guarantees, however, are only as useful as the optimisation procedure that realises them.

\parless{Sparse Coding and Amortisation.} Through point-wise inference, sparse coding infers a sparse code per input. Concretely, given samples $\{\rvy_i\}_{i=1}^p$ and fixed dictionary $\rmW$, the canonical \emph{pointwise} formulation solves, for each $i$,
\[
\rvh_i^\star \in \arg\min_{\rvh\in\sR^{d_h}} \tfrac12\|\rvy_i-\rmW\rvh\|_2^2 + \lambda\|\rvh\|_1,
\]
(or equivalently, a hard sparsity constraint $\rvh\in\Sigma_k^{(d_h)}$). Algorithms such as Iterative Shrinkage-Thresholding Algorithm (ISTA) and FISTA compute $\rvh_i^\star$ by iterating a sequence of code estimates $\rvh_i^{(0)},\rvh_i^{(1)},\ldots$ for each input, typically starting from $\rvh_i^{(0)}=\mathbf 0$ and applying repeated gradient-and-thresholding updates until convergence \citep{beck2009fast, daubechies2004iterative}.

When the dictionary is unknown, inference is paired with \emph{dictionary learning} \citep{olshausen1996emergence, mairal2010online}, to alternate between (i) estimating $\{\rvh_i\}_{i=1}^p$ given the current $\rmW$ and (ii) updating $\rmW$ to minimise reconstruction error under a column-norm constraint. A standard formulation is,
\begin{flalign}
\min_{\rmW,\{\rvh_i\}_{i=1}^p}\;\; \sum_{i=1}^p \Big(\tfrac12\|\rvy_i-\rmW\rvh_i\|_2^2 + \lambda\|\rvh_i\|_1\Big) \nonumber \\
\text{s.t.} \|\rmW_{:,j}\|_2 \le 1\;\;\forall j\in[d_h]. 
\end{flalign}


\begin{figure}
    \centering
    \includegraphics[width=\linewidth]{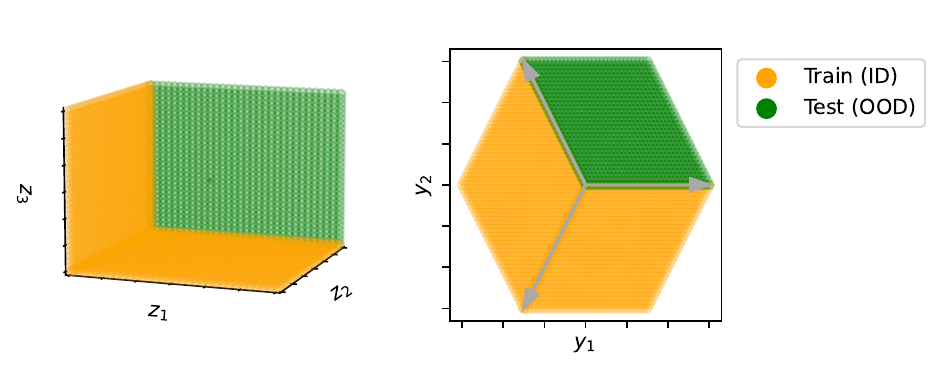}
    \caption{Compositional OOD split. \textit{Left:} In-distribution (ID) training data covers support pairs $(z_1,z_2)$ and $(z_2,z_3)$ and the novel combination $(z_1,z_3)$ is held out for OOD evaluation.
\textit{Right:} Same split in activation space $\rvy=\rmW\rvz$.}
    
    \label{fig:ood_split}
\end{figure}

In contrast, amortised methods replace per-input iterative solving with a feed-forward encoder that predicts $\rvh$ in a single forward pass, learning to approximate the solution across a training distribution \citep{vafaii2024poisson, vafaii2025brain}. We can have amortised sparse inference (e.g., LISTA 
\citep{gregor2010learning}) that unrolls ISTA iterations into learnable layers, so that its architecture is structurally tied to the sparse-coding objective and the dictionary is treated as given. SAEs, the dominant amortised approach in interpretability, are amortised autoencoding: they jointly learn both the encoder and the dictionary (the decoder $q$). An SAE encoder is free to learn any mapping that minimises reconstruction loss on the training data, including solutions that exploit distributional shortcuts rather than performing principled sparse decomposition.



In the fully unsupervised setting, the general impossibility of unsupervised identifiabilty implies typical SAEs have been shown not to be identifiable \citep{oneil2024compute}.
We hypothesise that this is one reason for the poor generalisation of SAEs OOD found in the literature \citep{kantamneni2025sparse}. In principle, a disentangled generalisation should allow for better OOD generalisation on downstream tasks \citep{Scholkopfetal21}---but whether this promise holds in practice depends on the entire pipeline---from dictionary quality, through the inference procedure, to the downstream task. In this paper, we focus on this downstream question. We operationalise the central question from \cref{sec:intro} by studying compositional distribution shifts (novel combinations of known concepts) which decompose it into two testable research questions:
\begin{tcolorbox}[
  enhanced jigsaw,
  breakable,
  colback=lavender!75,
  colframe=lavender!75, 
  boxrule=0pt,
  arc=0pt,
  left=1pt,right=2pt,top=2pt,bottom=2pt
]
    \begin{enumerate}[label=\roman*,wide, labelwidth=!, labelindent=0pt]
    \item  \textbf{Is sparse inference even necessary?} Linear probes outperforming SAEs on OOD tasks does not mean linear decoding is sufficient, it just means amortised inference is failing. Labels that are linearly separable in latent space become nonlinearly separable in activation space, and this nonlinearity is exposed precisely under compositional shifts. Rather than abandoning sparse coding for linear probes, we suggest the solution lies within the compressed-sensing framework---but the bottleneck may not be where one expects.
    \item \textbf{What is the bottleneck: inference or dictionary learning?} SAEs jointly learn a dictionary and an encoder. We decompose their failure to ask whether per-sample inference with an SAE-learned dictionary can close the gap, and if not, whether the dictionary itself---rather than the encoder---is the limiting factor
    \item\textbf{Does this transfer to real LLM activations?} The controlled experiments isolate the mechanism because the ground-truth dictionary is known. On real activations the concepts, their number, and their sparsity are unknown, so we test what remains observable: whether per-sample inference with a learned dictionary stays viable at scale, and whether sparse coding offers any benefits at scale.
\end{enumerate}
\end{tcolorbox}

\begin{figure}
    \centering
    \includegraphics[width=\linewidth]{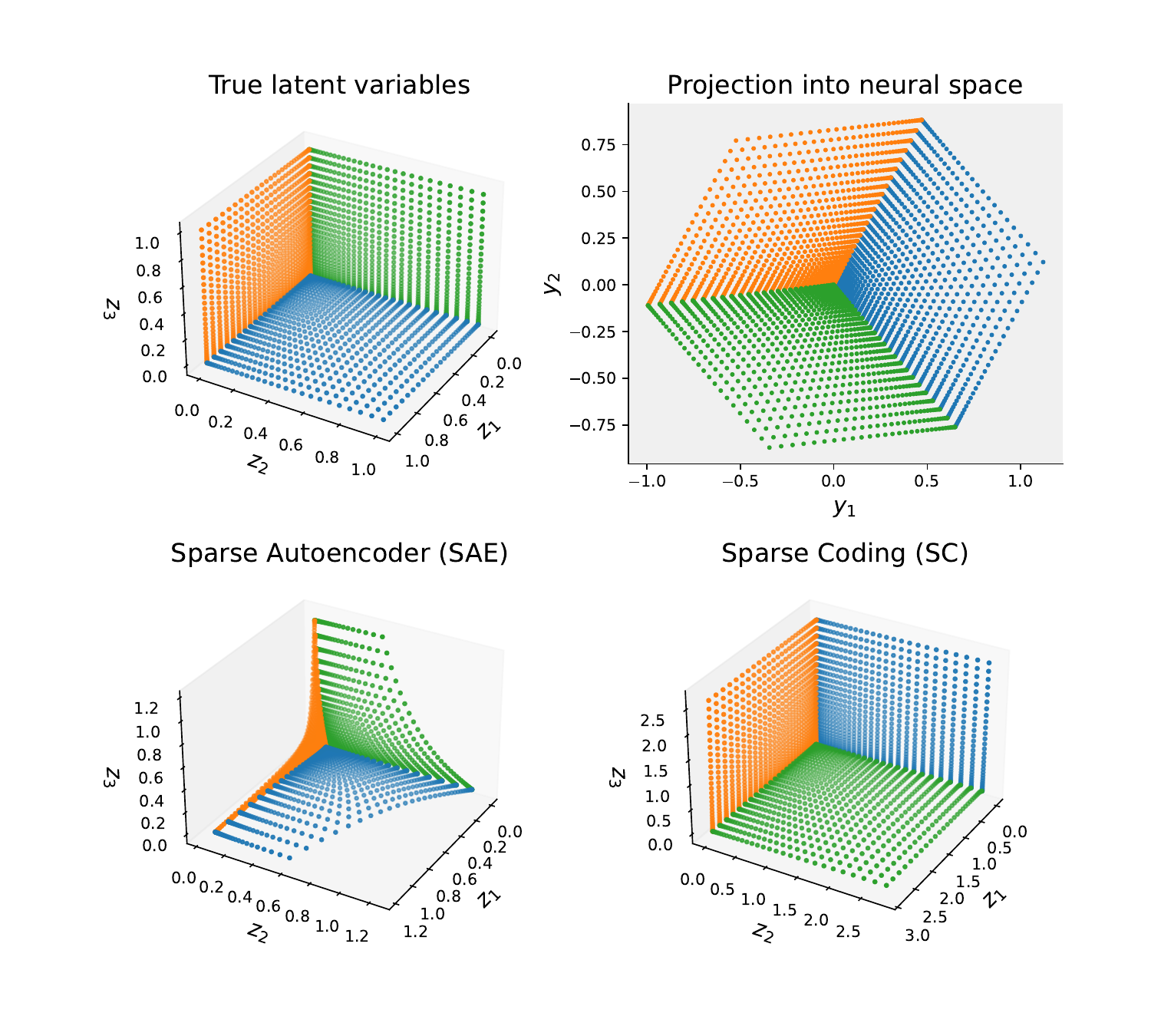}
    \caption{SAEs fail to recover latent variables under superposition, but sparse coding succeeds. \textbf{Top left:} Ground-truth latents ($d_z=3$, $k=2$); colors
denote active-variable combinations.
\textbf{Top right:} Activation space $\rvy = \rmW\rvz$ ($d_y=2$);
factors overlap after projection.
\textbf{Bottom left:} SAE reconstruction; planes are not recovered.
\textbf{Bottom right:} Sparse coding reconstruction; latents are
identified up to scaling.}
    \label{fig:toy3d}
\end{figure}

\parless{Downstream tasks.}
We evaluate inferred sparse codes via a downstream binary prediction task, where the target $t$ depends on the latent $\rvz$ through $t=u(\rvz)\in\{0,1\}$; hence, predicting $t$ from $\rvh$ through $\hat{t}=v(\rvh)$ is a standard supervised proxy for testing whether $\rvh$ preserves task-relevant information about $\rvz$, such as by fitting a logistic head on $\rvh$ and evaluating it through the log-odds $\log\frac{\Pr(t=1\mid \rvh)}{\Pr(t=0\mid \rvh)} = \rva^\top \rvh + a_0$ for weights $\rva$.

\parless{Compositional generalisation.}
Train and test sets differ in which combinations of generative
factors co-occur (\cref{fig:ood_split}).  In terms of supports
$S \subseteq [n]$ of the sparse latent $\rvz$
(cf.\ \cref{eqn:dgp_z}), the ID data excludes a structured subset
of support patterns while the OOD data concentrates on those
withheld combinations.  E.g., consider $k=2$, and three factors $z_1. z_2, z_3$. The ID data contains support pairs $(z_1, z_2)$ and $(z_2, z_3$ while the combination $(z_1, z_3)$ is held out for OOD evaluation (\cref{fig:ood_split}. A downstream label $t = \mathbf{1}\{z_1 > 0.5\}$ depends only on $z_1$, but a model trained ID may learn a shortcut: since $z_1$ always co-occurs with $z_2$, predictor $\rva^T\rvh + a_0$ can achieve high accuracy by tracking features correlated with $z_2$ instead of isolating $z_1$ itself. The OOD split breaks this shortcut by pairing $z_1$ with $z_3$ instead (\cref{fig:overview_boundary}). This follows the withheld-co-occurrence logic central to spurious-correlation benchmarks like Waterbirds~\citep{sagawa2019distributionally}.  Achieving OOD success therefore requires the encoder to recover coordinates where the evidence for $t$ remains stable across different latent factor recombinations. \cref{fig:toy3d} illustrates a toy example example of this failure, where even fully supervised SAEs fail to reconstruct the OOD plane
$(z_1, z_3)$, while sparse coding reconstructs all latents.


\begin{figure}
    \centering
    \includegraphics[width=0.98\linewidth]{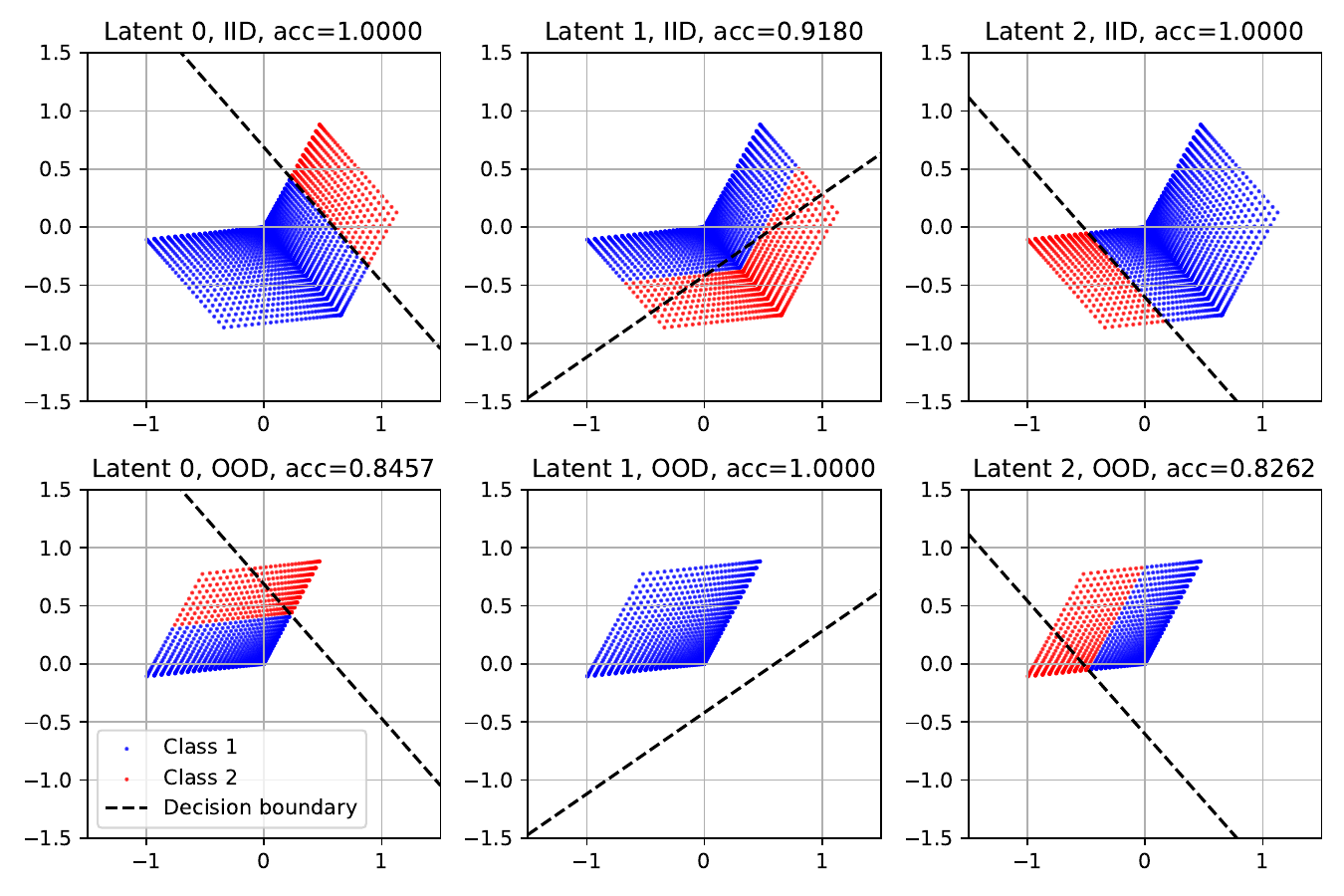}
    \caption{\textbf{Linear probes fail OOD under overcompleteness.}
    Each column sets $t = z_i$.  The linear classifier fits the
    ID decision boundary well, but the compression
    $\rvy = \rmA\rvz$ introduces nonlinearity that is only
    exposed OOD, causing catastrophic generalisation failure
    (columns 1, 3) or, even, poor ID accuracy (column 2).}
    \label{fig:classifiers}
\end{figure}

\parless{Linear Probing under the LRH.} When $\rmW$ is injective ($d_z \leq d_y$), linear separability is invariant: $\rmW$ can rotate or rescale latent space, but a label that is linear in $\rvz$ remains linear in $\rvy = \rmW\rvz$ (\cref{fig:boundary_2d}; also \citep{garg2026featureslanguagemodelstore} Fig.~1).


However, under overcompleteness, even labels that are linearly separable in latent space can become nonlinearly separable in activation space. When $\rmW$ is non-injective, multiple $\rvz$ map to the same $\rvy$, potentially lying on opposite sides of the latent hyperplane; in that case, no \emph{linear} rule in $\rvy$ can match the latent separator without error. This is the geometric phenomenon illustrated in \cref{fig:overview_boundary}b: a hyperplane separator in the full latent space can appear linear on the in-distribution slice of observed mixtures, yet become effectively nonlinear (or even ill-posed \footnote{Ill-posedness arises when $\exists\,\rvz\neq\rvz'$ with $\rmW\rvz=\rmW\rvz'$ but $t(\rvz)\neq t(\rvz')$, in which case $t$ is not a function of $\rvy$ at all.})
in activation space after being transformed through $\rmW$.
Figure \ref{fig:classifiers} illustrates different failure cases of linear probes under this compositional setting.
Under sparse coding, RIP does not make $\rmW$ globally invertible, but it ensures that $\rmW$ does not collapse \emph{sparse} directions, so that sparse latents remain (nonlinearly) distinguishable and pointwise sparse inference is well-posed.

\section{Experiments}
\label{sec:experiments}
We investigate two questions: (i)~is the OOD failure of SAEs a fundamental limitation of sparse inference under superposition, or is it a failure of specific components (dictionary, encoder, or both)?  (ii)~If the problem is solvable in principle, what is the bottleneck in practice?  We use an oracle baseline---FISTA with the ground-truth dictionary---to establish an upper bound on what per-sample inference can achieve, then progressively decompose the gap between this oracle and SAEs.  Methods span four families from fully per-sample to fully amortised inference, plus a linear-probe baseline.  All are evaluated on both in-distribution (ID) and out-of-distribution (OOD) test sets, where the OOD split withholds specific combinations of active latents during training (\cref{fig:ood_split}).

\begin{table}[t]
  \centering
  \caption{Key ratios. The bound requires $\delta \geq C\,\rho \ln(1/\rho)$ for a constant $C>0$: sparser codes (smaller $\rho$) and higher observation dimension (larger $\delta$) make recovery easier.}
  \label{tab:ratios}
  \setlength{\tabcolsep}{4pt}
  \renewcommand{\arraystretch}{1.12}
  \begin{tabularx}{\columnwidth}{@{}c l X@{}}
    \toprule
    Symbol & Definition & Interpretation \\
    \midrule
    $\rho$ &
    $\dfrac{k}{d_h}$ &
    \textbf{Sparsity.} Fraction of non-zeros in $\rvh$. \\
    $\delta$ &
    $\dfrac{d_y}{d_h}$ &
    \textbf{Undersampling.} Ratio of observation dimension $d_y$ to code dimension $d_h$. $\delta<1$ is the overcomplete regime. \\
    \bottomrule
  \end{tabularx}
\end{table}

We consider latent variables $\rvz \in [0,1]^{d_z}$ with at most $k$
non-zero entries, each sampled uniformly on $[0,1]$ when active, observed
through a linear mixing $\rvy = \rmA\rvz$, where
$\rmA \in \mathbb{R}^{d_y \times d_z}$ with $d_y < d_z$. Details in \cref{apx:data}. Two ratios govern recovery difficulty (\cref{tab:ratios}): sparsity
$\rho = k / d_z$ and undersampling $\delta = d_y / d_h$. We evaluate these using the metrics below. The target $t = \mathbf{1}\{z_1 > 0.5\}$ is predicted from inferred codes $\mathbf{h}$.

\parless{Metrics.} We report three quantities on both ID and OOD data (details in \cref{app:metrics}). The \emph{mean correlation coefficient} (\emph{MCC}) \citep{hyvarinen2016unsupervised} evaluates identifiability s.t. MCC${}=1$ for a representation identified up to permutation and rescaling.
\emph{Accuracy} trains a logistic probe on the inferred codes $\rvh$ to predict the binary label $t$, applying the \emph{same} supervised classifier to every method's codes, isolating the effect of the representation. \emph{AUC} is computed per-feature without a trained classifier: for each code dimension, we use the raw activation as a score and compute ROC AUC; the single best feature on ID data is selected and its AUC is reported on both splits. AUC tests whether the label is isolated in an individual feature---a stronger condition than accuracy, which can exploit combinations of features. In the main text we report MCC (identifiability) and Accuracy (for fair downstream comparison between an unsupervised and a supervied learning method). AUC and additional metrics are in the appendix.

\parless{Methods.} We compare methods spanning four families: \emph{sparse coding} (per-sample $\ell_1$
inference with oracle or learned dictionaries), \emph{SAEs}, \emph{frozen decoder} (per-sample
FISTA on a frozen SAE-learned dictionary), and \emph{refined hybrids}
(FISTA warm-started from SAE codes).  The last two families
disentangle dictionary quality from inference: frozen-decoder methods
replace only the encoder while reusing the SAE's dictionary, whereas refined
methods additionally test whether the SAE's output provides a useful
initialisation.  A linear-probe baseline (or, a supervised skyline) operates directly on
$\rvy$.  Architectural and optimisation
details are in \ref{app:sparse-algorithms}. All SAEs (ReLU \citep{cunningham2023sparseautoencodershighlyinterpretable}, JumpReLU \citep{rajamanoharan2024jumpingaheadimprovingreconstruction}, Top-K \citep{gao2024scalingevaluatingsparseautoencoders}, MP \citep{costa2025flat}) share the same decoder dimension and are trained with
identical optimiser settings, differing only in the activation
function governing the encoder's sparsity mechanism.

\parless{Undersampling ratio.} We run a sweep on $\delta$ to observe phase transition in MCC across a grid of latent dimensions and sparsity levels, consistently observing a transition around $C=2$ for $\delta \geq C\rho \ln(1/\rho)$ (details in \cref{apx:undersampling}).

\subsection{Scaling up Latent Dimension Does not close the compositional gap}
\label{sec:exp-vary-latents}

All unsupervised methods face a more challenging problem as $d_z$ grows, as the dictionary has more columns and the space of sparse patterns expands combinatorially. We hypothesise that per-sample methods should degrade more gracefully than SAEs if the bottleneck is amortised inference.

\begin{figure}[ht]
  \centering
  \includegraphics[width=\linewidth]{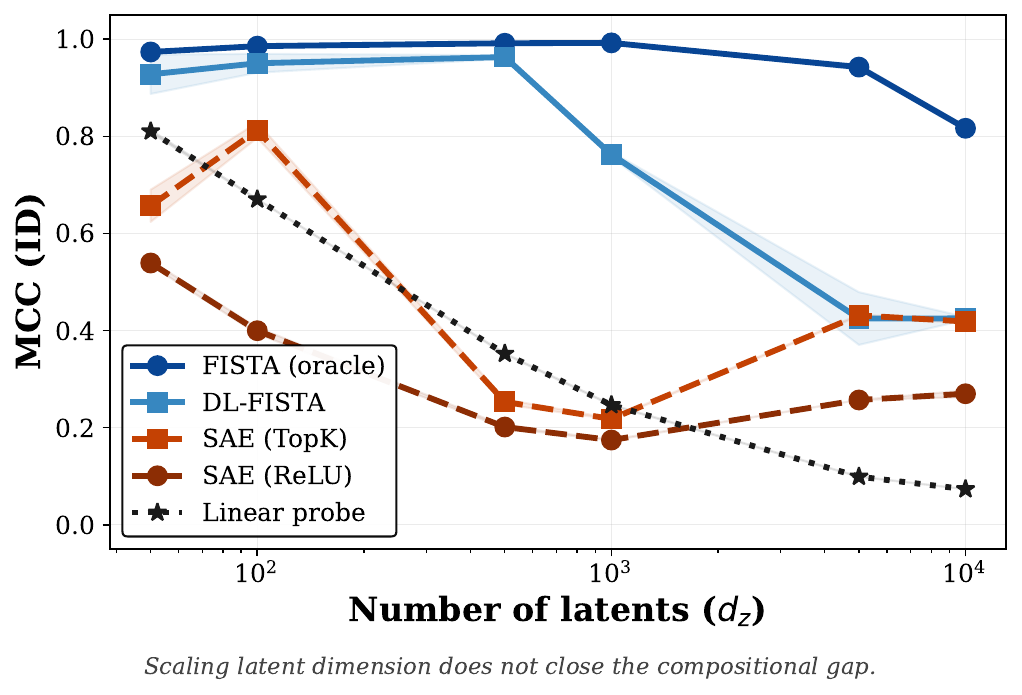}
  \caption{%
    DL-FISTA degrades due to dictionary learning difficulty but maintains a small ID--OOD gap.
  }
  \label{fig:vary-latents}
\end{figure}

\begin{figure}[h!]
  \centering
  \includegraphics[width=\linewidth]{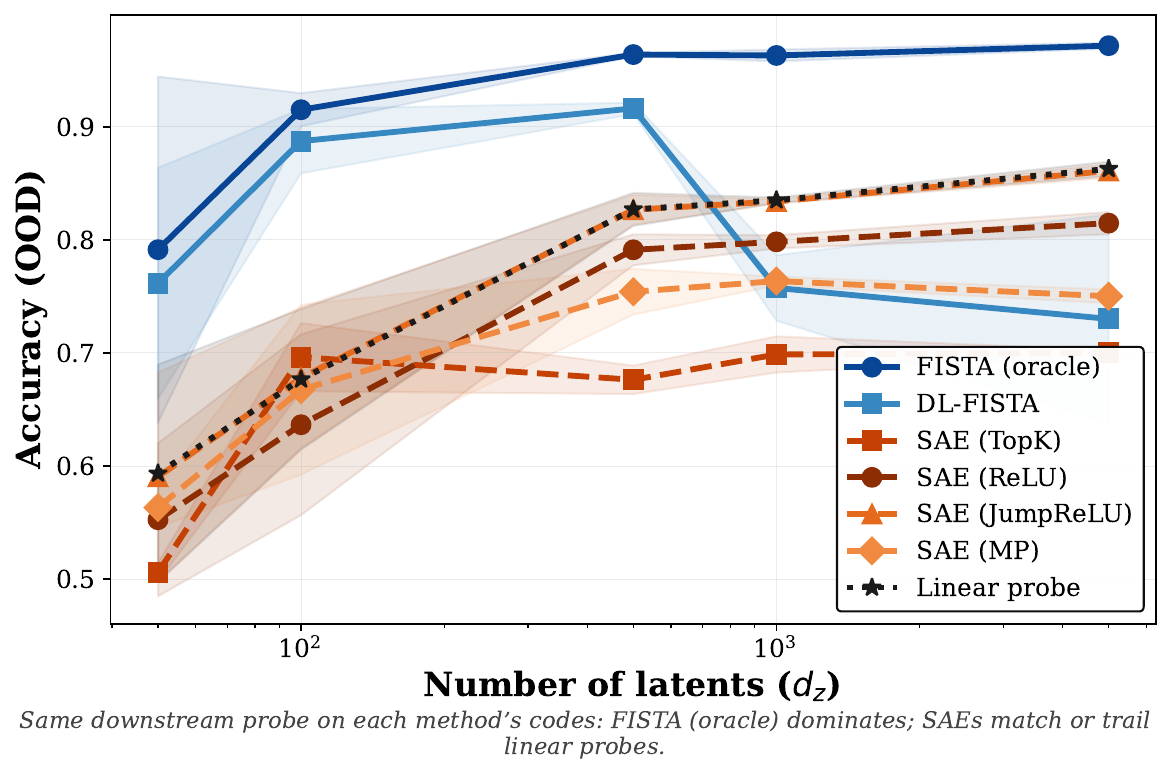}
  \caption{%
DL-FISTA beats linear probes when dictionary learning succeeds but falls
  behind at scale.
  }
  \label{fig:vary-latents-acc}
\end{figure}

We sweep $d_z \in \{50, 100, 500, 1\text{K}, 5\text{K},
10\text{K}\}$ with $k$ and $\delta$ held fixed and report ID MCC (\cref{fig:vary-latents}) and OOD accuracy (\cref{fig:vary-latents-acc}).  As $d_z$ grows, dictionary learning becomes harder for all unsupervised methods. We
   include FISTA with the ground-truth dictionary as an upper bound: it shows what is
  achievable with a perfect dictionary and isolates dictionary learning as the variable
   under test. FISTA (oracle) remains
   near-perfect (MCC $\geq 0.95$, OOD accuracy $\approx 0.97$), confirming the problem
  is solvable at all scales. DL-FISTA degrades in MCC as $d_z$ grows—dropping to
  ${\sim}0.3$ by $d_z = 10\text{K}$—but maintains a small ID–OOD gap, indicating that
  its failure is dictionary quality, not compositional generalisation. When dictionary
  learning succeeds ($d_z \leq 500$), DL-FISTA dominates the linear probe on OOD
  accuracy ($0.92$ vs $0.83$). SAE codes offer no consistent advantage over probing raw activations at
   any scale, consistent with \citet{kantamneni2025sparse}. Varying sparsity $k$ with
  $d_z$ fixed yields the same pattern (appendix, \cref{fig:vary_sparsity_appendix}).

\begin{tcolorbox}[
  colback=gray!34,
  colframe=gray!34, 
  boxrule=0pt,
  arc=0pt,
  left=2pt,right=2pt,top=2pt,bottom=2pt
]
\textbf{Takeaway.}\; The oracle proves the problem is solvable at all scales---the OOD failure is not inherent to superposition. DL-FISTA beats linear probes when dictionary learning succeeds ($d_z \leq 500$) but falls behind at scale, confirming dictionary learning as the universal bottleneck. SAE codes offer no downstream advantage over raw activations, consistent with \citet{kantamneni2025sparse}.
\end{tcolorbox}

\subsection{More Data Does Not Close the Amortisation Gap}
\label{sec:vary-samples}

One issue with the evaluation setup could be that SAEs are simply data-limited: with enough training samples, the encoder should learn to approximate per-sample
inference.  If this were the case, the gap between SAEs and sparse coding would shrink as the training set grows.  However, we find that: (\cref{apx:more-data} for details).

\begin{figure}[h!]
  \centering
  \includegraphics[width=\linewidth]{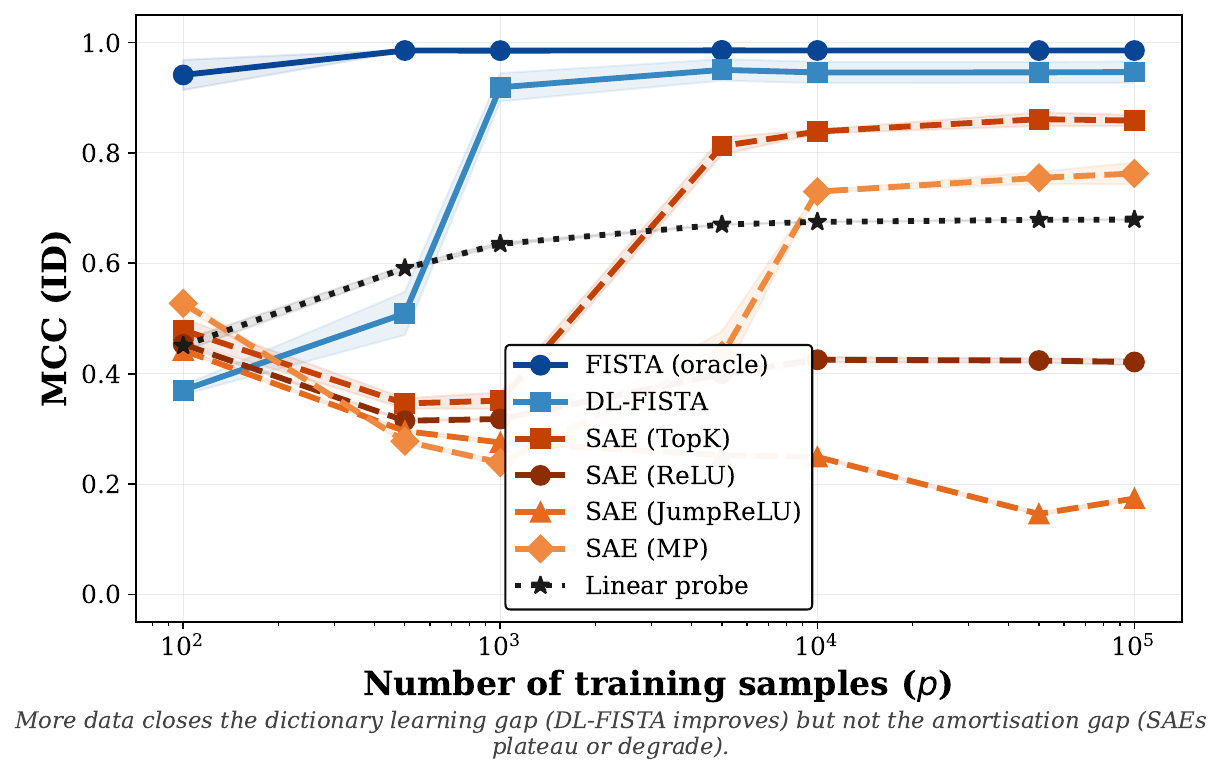}
  \caption{%
    \textbf{More data closes the dictionary learning gap but not the amortisation gap.} DL-FISTA's MCC jumps from ${\sim}0.5$ to ${\sim}0.98$ by $p=10^3$, confirming dictionary learning is sample-limited. $d_z=100$, $k=10$, $d_y=47$.}
    
  
  \label{fig:vary-samples}
\end{figure}

\begin{figure}[h!]
  \centering
  \includegraphics[width=\linewidth]{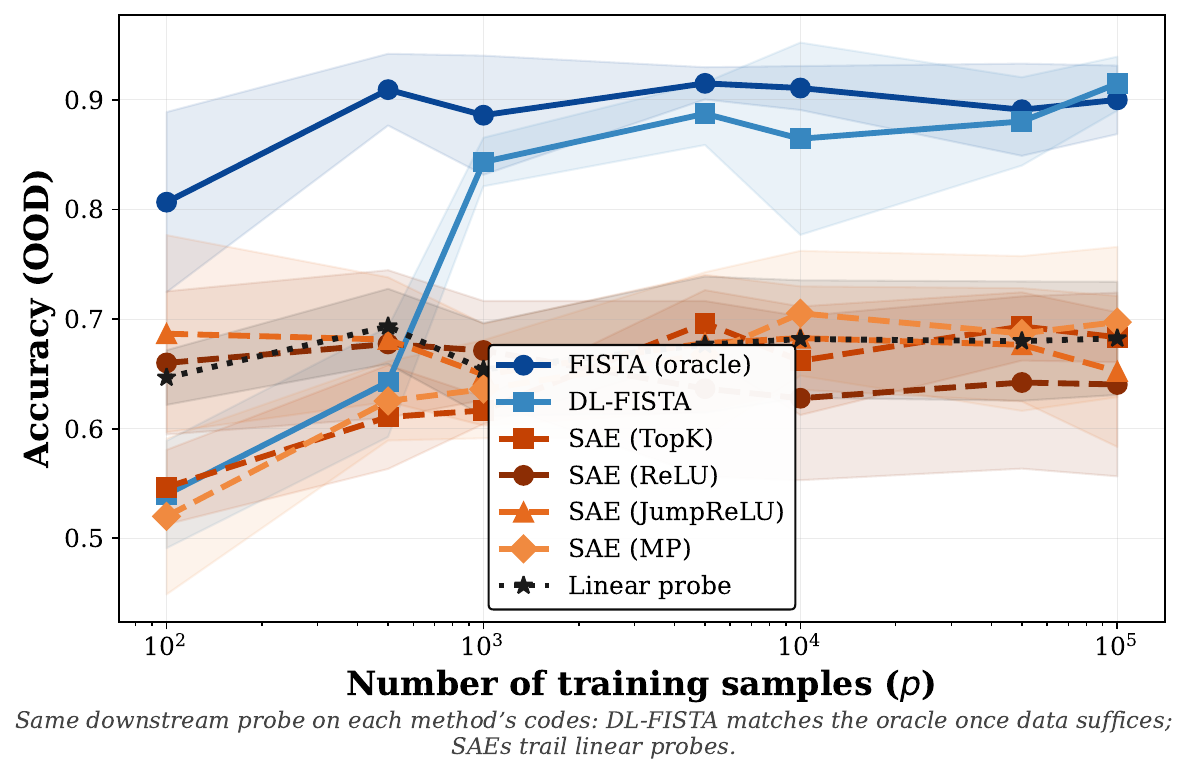}
  \caption{More data closes the gap for DL-FISTA but not for SAEs. 
  $d_z=100$, $k=10$, $d_y=47$.
  }
  \label{fig:vary-samples-acc}
\end{figure}

\begin{tcolorbox}[
  colback=gray!34,
  colframe=gray!34, 
  boxrule=0pt,
  arc=0pt,
  left=2pt,right=2pt,top=2pt,bottom=2pt
]
\textbf{Takeaway.}\; Additional training data closes the dictionary
learning gap (benefiting DL-FISTA) but does
not close the amortisation gap.  When dictionary learning succeeds, DL-FISTA dominates linear probes on the same downstream task, whereas SAE codes trail linear probes regardless of number of training samples.
\end{tcolorbox}


\begin{table*}[t] 
\centering
\footnotesize
\setlength{\tabcolsep}{4pt}
\renewcommand{\arraystretch}{0.95}
\begin{tabular*}{\textwidth}{@{\extracolsep{\fill}}lccc}
\toprule
\textbf{Dataset} & \textbf{DL-FISTA} & \textbf{Top-K SAE} & \textbf{Linear probe} \\
\midrule
\texttt{bias\_in\_bios\_class\_set1} & \textbf{96.96 $\pm$ 0.09} & 95.71 $\pm$ 0.29 & 96.86 $\pm$ 0.00 \\
\texttt{bias\_in\_bios\_class\_set2} & \textbf{95.53 $\pm$ 0.14} & 93.55 $\pm$ 0.05 & 95.28 $\pm$ 0.03 \\
\texttt{bias\_in\_bios\_class\_set3} & \textbf{93.47 $\pm$ 0.05} & 92.21 $\pm$ 0.03 & 93.11 $\pm$ 0.32 \\
\texttt{amazon\_reviews\_mcauley\_1and5} & \textbf{91.89 $\pm$ 0.23} & 89.65 $\pm$ 0.15 & 91.55 $\pm$ 0.16 \\
\texttt{amazon\_reviews\_mcauley\_1and5\_sentiment} & \textbf{96.97 $\pm$ 0.14} & 95.45 $\pm$ 0.46 & 96.80 $\pm$ 0.22 \\
\texttt{github-code} & 96.71 $\pm$ 0.09 & 96.33 $\pm$ 0.19 & \textbf{96.83 $\pm$ 0.04} \\
\texttt{ag\_news} & \textbf{94.93 $\pm$ 0.29} & 94.62 $\pm$ 0.20 & 94.55 $\pm$ 0.18 \\
\texttt{europarl} & 99.87 $\pm$ 0.05 & 99.50 $\pm$ 0.08 & \textbf{99.94 $\pm$ 0.01} \\
\midrule
\textbf{Mean} & \textbf{95.79 $\pm$ 0.06} & 94.63 $\pm$ 0.05 & 95.62 $\pm$ 0.04 \\
\bottomrule
\end{tabular*}
\caption{\textbf{DL-FISTA exceeds the Top-K SAE on all eight datasets and the raw-activation probe on six of eight.} Gemma-2-2B, layer-12 residual stream, $d_h=16{,}384$; test accuracy (\%), mean $\pm$ standard error over three seeds. Higher is better. \tablefootnote{This is a slightly updated version of the published manuscript. See Corrigendum at the end of the paper.}}
\label{tab:saebench_gemma}
\end{table*}

\subsection{The Bottleneck is the Dictionary, Not the Encoder}
\label{sec:disentangle}

The preceding experiments confound two potential sources of SAE failure: a poor dictionary and poor inference.  We isolate these with controlled experiments that hold one component fixed while varying the other,  finding that the SAE’s
dictionary provides a useful starting point for sparse coding
while lagging in inference,  finding that the SAE’s
dictionary provides a useful starting point for sparse coding
while lagging in inference. Across SAEs and DL-FISTA the
bottleneck that emerges is dictionary learning. 

\parless{Swapping the encoder does not close the gap.} Given a trained SAE with decoder $\hat{\rmW}$ and bias $\hat{\rvb}$, we compare three inference strategies on the same test input $\rvy$:
\begin{enumerate}[leftmargin=*,itemsep=2pt]
    \item \textbf{SAE encoder} (baseline): a single forward pass, $\rvh = r(\rvy)$.
    \item \textbf{Frozen decoder}: FISTA on $\rvy - \hat{\rvb}$ using $\hat{\rmW}$ as a fixed dictionary, initialised from $\rvh^{(0)} = \mathbf{0}$.
    \item \textbf{Refined} (warm-start): FISTA on the same frozen $\hat{\rmW}$, initialised from $\rvh^{(0)} = r(\rvy)$.
\end{enumerate}

\cref{fig:frozen-faceted} and \cref{fig:frozen-faceted-acc} show that swapping the SAE encoder for per-sample inference on the same dictionary does not meaningfully improve recovery. To diagnose the learned dictionary, we track cosine similarity between the learned decoder columns and the ground-truth while training in \cref{fig:learning-dynamics} and compare the binary activation pattern of the learned SAE codes with the ground truth to measure support recovery in \cref{fig:support-recovery} (\cref{apx:diagnostics} for details). We also test if the SAE dictionary acts as a useful starting point to accelerate dictionary learning, finding that it provides a clear head-start only when using the learned dictionary of the Top-K SAE (\cref{apx:warmstart,fig:warmstart-decoder}).

\begin{tcolorbox}[
  colback=gray!34,
  colframe=gray!34,
  boxrule=0pt,
  arc=0pt,
  left=2pt,right=2pt,top=2pt,bottom=2pt
]
\textbf{Takeaway.}\; The gap between oracle and other methods seems to be a dictionary learning problem. SAEs fail at both dictionary learning and inference, but the dictionary is the limiting factor: per-sample inference cannot rescue wrong column directions. At scale ($d_z=5000$), DL-FISTA also fails at dictionary learning (cosine ${\sim}0.25$), converging to similar MCC as SAEs. 
\end{tcolorbox}

\subsection{Sparse Coding is Viable on Real LLM Activations}

\begin{table}[h]
\centering
\begin{tabular}{lc}
\hline
\textbf{Model} & \textbf{Test accuracy} \\ \hline
Linear probe (raw activations) & $92.21 \pm 0.16$ \\
SAE & $92.21 \pm 0.02$ \\
\textbf{DL-FISTA} (ours) & \textbf{92.38} $\pm$ \textbf{0.07} \\ \hline
\end{tabular}
\caption{\textbf{DL-FISTA exceeds the Top-K SAE on most eight datasets and the raw-activation probe on six of eight.} Gemma-2-2B, layer-12 residual stream, $d_h=16{,}384$; test accuracy (\%), mean $\pm$ s.d. over three seeds. Higher is better.}
\label{tab:saebench_pythia}
\end{table}

The controlled experiments in previous sections identify the mechanism of failure, but on real activations the true dictionary is unavailable, so recovery cannot be measured directly. What can be measured is the downstream ordering among methods and whether per-sample inference survives realistic dictionary sizes. We
run the sparse-probing task of \citet{kantamneni2025sparse} from SAEBench \citep{karvonen2025saebench}: for each of eight text-classification datasets, a logistic probe is trained on each method's codes and evaluated on held-out text. We evaluate the layer-4 residual stream of Pythia-70M-deduped ($d_y=512$, $d_h=2048$) and the layer-12 residual
stream of Gemma-2-2B ($d_h=16{,}384$), three seeds each; details in \cref{apx:saebench}.

\cref{tab:saebench_pythia} reproduces the null result of \citet{kantamneni2025sparse}: SAE codes match probes on raw activations
($92.21$ for both). DL-FISTA is the only unsupervised method to exceed
raw-activation probing ($92.38 \pm 0.07$ vs.\ $92.21 \pm 0.16$), a small
but seed-consistent margin. On Gemma-2-2B (\cref{tab:saebench_gemma}), DL-FISTA
exceeds the Top-K SAE on all eight datasets, by $0.31$--$2.24$pp (percentage points)(mean $+1.2$pp), and the raw-activation probe on six of eight (mean
$+0.2$pp). This is the expected ordering from our controlled experiments.

Sparse probing is an in-distribution evaluation: no public benchmark
withholds known combinations of natural-language concepts, so compositional generalisation
is not directly testable here. The experiment establishes feasibility and
ordering, not the OOD failure mode. Constructing compositional splits for
natural text is future work.

\begin{tcolorbox}[
  colback=gray!34,
  colframe=gray!34,
  boxrule=0pt,
  arc=0pt,
  left=2pt,right=2pt,top=2pt,bottom=2pt
]
\textbf{Takeaway.}
Per-sample sparse inference is practical on real LLM activations at
$d_h$ up to $16{,}384$ and preserves the ordering from the controlled
experiments: DL-FISTA $>$ raw-activation probes $\geq$ SAE codes. SAE
codes never exceed probing raw activations.
\end{tcolorbox}



\section{Conclusion}

In this paper, we asked two questions: is sparse inference necessary under superposition, and if so, what is the bottleneck?

\textbf{Sparse inference is necessary.} Under superposition, concepts are linearly \emph{represented} in activations but not linearly \emph{accessible}---decision boundaries that are linear in latent space become nonlinear after projection. Linear probes, even with oracle access to ground-truth latents, degrade sharply in identifiability as superposition intensifies (MCC $< 0.1$ at $d_z = 10{,}000$). Per-sample FISTA with the ground-truth dictionary achieves near-perfect OOD recovery (MCC $\geq 0.83$, accuracy $\geq 0.97$) at all scales, proving the problem is solvable within the compressed-sensing framework.

\textbf{The bottleneck is dictionary learning, not inference.} SAE-learned decoder columns point in substantially wrong directions, and replacing the encoder with per-sample FISTA on the same dictionary does not help---the dictionary itself is the limiting factor. On the same downstream task, in our experiments, SAE codes offer no advantage over probing raw activations. Classical dictionary learning (DL-FISTA) dominates linear probes when it succeeds ($d_z \leq 1{,}000$). The gap between oracle and the best unsupervised method identifies \textbf{scalable dictionary learning} as the key open problem for sparse inference under superposition.

Both patterns persist on real LLM activations
(\cref{tab:saebench_pythia,tab:saebench_gemma}), where per-sample inference runs at $d_h = 16{,}384$ without collapse. \footnote{Code for all experiments in this paper can be found \href{https://github.com/shrutij01/compositional-sparse-ood}{here}.}

\subsection*{Acknowledgements} 
This research was partially supported by the Canada CIFAR AI Chair program (Mila) and the NSERC Discovery Grant RGPIN-2025-05123. Simon Lacoste-Julien is a CIFAR Associate Fellow
in the Learning in Machines \& Brains program. This research was enabled in part by computing resources provided by Mila--Québec AI Institute.

\subsection*{Corrigendum}
Table \ref{tab:saebench_gemma} has been updated to provide minor presentation fixes that do not alter the paper's original claims or conclusions. We have added the standard error to the linear probe column, replaced the previously reported standard deviations with standard errors, and corrected the row highlighting.

\bibliography{uai2026-template}

\appendix
\thispagestyle{empty}

\onecolumn

\section{Synthetic data details}
\label{apx:data}
We generate the synthetic data as follows.
We generate a projection matrix \(A \in \mathbb{R}^{m \times n}\) with \(m < n\), where the elements of A are drawn from a standard Normal distribution, in agreement with the Restricted Isometry Property from compressed sensing.
The rows of \(A\) are normalized to have unit norm.
We generate the latent variables \(z \in [0, 1]^n\) with \(k\) non-zero components, where the non-zero components are sampled uniformly from \([0, 1]\).
The observed variables are then generated as \(y = Az\).

When selecting which combinations of latent variables to be active, which $k$ out of $n$, we consider the particular "out-of-variable" case for OOD generalizations, where some combinations of the variables are not available in the training data.
The number of OOD variables is $n/2$. Then, we consider two possibilities:
\begin{itemize}
    \item \textbf{ID data}: Divided into two cases:
    \begin{itemize}
        \item The first latent variable is active and the other $k-1$ are drawn between the variables of indices $[2, n/2]$.
        \item The first latent variable is not active. The $k$ active indices are drawn from the fool pool of indices $[2, n]$.
    \end{itemize}
    \item \textbf{OOD data}: The first latent variable is active and the other $k-1$ are drawn between the variables of indices $[n/2 + 1, n]$.
\end{itemize}
The training set consists of ID data, and the test set consists of OOD data.




\section{Training details}
We implement the models in PyTorch and train them on a single NVIDIA A100 GPU.


\section{Sparse Inference Methods for Interpretability}
\label{app:sparse-inference}

Sparse autoencoders (SAEs) have become the dominant tool for extracting
interpretable features from neural network representations
\citep{bricken_towards_2023, cunningham2023sparseautoencodershighlyinterpretable}.  An SAE
decomposes an activation $\rvy \in \sR^{d_y}$ as $\rvy  \approx \rmD \rvh$,
where $\rmD  \in \sR^{d_h \times d_y}$ is an overcomplete dictionary ($d_h > d_y$)
and $\rvh \in \sR^{d_h}$ is a sparse code whose nonzero entries identify the
active features.  The quality of the interpretation depends entirely on
the quality of $\rvh$: if the codes are wrong, the resulting feature
attribution is wrong, regardless of reconstruction fidelity.

Standard SAEs compute codes in a single feedforward pass,
$\rvh = \sigma(\rmW^\top (\rvy - b_{\mathrm{pre}}) + b)$, where $\sigma$
is ReLU or a top-$k$ or JumpReLU activation.  This is an \emph{amortised}
approximation to the sparse inference problem
\begin{equation}
  \label{eq:lasso}
  \rvh^* = \argmin_{\rvh} \;\tfrac{1}{2}\|\rvy - \rmD\rvh\|_2^2
  + \lambda \|\rvh\|_1\,,
\end{equation}
which is the Lasso \citep{tibshirani1996regression}, a convex problem with a
unique solution (under mild conditions on $\rmD$).  The amortisation gap
$\rvh - \rvh^*$ is a structured error that is largest precisely when
features are correlated or hierarchically organised
\citep{costa2025flat,chanin2025feature}---the regimes most relevant
to real neural network representations.

Below we compare three inference strategies that move progressively
closer to solving \cref{eq:lasso}: FISTA, LISTA, and Matching Pursuit.  The comparison focuses on properties that
matter for interpretability rather than reconstruction.

\subsection{Algorithms}
\label{app:sparse-algorithms}

\paragraph{FISTA (Fast Iterative Shrinkage-Thresholding).}
FISTA \citep{beck2009fast} solves \cref{eq:lasso} by alternating a
gradient step on the reconstruction loss with the proximal operator for
the $\ell_1$ penalty (soft-thresholding $S_\lambda$), accelerated by
Nesterov momentum.  Let $\rvh^{(t)}$ denote the code estimate at
iteration $t$ and $\rvq^{(t)}$ a momentum-extrapolated point:
\begin{align}
  \rvq^{(t)} &= \rvh^{(t)} + \frac{t_k - 1}{t_{k+1}}
    \bigl(\rvh^{(t)} - \rvh^{(t-1)}\bigr)\,,
    \label{eq:fista-momentum}\\
  \rvh^{(t+1)} &= S_{\eta\lambda}\!\bigl(\rvq^{(t)}
    - \eta\, \rmD^\top(\rmD\,\rvq^{(t)} - \rvy)\bigr)\,,
    \label{eq:fista-update}
\end{align}
where $\eta \leq 1/\|\rmD^\top \rmD\|_{\mathrm{op}}$ is the step size.
Every iteration updates \emph{all} $d_h$ coefficients simultaneously.
The support (which atoms are active) is fluid: a coefficient can be
driven to zero by soft-thresholding at step $t$ and revived at step
$t' > t$.  Convergence to the global optimum is guaranteed at rate
$O(1/t^2)$ \citep{beck2009fast}.  There are no learned parameters;
the algorithm is fully determined by $\rmD$ and $\lambda$.

\emph{Practical note.}\enspace
Precomputing $\rmW = \rmI - \eta\, \rmD^\top \rmD$ and $\rvb = \eta\, \rmD^\top \rvy$
reduces each iteration to $\rvh^{(t+1)} = S_{\eta\lambda}(\rmW\,\rvh^{(t)} + \rvb)$:
a single matrix--vector multiply plus elementwise thresholding, both
fully batchable on GPU.

\paragraph{LISTA (Learned ISTA).}
LISTA \citep{gregor2010learning} takes the ISTA update (i.e.\
\cref{eq:fista-update} without momentum) and untethers its parameters
from $\rmD$.  Each layer $t$ computes:
\begin{equation}
  \label{eq:lista}
  \rvh^{(t+1)} = S_{\theta_t}\!\bigl(\rmW_t\, \rvh^{(t)}
    + \rmB_t\, \rvy\bigr)\,,
\end{equation}
where $\rmW_t \in \sR^{d_h \times d_h}$, $\rmB_t \in \sR^{d_h \times d_y}$, and
$\theta_t \in \sR^{d_h}$ are \emph{free parameters learned by
backpropagation}, independently at each layer.  In ISTA,
$\rmW_t = \rmI - \eta\, \rmD^\top \rmD$, $\rmB_t = \eta\, \rmD^\top$, and
$\theta_t = \eta\lambda\,\mathbf{1}$ for all $t$; LISTA relaxes these
constraints, allowing the network to learn iteration-dependent
acceleration.  Empirically, LISTA matches FISTA's solution quality in
10--20 layers rather than 100+ iterations
\citep{gregor2010learning}.

Crucially, LISTA retains the structural properties of ISTA/FISTA: all
coefficients are updated jointly at every layer, soft-thresholding
provides a continuous sparsity mechanism, and the architecture is fully
parallelisable across the batch dimension.  The dictionary $\rmD$ (or its
learned analogue in $\rmW_t, \rmB_t$) can be trained end-to-end.

\paragraph{MP-SAE (Matching Pursuit SAE).}
MP-SAE \citep{costa2025flat} unrolls the classical matching pursuit
algorithm \citep{mallat1993matching} into a differentiable encoder.
Let $\rvd_j$ denote the $j$-th column of $\rmD$.
At each step $t = 1, \ldots, T$:
\begin{align}
  j^{(t)} &= \argmax_{j \in \{1,\ldots,d_h\}}\; \rvd_j^\top \rvr^{(t-1)}\,,
    \label{eq:mp-select}\\
  h_{j^{(t)}} &= \rvd_{j^{(t)}}^\top \rvr^{(t-1)}\,,
    \label{eq:mp-coeff}\\
  \rvr^{(t)} &= \rvr^{(t-1)} - h_{j^{(t)}}\, \rvd_{j^{(t)}}\,,
    \label{eq:mp-residual}
\end{align}
where $\rvr^{(0)} = \rvy - b_{\mathrm{pre}}$.  One atom is selected per
step; its coefficient is computed by projection onto the residual; the
residual is updated by subtracting the selected atom's contribution.
Previous coefficients are never revised.  The dictionary is trained
end-to-end via backpropagation through the unrolled steps.

MP-SAE approximately solves a different problem from \cref{eq:lasso}:
it targets $\min_{\rvh} \|\rvy - \rmD\,\rvh\|_2^2$ subject to
$\|\rvh\|_0 \leq T$, which is NP-hard; matching pursuit is a greedy
approximation with no global optimality guarantee.
\subsection{Comparison on Interpretability-Relevant Axes}
\label{app:sparse-comparison}

\paragraph{Well-posedness of codes.}
FISTA computes the unique minimiser of the Lasso objective
\cref{eq:lasso}.  The codes are \emph{defined} by a convex optimisation
problem: one can point to the objective and state precisely what the
codes mean.  LISTA approximates this same solution with learned
acceleration.  MP-SAE computes the output of a greedy procedure that
does not correspond to the global minimum of any fixed objective; the
codes depend on the selection order, which is itself a function of the
dictionary geometry and the input.  For identifiability --- where
``meaning'' is invariance across the equivalence class of valid solutions
--- the distinction matters: the Lasso solution is unique and
characterisable; the MP output is not.

\paragraph{Joint coefficient adjustment.}
FISTA and LISTA update all $d_h$ coefficients at every iteration.  If
activating atom $i$ changes the optimal coefficient for atom $j$ (as occurs whenever $\rvd_{i}^{\top} \rvd_j \neq 0$), subsequent iterations correct
for this.  MP-SAE sets each coefficient once, at the step the atom is selected, and never revises it.  Consider $\rvy = \alpha_1 \rvd_1 + \alpha_2 \rvd_2$ with $\rvd_1^\top \rvd_2 = \rho > 0$.  MP selects $\rvd_1$ first
(assuming $\alpha_1 > \alpha_2$) and assigns $h_1 = \rvd_1^\top \rvy = \alpha_1 + \alpha_2 \rho$, which is inflated by $\rvd_2$'s contribution
leaking through the correlation.  The coefficient $h_2$ computed on the
residual is correspondingly deflated.  FISTA converges to the correct
$(\alpha_1, \alpha_2)$ because it jointly adjusts both coefficients
across iterations.

\paragraph{Support dynamics.}
In FISTA/LISTA, the active set (support of $\rvh$) is fluid: an atom
can be activated, deactivated, and reactivated across iterations as the
algorithm converges.  This self-correction is critical when the initial
support estimate is wrong.  In MP-SAE, the support grows monotonically
--- once an atom is selected, it remains active.  There is no mechanism
to deselect an incorrectly chosen atom, and the error propagates through
all subsequent residuals.

\paragraph{Correlated and hierarchical features.}
Standard SAEs compute all inner products $\langle \rvd_j, \rvy \rangle$
simultaneously and threshold, making all activation decisions in
parallel.  This implicitly assumes quasi-orthogonality of the dictionary
\citep{costa2025flat}: if $\rvd_i$ and $\rvd_j$ are correlated, activating
$\rvd_i$ should reduce the evidence for $\rvd_j$, but the one-shot encoder
cannot express this.

MP-SAE fixes the conditioning problem via the residual update
\cref{eq:mp-residual}: after selecting $\rvd_i$, atom $\rvd_j$ is evaluated
against the residual $\rvr$ rather than the raw input, so correlated
atoms no longer double-count shared variance.  This is also why MP-SAE
naturally recovers hierarchical structure: the first iteration selects
the dominant (coarse) feature, and subsequent iterations select
progressively finer features on the residual.

FISTA/LISTA handle correlated features correctly \emph{and} with correct
magnitudes, because the joint coefficient update avoids the inflation
effect described above.  However, they do not provide a natural ordering
over features --- all coefficients converge simultaneously rather than
being produced in sequence.  When a hierarchy readout is desired, the
convergence order or coefficient magnitude in FISTA can serve as a proxy,
but the sequential atom selection in MP provides this more directly.

\paragraph{Computational cost.}
\Cref{tab:sparse-cost} summarises the per-step and total cost for a
batch of $B$ samples.  FISTA and LISTA are fully parallelisable across
the batch; MP-SAE's sequential atom selection (the $\argmax$ in
\cref{eq:mp-select} depends on the previous step's residual)
limits GPU utilisation.  LISTA compensates for its per-step cost by
converging in far fewer steps than FISTA.

\begin{table}[h]
\centering
\caption{Computational comparison of sparse inference methods.
  $d_y$: input dimension, $d_h$: dictionary size, $T$: number of
  steps/layers, $B$: batch size.}
\label{tab:sparse-cost}
\small
\begin{tabular}{@{}lccc@{}}
\toprule
 & \textbf{FISTA} & \textbf{LISTA} & \textbf{MP-SAE} \\
\midrule
Per-step cost & $O(B\,d_h^2)$ & $O(B\,d_h^2)$ & $O(B\,d_h)$ \\
Typical steps $T$ & 100--300 & 10--20 & $k$ (active atoms) \\
GPU parallelism & Full & Full & Limited \\
End-to-end trainable & No\footnotemark & Yes & Yes \\
\bottomrule
\end{tabular}
\end{table}
\footnotetext{FISTA itself is not trained; the dictionary is updated in
a separate alternating minimisation step.  However, FISTA can be used at
evaluation time with a dictionary trained by any method, including an
SAE.}
\paragraph{Trainability.}
LISTA and MP-SAE are both end-to-end trainable: the dictionary is
updated by backpropagation through the unrolled inference steps, using
standard deep learning optimisers.  FISTA requires alternating
optimisation --- an outer loop updating $\rmD$ and an inner loop running
FISTA to convergence for each batch --- which is slower but provides
stronger guarantees on code optimality.  A practical middle ground is to
train the dictionary using a standard SAE or LISTA, then compute codes
at evaluation time using FISTA with the learned dictionary, optionally
warm-started from the encoder's output.

\section{Experimental results}
\label{app:experiments}

\subsection{Metric definitions and what each diagnostic isolates}
\label{app:metrics}

We use three levels of evaluation metrics, each isolating a different component of the SAE pipeline. \cref{tab:metric-summary} summarises the distinctions.

\begin{table}[h]
\centering
\caption{Summary of evaluation metrics and the questions they answer.}
\label{tab:metric-summary}
\small
\setlength{\tabcolsep}{4pt}
\begin{tabularx}{\linewidth}{@{}l l X@{}}
\toprule
\textbf{Metric} & \textbf{Operates on} & \textbf{Question it answers} \\
\midrule
MCC & Codes vs $\rvz$ (samples) & Do learned code dimensions track the same variation as the true latents, up to permutation and rescaling? \\
Accuracy & Probe on codes vs labels & Does a logistic classifier on the codes predict the label OOD? (Same classifier for all methods---fair comparison.) \\
Per-feature AUC & Codes vs labels (samples) & Does a single code dimension separate the binary label? \\
\midrule
Cosine similarity & Decoder $\hat{\rmW}$ vs $\rmA$ (weights) & Do the dictionary atoms point in the same directions as the ground-truth columns? \\
Norm ratio & Decoder $\hat{\rmW}$ vs $\rmA$ (weights) & Are the dictionary atoms the correct magnitude? \\
\midrule
Support precision & Codes vs $\rvz$ (per-sample binary) & Of the features the encoder activates, how many are truly active? \\
Support recall & Codes vs $\rvz$ (per-sample binary) & Of the truly active features, how many does the encoder find? \\
\bottomrule
\end{tabularx}
\end{table}

\parless{Accuracy vs AUC: fairness considerations.} The linear-probe baseline uses supervised ridge regression from $\rvy$ to the ground-truth $\rvz$, producing $d_z$-dimensional codes where each dimension is explicitly trained to track one latent. Per-feature AUC directly benefits from this alignment: each output dimension is already optimised to separate its target, giving the linear probe an inherent advantage over unsupervised methods whose code dimensions need not correspond to individual latents. Accuracy---training a logistic probe on each method's inferred codes---provides a fairer comparison because it applies the same downstream classifier to all methods and can exploit feature combinations, not just individual dimensions. In the main text we therefore report accuracy for downstream comparisons and reserve AUC for the appendix.

\parless{MCC (end-to-end).} The mean correlation coefficient \citep{hyvarinen2016unsupervised} computes the Pearson correlation between each code column and each ground-truth latent column across samples, then finds the best one-to-one matching via the Hungarian algorithm. It measures overall recovery quality: MCC${}\!=\!1$ when codes reproduce the true latents up to permutation and rescaling. Pearson correlation is invariant to linear scaling, so MCC does not penalise magnitude differences. However, MCC conflates dictionary quality and encoder quality---a low MCC does not tell you whether the dictionary atoms are wrong or the encoder is selecting the wrong atoms.

\parless{Dictionary diagnostics (model-level).} Cosine similarity and norm ratio between matched decoder columns and ground-truth dictionary columns isolate \emph{dictionary quality} independent of any test data or encoder. If cosine is high but MCC is low, the dictionary is good but the encoder fails. If cosine is low, the dictionary itself is the bottleneck regardless of the encoder. In our experiments, norm ratios are ${\approx}1.0$ throughout (column magnitudes are correct) while cosine similarity varies widely across SAE types ($0.33$--$0.93$), pinpointing the error as directional.

\parless{Support diagnostics (encoder-level).} Support precision and recall compare the binary nonzero pattern of codes against ground truth (after Hungarian-matching via the decoder columns). These isolate \emph{encoder quality} on the feature-selection task: does the encoder activate the right atoms, independent of magnitude accuracy? MCC cannot distinguish ``activated the wrong 10 features'' from ``activated all 100 features and relied on magnitude differences''---the support diagnostics can. In our experiments, ReLU and JumpReLU activate ${\sim}90\%$ of all features (precision ${\sim}0.1$), revealing that they are not performing sparse selection at all.

\parless{Why all three levels are needed.} Consider two failure modes that produce the same MCC:
\begin{enumerate}[leftmargin=*,itemsep=2pt]
\item An SAE with correct dictionary directions but an encoder that activates the wrong atoms. Dictionary cosine would be high; support precision would be low.
\item An SAE with wrong dictionary directions but an encoder that compensates by routing information through correlated atoms. Dictionary cosine would be low; support precision could be moderate.
\end{enumerate}
MCC alone cannot distinguish these. The layered diagnostics---dictionary geometry (cosine) $\to$ feature selection (support) $\to$ overall recovery (MCC)---tell you \emph{where} in the pipeline things break.

\subsection{The Amortisation gap persists across undersampling ratios}
\label{apx:undersampling}
\begin{figure}
    \centering
    \includegraphics[width=\linewidth]{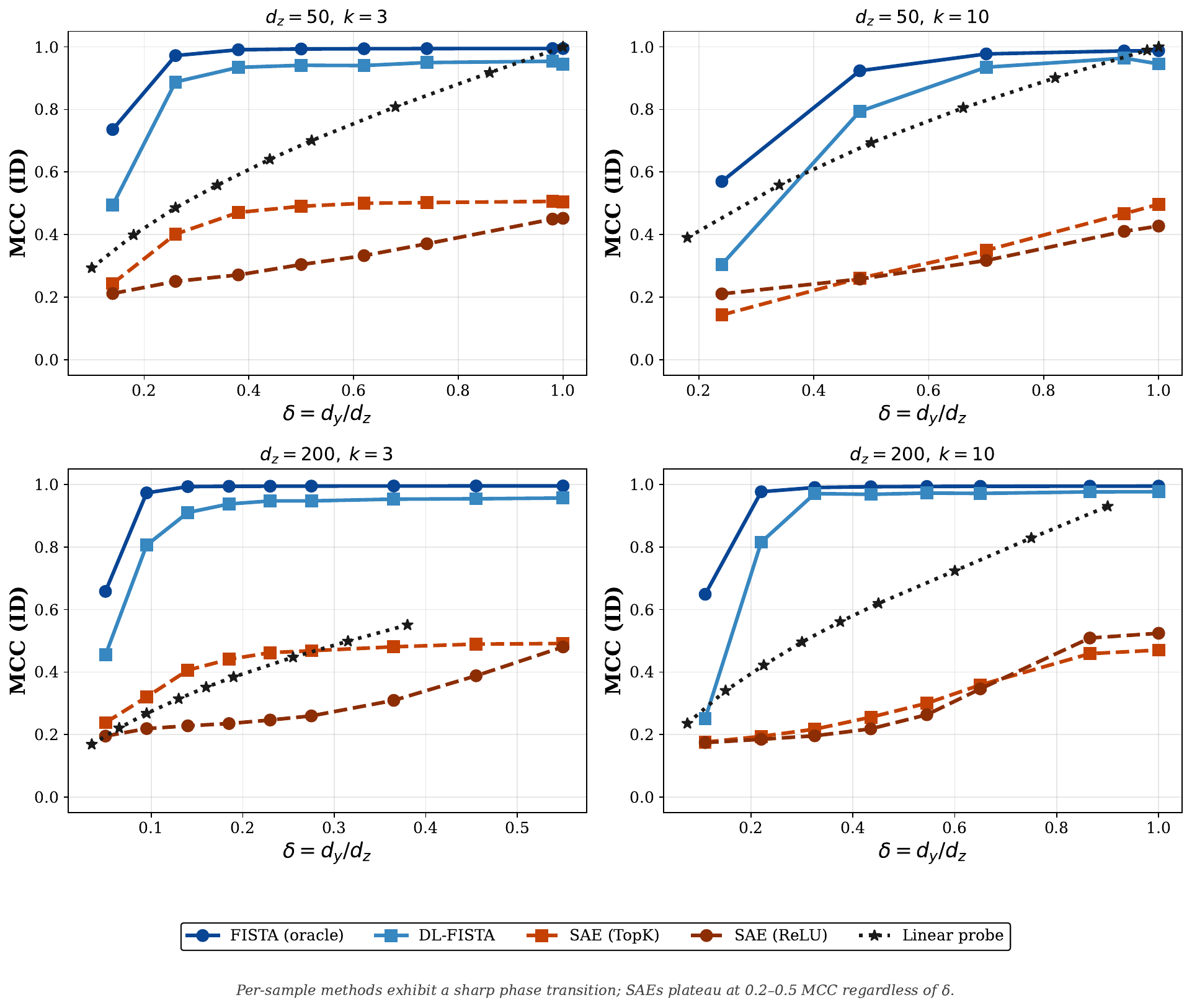}
    \caption{\textbf{The amortisation gap persists across undersampling ratios.} Per-sample methods (FISTA) exhibit a sharp phase transition to near-perfect MCC once $\delta$ exceeds the compressed-sensing threshold; SAEs plateau at $0.2$--$0.5$ MCC regardless of $\delta$. Each panel shows a different $(d_z, k)$ combination.}
    \label{fig:phase}
\end{figure}

Compressed sensing theory predicts a phase transition in sparse recovery \citep{candes2006}: once the number of suitably random observations $d_y$ exceeds $O(k \ln(d_z/k))$, per-sample $\ell_1$ methods recover the latent code exactly.  
We sweep the undersampling ratio $\delta = d_y/d_h$ across a grid of latent dimensions $d_z \in \{50, 100, 200\}$ and sparsity levels $k \in \{3, 5, 10\}$, and report ID MCC (\cref{fig:phase}). Phase transition with other metrics are reported in \cref{app:experiments}.
If SAEs solved the same sparse inference problem, they would exhibit the same transition and reach the same asymptotic performance. A persistent gap between the two would reveal the cost of amortising inference into a fixed encoder.

Per-sample methods exhibit the predicted phase transition. FISTA (oracle) transitions sharply to near-perfect MCC once $\delta$ passes a critical threshold, and the transition sharpens with $d_z$, matching the theoretical prediction. The empirical transition point is consistent with the constant $C \approx 2$ used in the compressed-sensing bound $\delta \geq C \rho \ln(1/\rho)$, validating the bound's practical relevance. DL-FISTA follows the same curve, shifted right by the cost of learning the dictionary.  SAE variants also improve with $\delta$---they are not insensitive to the undersampling ratio---but plateau at $0.2$--$0.5$ MCC, well below the near-perfect recovery that per-sample methods achieve in the same regime.  Crucially, the gap does not close at high $\delta$, implying that even when the problem is well within the regime where compressed sensing guarantees exact recovery, the amortised encoder remains the bottleneck.

\begin{tcolorbox}[
  colback=gray!34,
  colframe=gray!34, 
  boxrule=0pt,
  arc=0pt,
  left=2pt,right=2pt,top=2pt,bottom=2pt
]
\textbf{Takeaway.}\; Both SAEs and sparse coding benefit from less aggressive undersampling (higher $\delta$), but SAEs saturate far below per-sample methods.
\end{tcolorbox}

\subsection{SAE Bench Sparse Probing}
\label{apx:saebench}
\parless{Pythia-70m-deduped:}
The setup follows the SAE Bench protocol (d\_model=512, d\_sae=2048, L1 weight 0.1, 50k training tokens from NeelNanda/pile-10k, evaluation across 8 text classification datasets with 4k train / 1k test examples). DL-FISTA uses 100 inner FISTA iterations with dictionary updates every 50 steps during training, and 200 FISTA iterations at evaluation. We evaluate the models on 3 different seeds each and report the mean and standard deviation (extending the original SAE Bench results that had been published only for seed 42).

\parless{Gemma-2-2b:} Layer 12 residual stream, d\_sae=16384, $\lambda$=0.1, FISTA inner=200 at evaluation.

\subsection{Dictionary is the Bottleneck}
\label{apx:diagnostics}

\parless{Swapping the encoder does not close the gap.} Given a trained SAE with decoder $\hat{\rmW}$ and bias $\hat{\rvb}$, we compare three inference strategies on the same test input $\rvy$:
\begin{enumerate}[leftmargin=*,itemsep=2pt]
    \item \textbf{SAE encoder} (baseline): a single forward pass, $\rvh = r(\rvy)$.
    \item \textbf{Frozen decoder}: FISTA on $\rvy - \hat{\rvb}$ using $\hat{\rmW}$ as a fixed dictionary, initialised from $\rvh^{(0)} = \mathbf{0}$.
    \item \textbf{Refined} (warm-start): FISTA on the same frozen $\hat{\rmW}$, initialised from $\rvh^{(0)} = r(\rvy)$.
\end{enumerate}

\begin{figure*}
  \centering
  \includegraphics[width=0.95\textwidth]{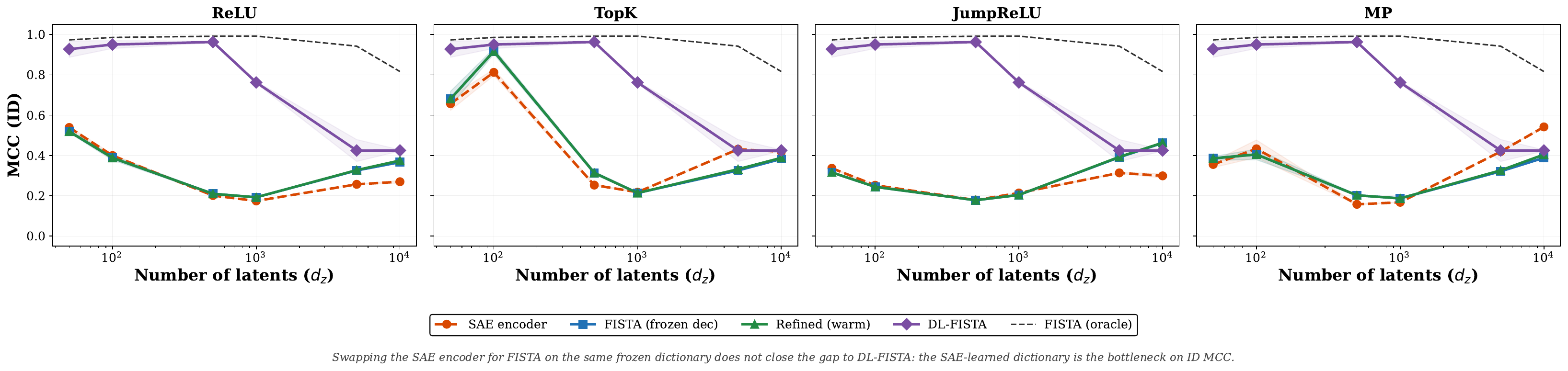}
  \caption{%
    \textbf{The SAE-learned dictionary is the bottleneck, not the encoder.} Each panel shows one SAE type. Swapping the SAE encoder for FISTA on the same frozen dictionary (blue) or warm-starting from SAE codes (green) does not close the gap to DL-FISTA (purple), which learns its own dictionary. The oracle (gray dashed) confirms the problem is solvable. $k=10$, $n=5000$.
  }
  \label{fig:frozen-faceted}
\end{figure*}

\begin{figure*}[h!]
  \centering
  \includegraphics[width=0.95\textwidth]{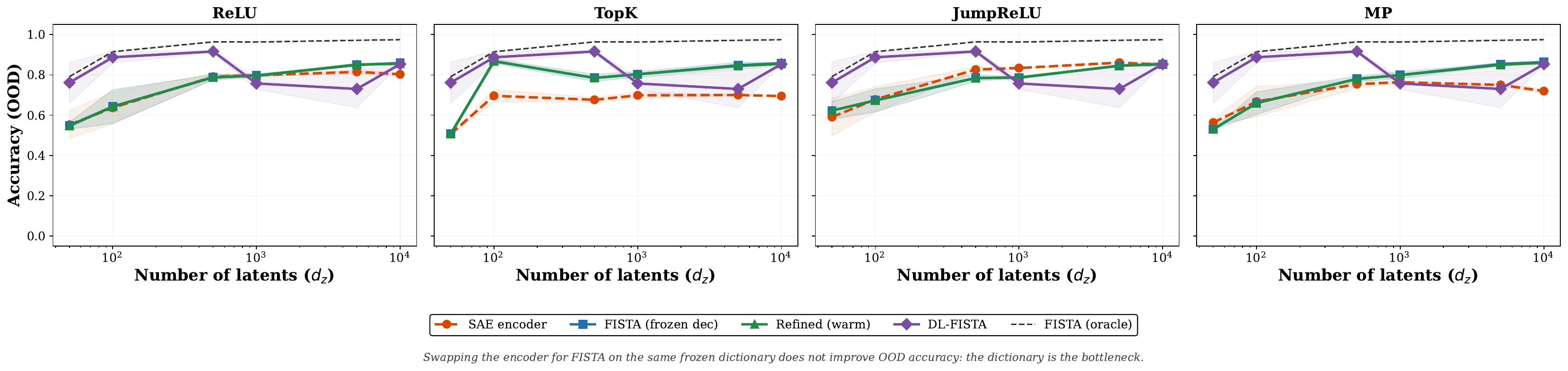}
  \caption{%
    \textbf{Swapping the encoder does not improve OOD accuracy: the dictionary is the bottleneck.} Same layout as \cref{fig:frozen-faceted}. Replacing the SAE encoder with FISTA on the same frozen dictionary yields no consistent accuracy gain. The gap to DL-FISTA confirms that the dictionary directions, not the inference procedure, limit OOD performance. $k=10$, $p=5000$.
  }
  \label{fig:frozen-faceted-acc}
\end{figure*}

All three use the \emph{same} dictionary---only the inference procedure differs. \cref{fig:frozen-faceted} shows the comparison alongside DL-FISTA (which learns its own dictionary independently). The result shows that SAE encoders $\approx$ frozen FISTA $\approx$ refined ($\sim 0.15$--$0.4$ MCC), while DL-FISTA (which learns its own dictionary via classical alternating minimisation) achieves an MCC of $\sim 0.75$--$0.95$. This implies that swapping the encoder for per-sample inference on the same dictionary does not meaningfully improve recovery. The SAE-learned dictionary itself is the limiting factor. This conclusion is robust to the choice of FISTA regularisation strength $\lambda$: sweeping $\lambda$ across three orders of magnitude yields at most modest improvements over the SAE encoder, far below the oracle at every $\lambda$ (appendix, \cref{app:lambda}). Note that DL-FISTA also degrades at large $d_z$ (to ${\sim}0.3$ at $d_z=10\text{K}$, \cref{sec:exp-vary-latents}), reflecting the inherent difficulty of dictionary learning at scale---but it degrades uniformly across ID and OOD, unlike SAEs which exhibit a compositional generalisation failure. The linear probe, operating directly on $\rvy$ without sparse inference, degrades in MCC even faster as superposition intensifies (\cref{fig:vary-latents}), but maintains surprisingly high downstream accuracy (\cref{fig:vary-latents-acc}) because it is supervised for that specific task---a divergence that illustrates how single-task metrics can mask identifiability failure.

At $100$ FISTA iterations (sufficient for convergence of the convex Lasso objective) warm-starting from SAE codes yields the same solution as cold-starting from zeros, confirming that the per-sample optimisation converges regardless of initialisation. The one exception is TopK, whose SAE decoder achieves ${\sim}0.87$ OOD MCC on its own (comparable to DL-FISTA at small $d_z$), consistent with TopK's structurally enforced sparsity producing a more faithful dictionary.

\begin{figure*}
  \centering
  \includegraphics[width=0.95\textwidth]{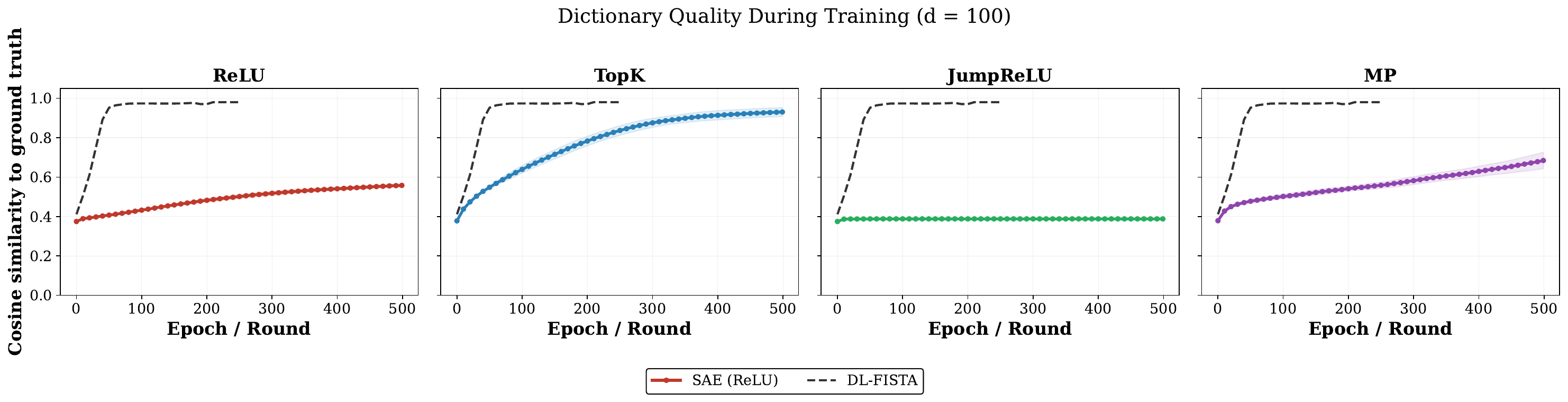}
  \caption{\textbf{DL-FISTA learns correct dictionary directions; SAEs do not.} SAE decoder cosine to ground truth vs training epoch (coloured) and DL-FISTA dictionary cosine vs outer round (gray dashed) at $d_z=100$. At large $d_z$ ($\geq 5000$), neither method converges (\cref{fig:learning-dynamics-n5000}).}
  \label{fig:learning-dynamics}
\end{figure*}

\parless{The dictionary fails because the columns point in the wrong directions.} We track cosine similarity between decoder columns and ground truth during training (\cref{fig:learning-dynamics}) and decompose column-level errors at convergence (\cref{fig:dict-diagnostics-appendix}). At $d_z=100$, DL-FISTA converges to high cosine ($>0.9$) while SAEs plateau earlier---the SAE optimisation landscape is harder even when the problem is tractable for alternating minimisation. At $d_z=5000$, neither method finds the right directions (cosine ${\sim}0.2$--$0.35$), implying that the SAE dictionaries do not overfit or drift, they simply never converge. Norm ratios remain ${\approx}1.0$ throughout, and re-normalising columns or substituting oracle norms does not improve MCC ( \cref{fig:dict-quality-appendix}). TopK is an outlier at small $d_z$ (cosine $0.93$, $90\%$ of columns close to ground truth at $d_z=100$), but this advantage vanishes at scale.

\parless{Why the encoder fails: wrong support.} We measure support recovery---whether the SAE activates the correct features---by comparing the binary nonzero pattern of SAE codes against the ground truth (after Hungarian matching). At $d_z=5000$ with true sparsity $k=10$: ReLU activates ${\sim}120$ of $5000$ features (precision $0.009$); JumpReLU activates ${\sim}300$ (precision $0.005$). Even TopK, which activates the correct number of features (${\sim}10$), selects almost entirely wrong atoms (precision $0.03$). Re-estimating magnitudes via least-squares on the SAE's support makes things \emph{worse} (\cref{fig:support-recovery}), confirming the support itself is incorrect. At smaller $d_z$, TopK's support precision is substantially better ($0.49$ at $d_z=100$; \cref{fig:support-diagnostics-appendix}), explaining its stronger performance at small scale.

\subsection{The SAE Dictionary is a Useful Starting Point}
\label{apx:warmstart}

Although the SAE decoder is a poor standalone dictionary, it may still encode useful structure that accelerates dictionary learning. We test this by using the SAE decoder as the initial dictionary for DL-FISTA (unsupervised alternating optimisation), comparing against DL-FISTA initialised from a random dictionary.

\begin{figure*}
  \centering
  \includegraphics[width=0.95\textwidth]{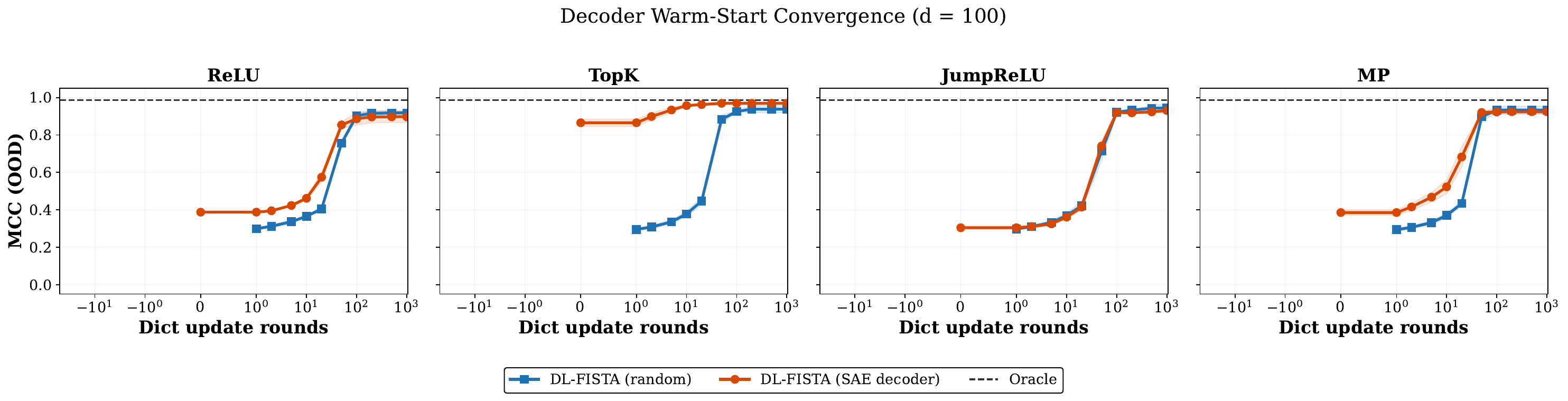}
  \caption{\textbf{The SAE decoder is a useful warm-start for dictionary learning.} Orange: DL-FISTA initialised from SAE decoder. Blue: DL-FISTA from random dictionary. Gray dashed: oracle. The SAE decoder provides a clear head start, especially for TopK, but both initialisations converge to the same optimum. $d_z=5000$, $k=10$. Results at other $d_z$ in appendix.}
  \label{fig:warmstart-decoder}
\end{figure*}

\cref{fig:warmstart-decoder} shows convergence curves at $d_z=100$, sweeping the number of dictionary-update rounds. The SAE decoder provides a clear head start: for TopK, warm-starting begins at OOD MCC $0.87$ and reaches $0.94$ within $5$ rounds, while random initialisation starts at $0.30$ and requires ${\sim}50$ rounds to reach the same level. For ReLU and JumpReLU, the advantage is smaller but consistent---the SAE decoder saves ${\sim}20$--$50$ rounds of dictionary learning. Both initialisations converge to the same final MCC, confirming that the SAE decoder biases the optimisation toward the correct basin without trapping it in a suboptimal one.

\parless{Encoder warm-start is marginal.} We also test whether the SAE encoder's output provides a useful initialisation for per-sample FISTA (with frozen dictionary). Since the Lasso objective is convex, cold-start and warm-start must converge to the same optimum given sufficient iterations. At $100$ iterations, both yield identical MCC. At low iteration budgets ($1$--$10$), warm-starting from SAE codes provides a modest advantage for TopK but negligible benefit for other types (appendix, \cref{fig:warmstart-encoder-appendix}). The practical value of the SAE encoder as a FISTA initialiser is limited.

\begin{tcolorbox}[
  colback=gray!34,
  colframe=gray!34,
  boxrule=0pt,
  arc=0pt,
  left=2pt,right=2pt,top=2pt,bottom=2pt
]
\textbf{Takeaway.}\; At small $d_z$, the SAE decoder is a useful warm-start for DL-FISTA, saving ${\sim}50$ dictionary-update rounds. At large $d_z$ ($\geq 5000$), this advantage vanishes---neither initialisation leads to a good dictionary. The practical path forward requires better dictionary learning algorithms that scale, not just better initialisations.
\end{tcolorbox}

\subsection{Scaling number of latents}
\begin{figure}
    \centering
    \includegraphics[width=\linewidth]{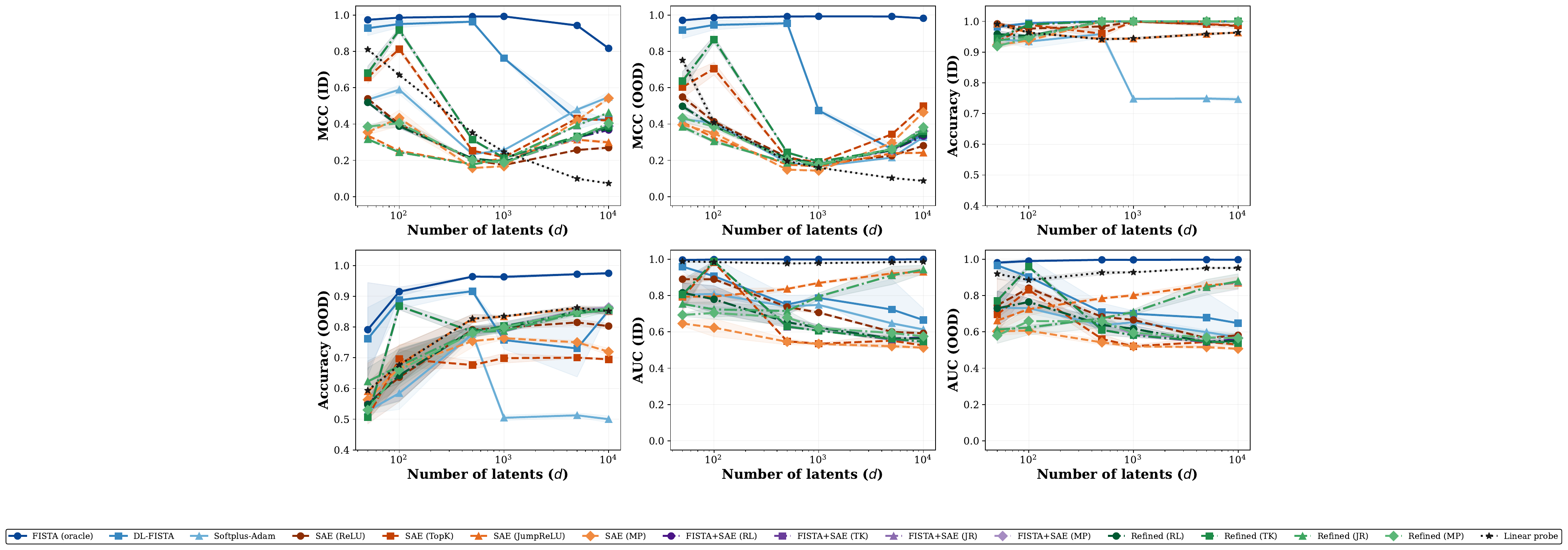}
    \caption{\textbf{All metrics vs number of latent variables.} Per-sample methods (blue) degrade uniformly across ID and OOD as $d_z$ grows. SAEs (orange, dashed) show a persistent ID--OOD gap across all metrics. Frozen-decoder and refined hybrids (purple, green) close part of the gap by swapping inference. $k=10$, $p=5000$.}
    \label{fig:vary_frozen_appendix}
\end{figure}

\cref{fig:vary_frozen_appendix} extends the main-text \cref{fig:vary-latents,fig:frozen-faceted} with all six metric panels (MCC and AUC on both ID and OOD, plus accuracy). The key patterns from the main text hold across all metrics: per-sample methods maintain a small ID--OOD gap while SAEs exhibit a persistent and large gap. The frozen-decoder and refined variants consistently improve over raw SAEs, with gains most pronounced on OOD metrics.

\begin{figure}
    \centering
    \includegraphics[width=0.5\linewidth]{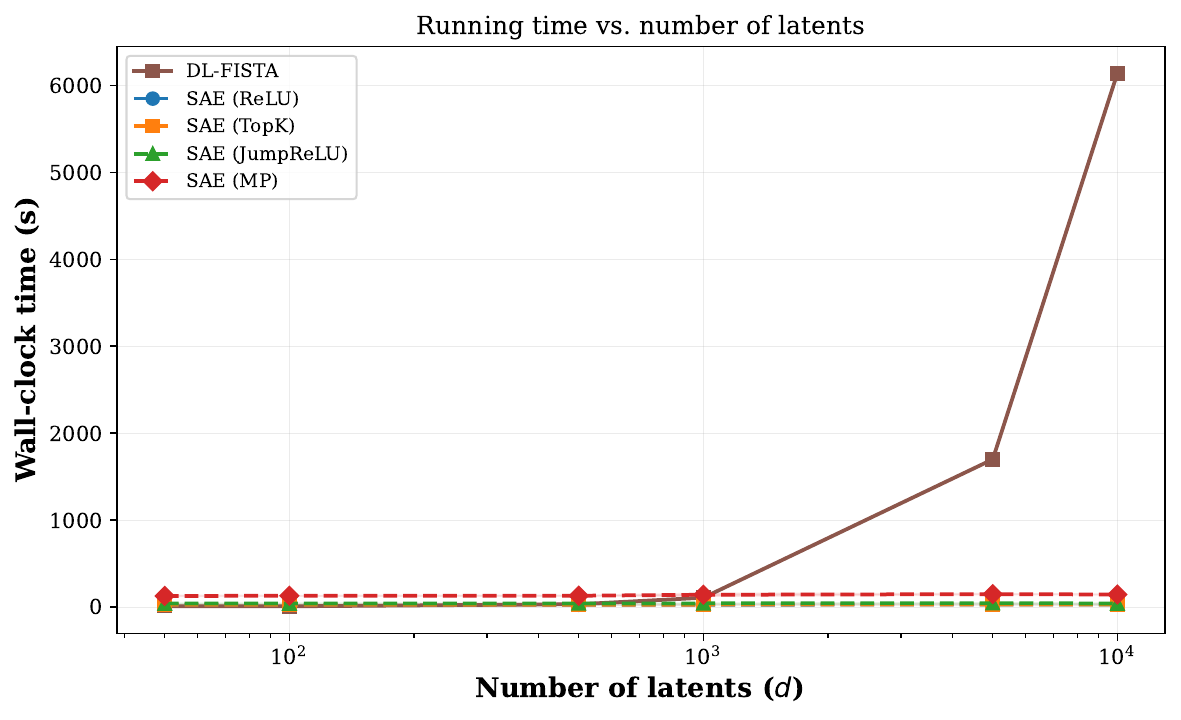}
    \caption{Running time X number of latent variables. The runtime increases significantly for over $10^4$ latents.}
    \label{fig:exp_vary_latents_runtime}
\end{figure}

Figure \ref{fig:exp_vary_latents_runtime} explores running time vs number of latents for DL-FISTA (unsupervised sparse coding) and SAEs. We find that DL-FISTA (running for 200 iterations) has a comparable running time to SAEs for up to 1000 latents, after which the running time increases up to 6000s. In practice, this was not a limitation when applying this experiment to real data, and it can be made more efficient by reducing the number of iterations and improving parallelization.


\subsection{More data does not close the gap}
\label{apx:more-data}

We vary the number of training samples $p \in \{10^2, \ldots, 10^5\}$ with all other parameters held fixed and report ID MCC (\cref{fig:vary-samples}) and OOD accuracy (\cref{fig:vary-samples-acc}).
FISTA (oracle) is constant by construction (it uses no
training data beyond the oracle dictionary). DL-FISTA benefits substantially from more data---its MCC jumps from ${\sim}0.5$ at $p = 10^2$ to ${\sim}0.98$ by $p = 10^3$ and saturates---confirming that dictionary learning is genuinely sample-limited and that per-sample inference learns a better dictionary with enough data. SAE variants show a different pattern.  SAE (TopK) and SAE (ReLU) improve only marginally, plateauing around
$0.35$--$0.45$ MCC.  SAE (JumpReLU) \emph{degrades} with
more data, dropping from ${\sim}0.45$ to below $0.1$, suggesting
that additional training leads to a poor local solution rather than
correcting it.  The gap between per-sample and amortised methods is
stable or widening across two orders of magnitude of additional data. On the downstream prediction task (\cref{fig:vary-samples-acc}), DL-FISTA clearly separates from the linear probe once $p \geq 10^3$ ($0.88$ vs $0.68$ accuracy), while all SAE variants trail the linear probe---this is the regime ($d_z=100$) where dictionary learning works, and it translates directly into better OOD generalisation. OOD AUC shows the same pattern (appendix, \cref{fig:vary_samples_auc_ood_v2})

\begin{figure}
    \centering
    \includegraphics[width=\linewidth]{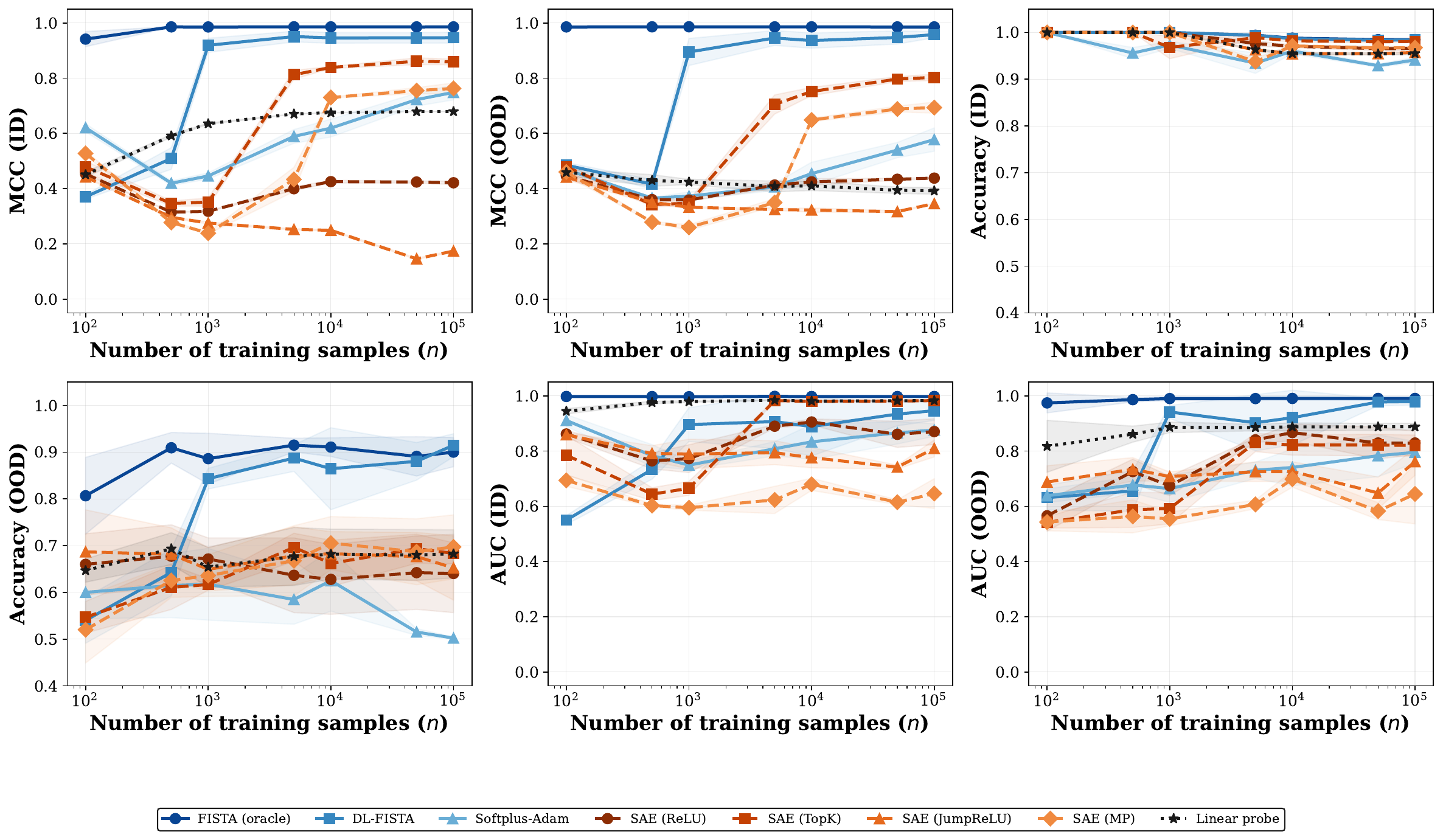}
    \caption{\textbf{All metrics vs number of training samples.} DL-FISTA's MCC and AUC improve sharply with more data and saturate by $p=10^3$. SAE variants plateau around $0.35$--$0.45$ MCC or degrade (JumpReLU), and OOD AUC remains scattered between $0.5$--$0.85$ with high variance regardless of $p$. $d_z=100$, $k=10$, $d_y=47$.}
    \label{fig:vary_samples_appendix}
\end{figure}

\cref{fig:vary_samples_appendix} extends \cref{fig:vary-samples} with all six metrics. DL-FISTA benefits substantially from more data across all metrics, while SAEs plateau or degrade---confirming that the amortisation gap is not a sample-complexity issue.

\subsection{Varying sparsity level}
\begin{figure}
    \centering
    \includegraphics[width=\linewidth]{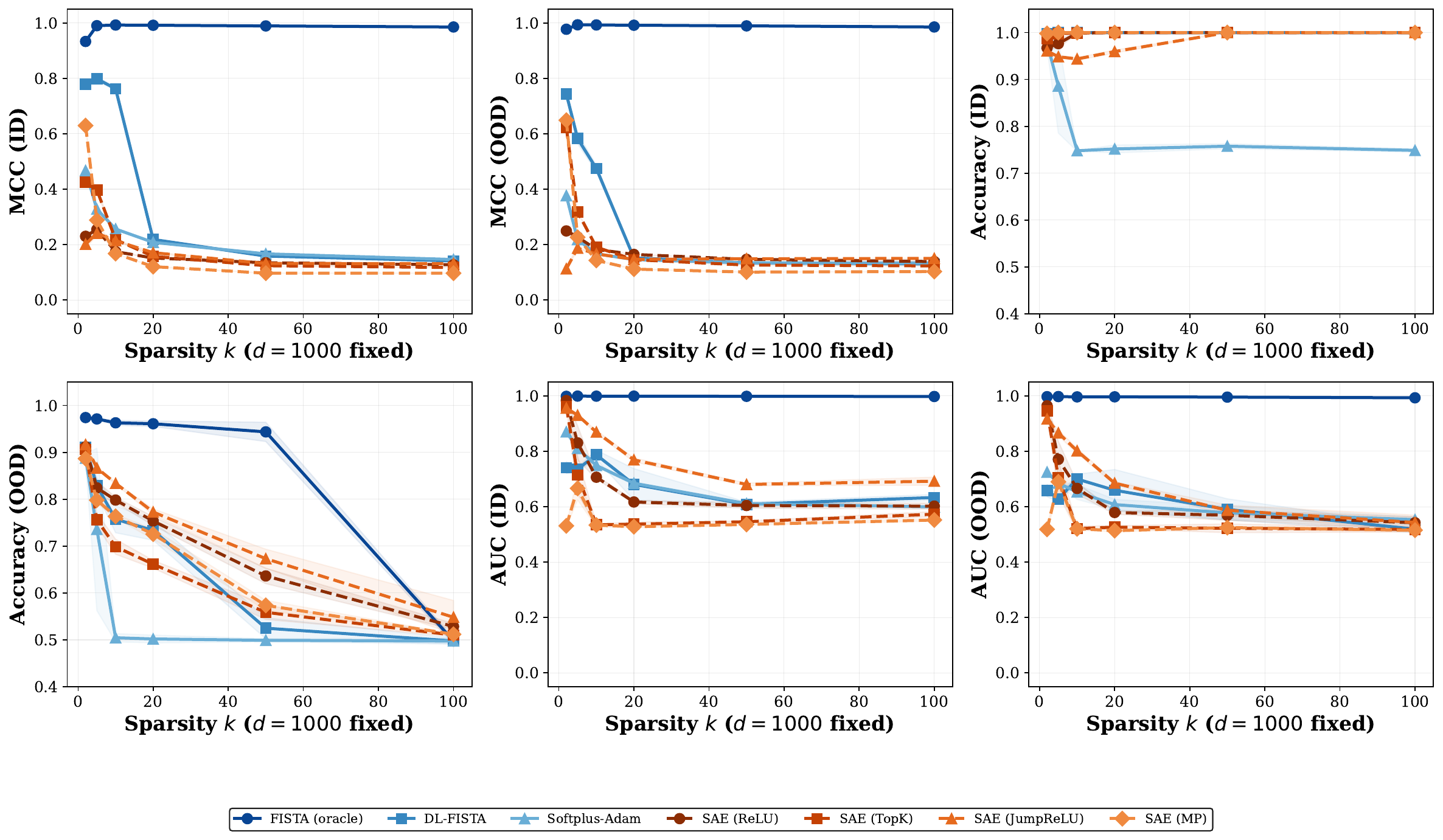}
    \caption{\textbf{All metrics vs sparsity $k$.} Per-sample methods degrade gracefully with increasing $k$; SAE OOD AUC converges toward chance at high $k$. $d_z=1000$, $d_y$ follows the CS bound.}
    \label{fig:vary_sparsity_appendix}
\end{figure}

\cref{fig:vary_sparsity_appendix} sweeps sparsity $k$ with $d_z=1000$ fixed. Per-sample methods degrade gracefully as sparsity increases (the inference problem becomes harder), maintaining consistent ID--OOD performance. SAE OOD AUC converges toward chance (${\sim}0.5$) at high $k$, further confirming the compositional generalisation failure.

\subsection{Additional main-text metrics (v2 figures)}

\parless{AUC (OOD).} AUC evaluates single-feature separability without a trained classifier. Note that the linear probe's AUC is computed on its supervised regression output---each output dimension targets one latent---giving it an inherent advantage over unsupervised methods whose code dimensions need not align with the label. AUC can therefore overstate the linear probe's generalisation capability relative to unsupervised methods; the accuracy metric (\cref{fig:vary-latents-acc,fig:vary-samples-acc}) provides the fairer comparison.

\begin{figure}
    \centering
    \includegraphics[width=\linewidth]{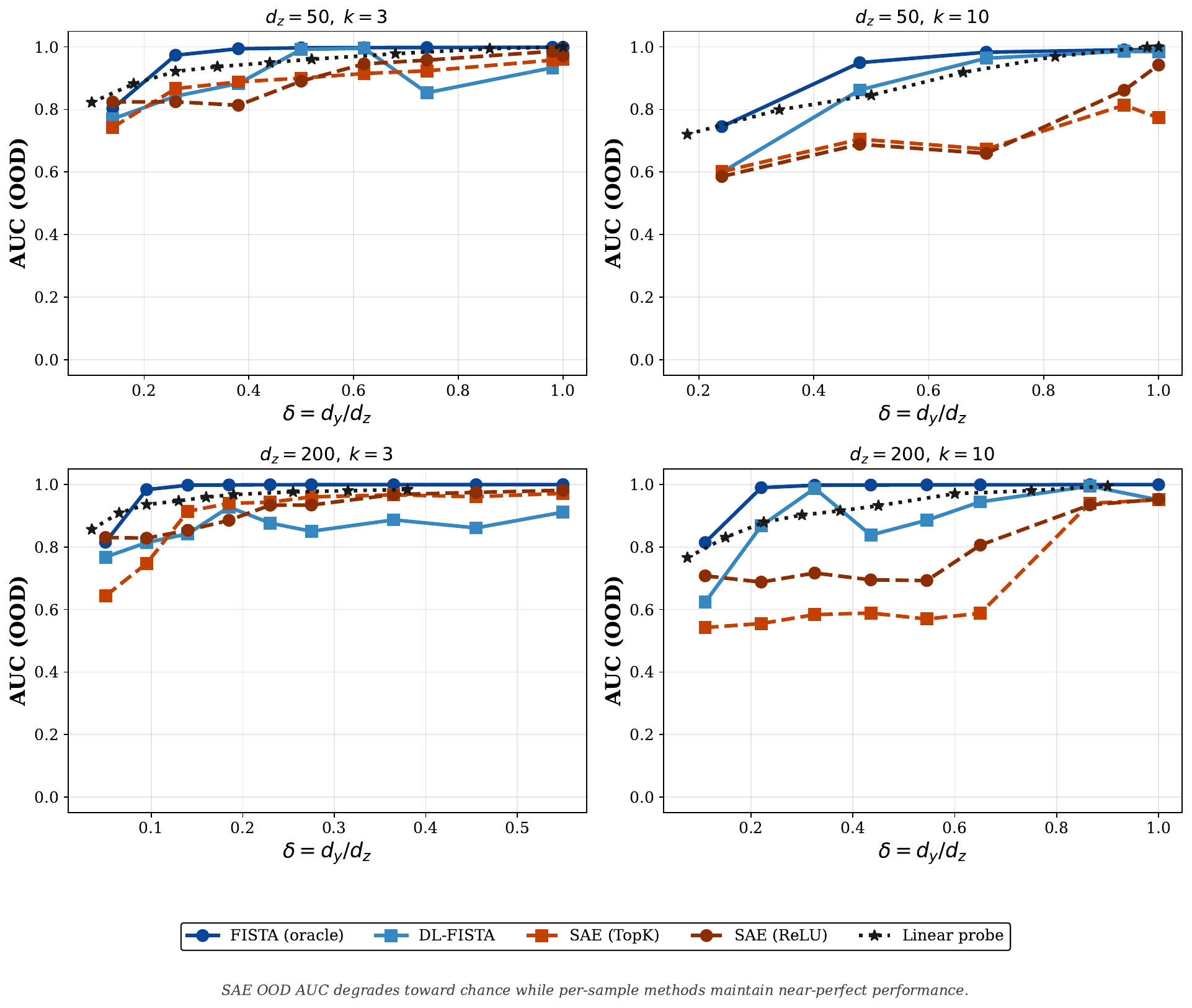}
    \caption{\textbf{Phase transition: AUC (OOD).} SAE OOD AUC degrades toward chance while per-sample methods maintain near-perfect performance across all $(d_z, k)$ settings.}
    \label{fig:phase_auc_ood_v2}
\end{figure}

\begin{figure}
    \centering
    \includegraphics[width=\linewidth]{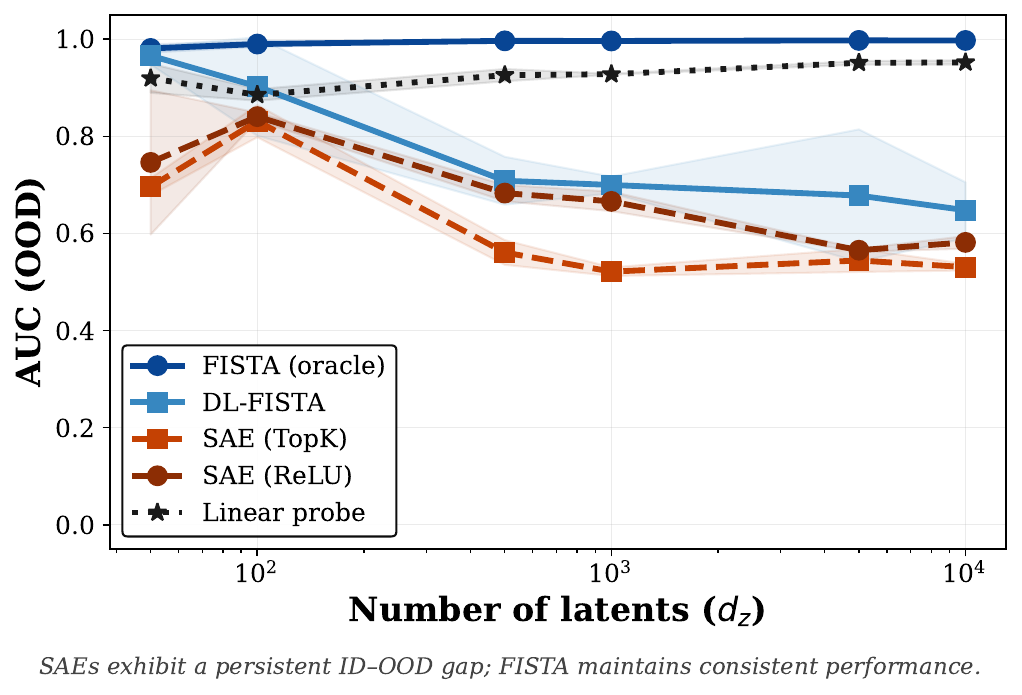}
    \caption{\textbf{Scaling latent dimension: AUC (OOD).} The linear probe maintains high AUC (${\sim}0.93$) even as its MCC collapses to $0.07$ (\cref{fig:vary-latents}), illustrating how a supervised single-label metric can mask identifiability failure. FISTA (oracle) dominates; SAEs degrade. $k=10$, $p=5000$.}
    \label{fig:vary_latents_auc_ood_v2}
\end{figure}

\begin{figure}
    \centering
    \includegraphics[width=\linewidth]{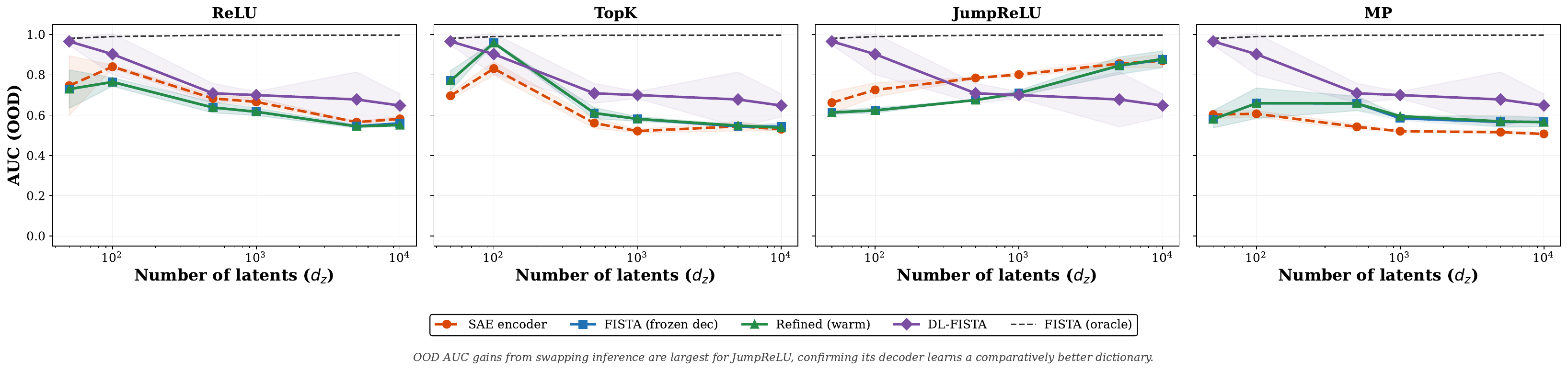}
    \caption{\textbf{Frozen decoder ablation: AUC (OOD).} Same layout as \cref{fig:frozen-faceted}. FISTA on frozen TopK and JumpReLU decoders yields modest OOD AUC gains, but the gap to DL-FISTA remains large for all types. $k=10$, $p=5000$.}
    \label{fig:frozen_faceted_auc_ood_v2}
\end{figure}

\begin{figure}
    \centering
    \includegraphics[width=\linewidth]{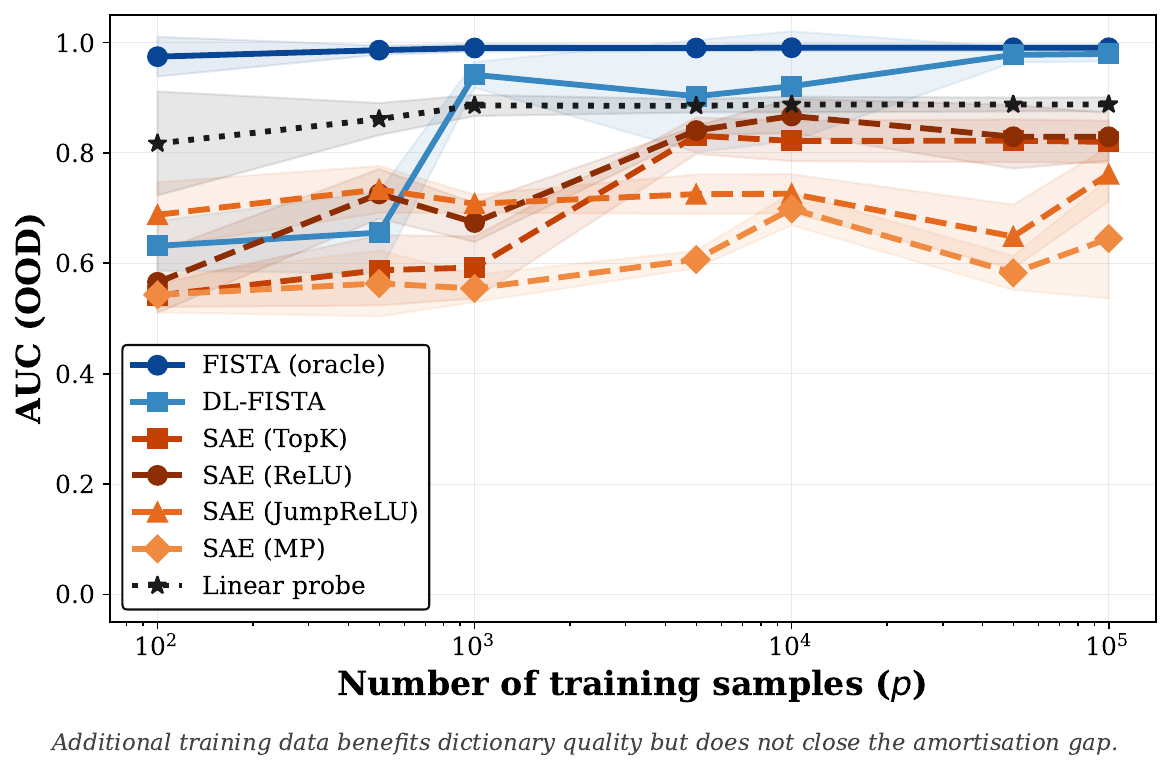}
    \caption{\textbf{More data: AUC (OOD).} Additional training data benefits DL-FISTA but does not close the amortisation gap on OOD AUC. $d_z=100$, $k=10$, $d_y=47$.}
    \label{fig:vary_samples_auc_ood_v2}
\end{figure}

\begin{figure}
    \centering
    \includegraphics[width=\linewidth]{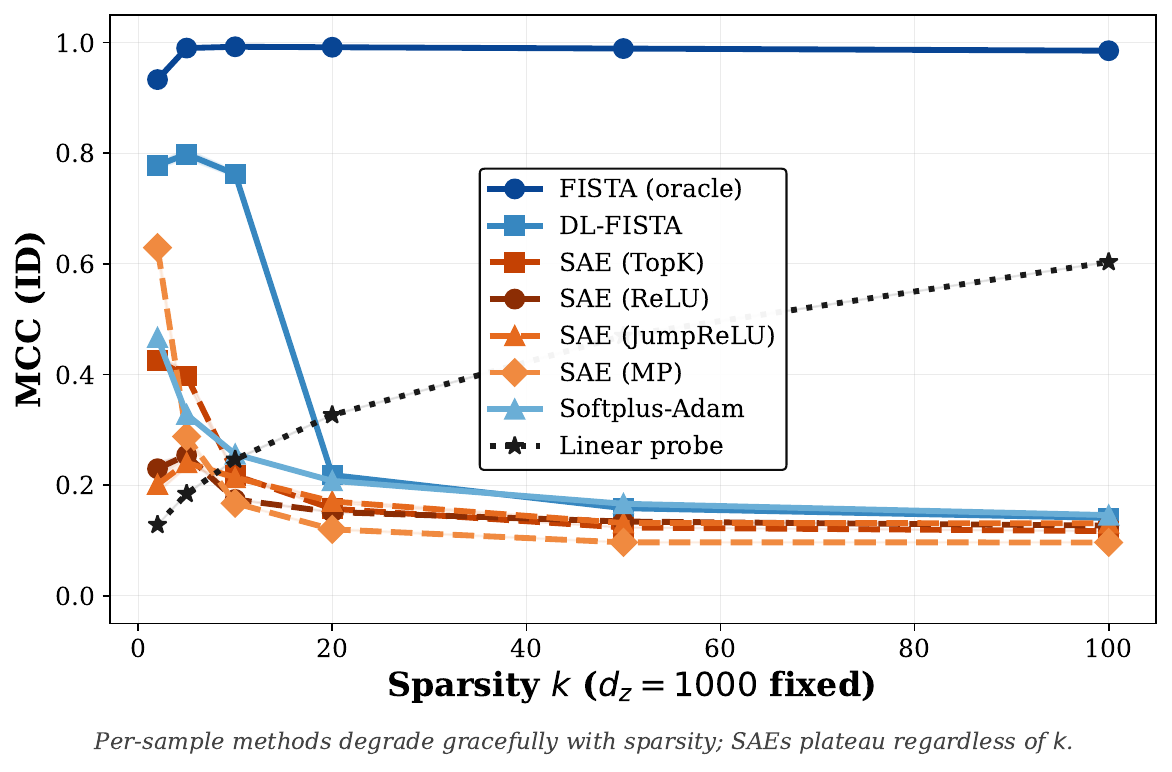}
    \caption{\textbf{Varying sparsity: MCC (ID).} Per-sample methods degrade gracefully with increasing $k$; SAEs plateau. $d_z=1000$, $d_y$ follows the CS bound.}
    \label{fig:vary_sparsity_mcc_id_v2}
\end{figure}

\begin{figure}
    \centering
    \includegraphics[width=\linewidth]{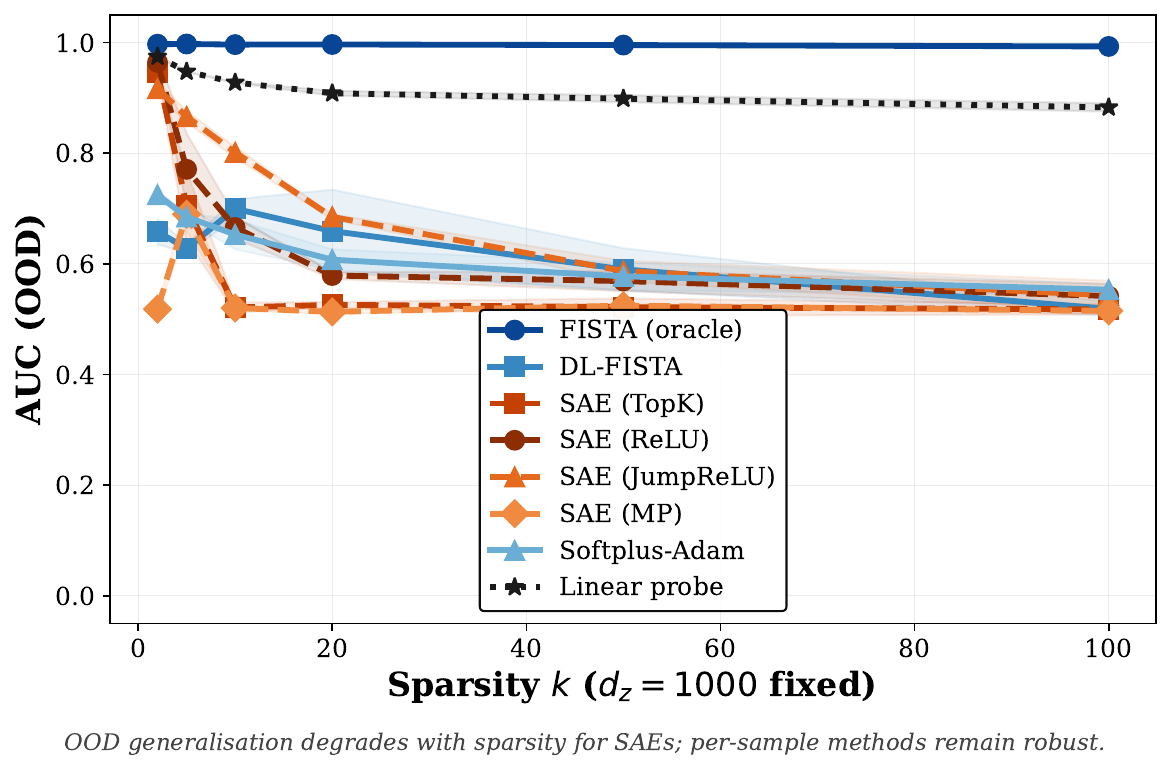}
    \caption{\textbf{Varying sparsity: AUC (OOD).} FISTA (oracle) remains near $1.0$; all other methods degrade with increasing $k$, with SAEs and DL-FISTA converging toward $0.5$. The linear probe degrades more gracefully ($0.97 \to 0.88$) due to its supervised advantage. $d_z=1000$.}
    \label{fig:vary_sparsity_auc_ood_v2}
\end{figure}

\begin{figure}
    \centering
    \includegraphics[width=\linewidth]{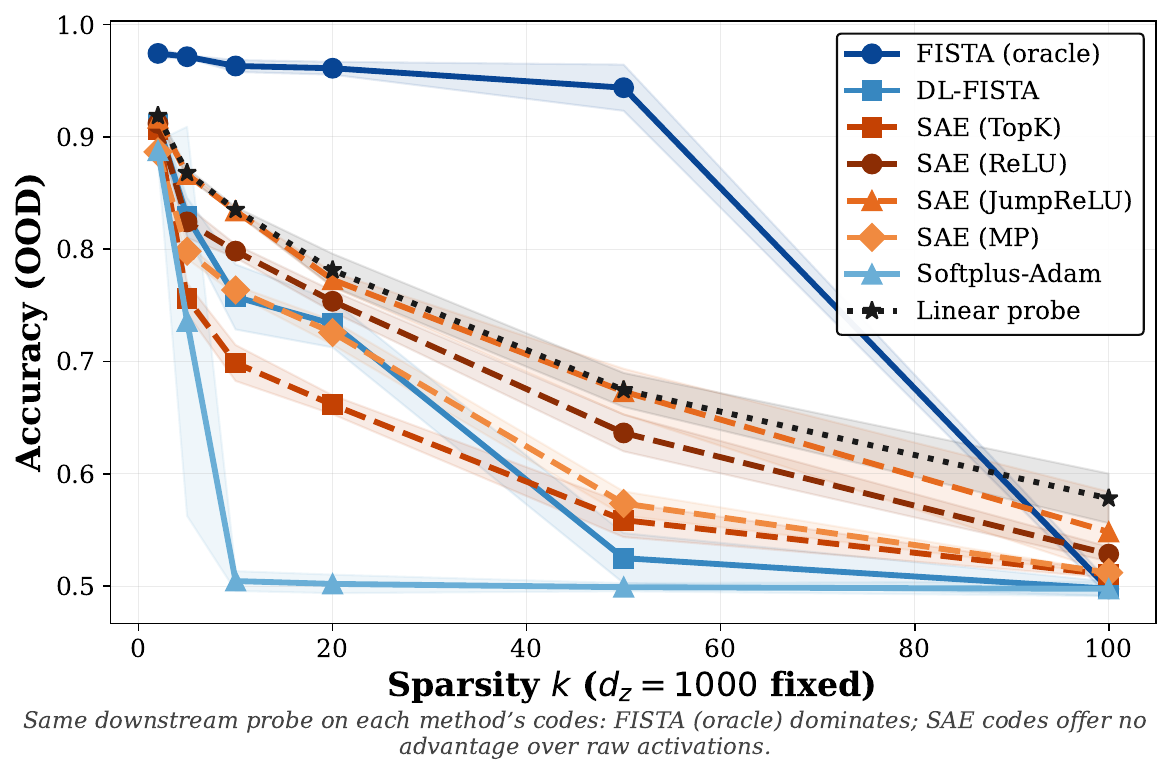}
    \caption{\textbf{Varying sparsity: Accuracy (OOD).} Same downstream probe on each method's codes. FISTA (oracle) dominates. DL-FISTA beats the linear probe at low $k$ ($\leq 10$) but collapses at high $k$ as dictionary learning fails. SAE codes offer no consistent advantage over raw activations. $d_z=1000$.}
    \label{fig:vary_sparsity_acc_ood_v2}
\end{figure}

\begin{figure}
    \centering
    \includegraphics[width=\linewidth]{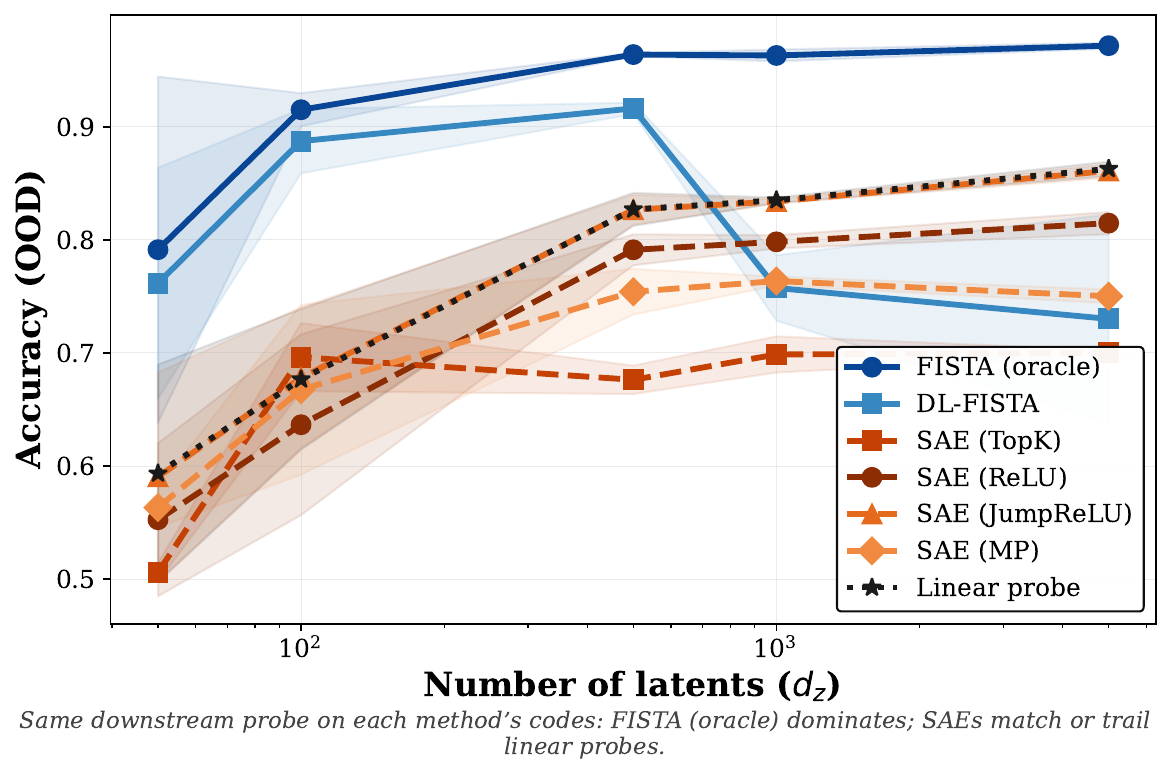}
    \caption{\textbf{Varying latent dimension: Accuracy (OOD).} Duplicate of \cref{fig:vary-latents-acc} for completeness alongside other appendix metrics.}
    \label{fig:vary_latents_acc_ood_v2}
\end{figure}

\begin{figure}
    \centering
    \includegraphics[width=\linewidth]{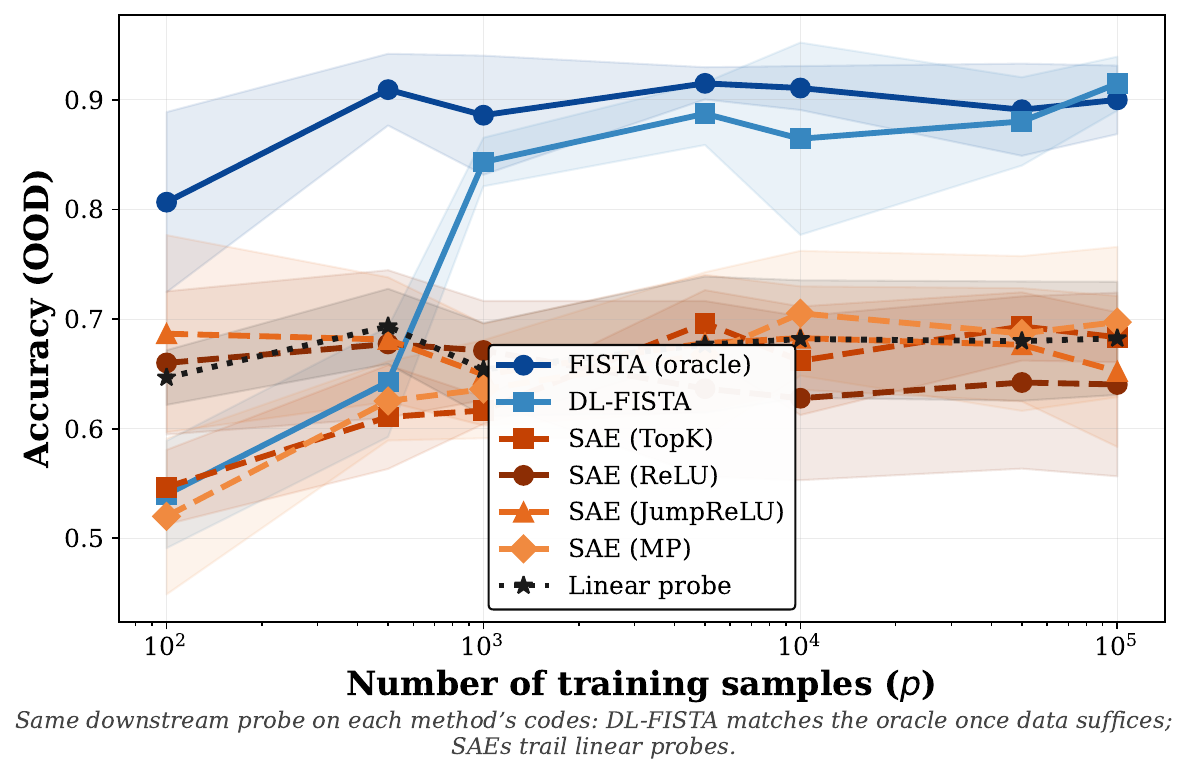}
    \caption{\textbf{Varying training data: Accuracy (OOD).} Duplicate of \cref{fig:vary-samples-acc} for completeness alongside other appendix metrics.}
    \label{fig:vary_samples_acc_ood_v2}
\end{figure}

\subsection{Controlled experiment details}

\begin{figure}
    \centering
    \includegraphics[width=\linewidth]{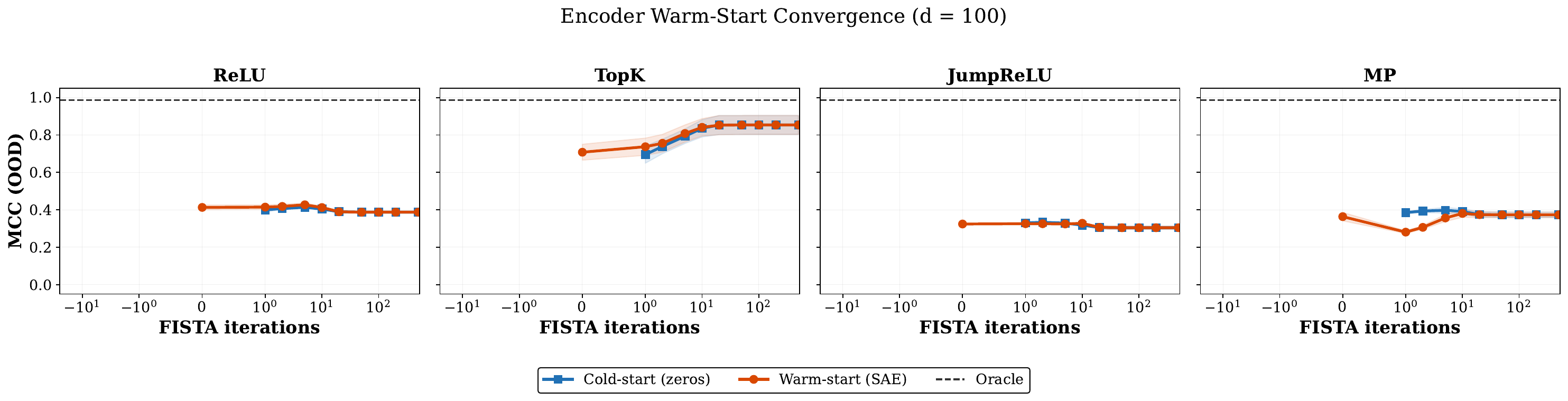}
    \caption{\textbf{Encoder warm-start convergence ($d_z=100$).} Cold-start (blue) and warm-start (orange) FISTA on frozen SAE decoder. The convex Lasso objective means both converge to the same optimum. Warm-starting provides a modest advantage for TopK at low iteration budgets but negligible benefit for other types.}
    \label{fig:warmstart-encoder-appendix}
\end{figure}

\begin{figure}
    \centering
    \includegraphics[width=\linewidth]{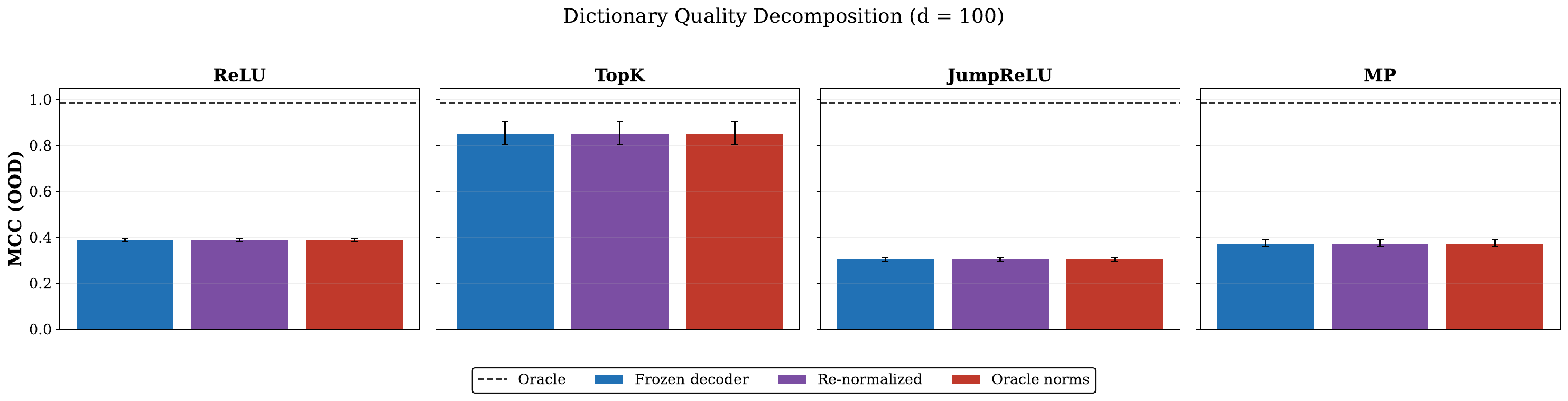}
    \caption{\textbf{Dictionary quality decomposition ($d_z=100$).} Blue: FISTA with frozen SAE decoder. Purple: after re-normalising decoder columns to unit norm. Red: SAE directions with oracle column magnitudes. Re-normalising and norm substitution have no effect---the error is in the column \emph{directions}, not magnitudes.}
    \label{fig:dict-quality-appendix}
\end{figure}

\begin{figure}
    \centering
    \includegraphics[width=\linewidth]{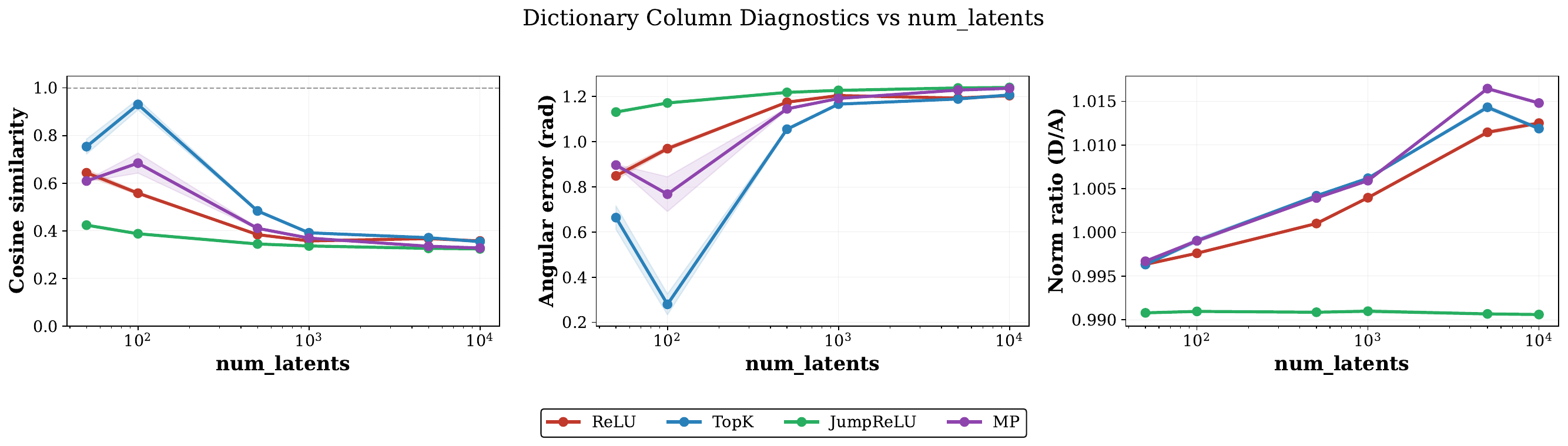}
    \caption{\textbf{Dictionary column diagnostics across $d_z$.} Left: mean cosine similarity between SAE and ground-truth columns. Centre: angular error. Right: norm ratio. TopK maintains high cosine ($>0.9$) across all $d_z$; other types degrade. Norm ratios are ${\approx}1.0$ throughout, confirming the error is directional.}
    \label{fig:dict-diagnostics-appendix}
\end{figure}

\begin{figure}
    \centering
    \includegraphics[width=\linewidth]{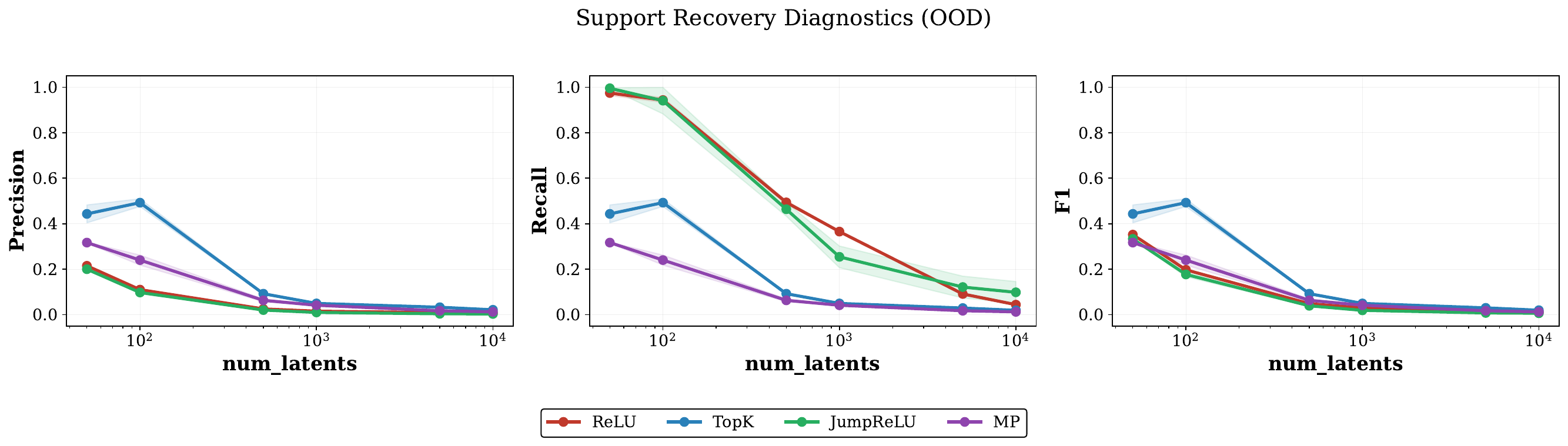}
    \caption{\textbf{Support recovery diagnostics across $d_z$.} Precision, recall, and F1 of the SAE's binary support compared to ground truth. ReLU and JumpReLU have high recall but catastrophically low precision (${\sim}0.1$): they activate nearly all features. TopK maintains balanced precision and recall (${\sim}0.5$).}
    \label{fig:support-diagnostics-appendix}
\end{figure}

\subsection{Lambda sensitivity}
\label{app:lambda}

\begin{figure}
    \centering
    \includegraphics[width=\linewidth]{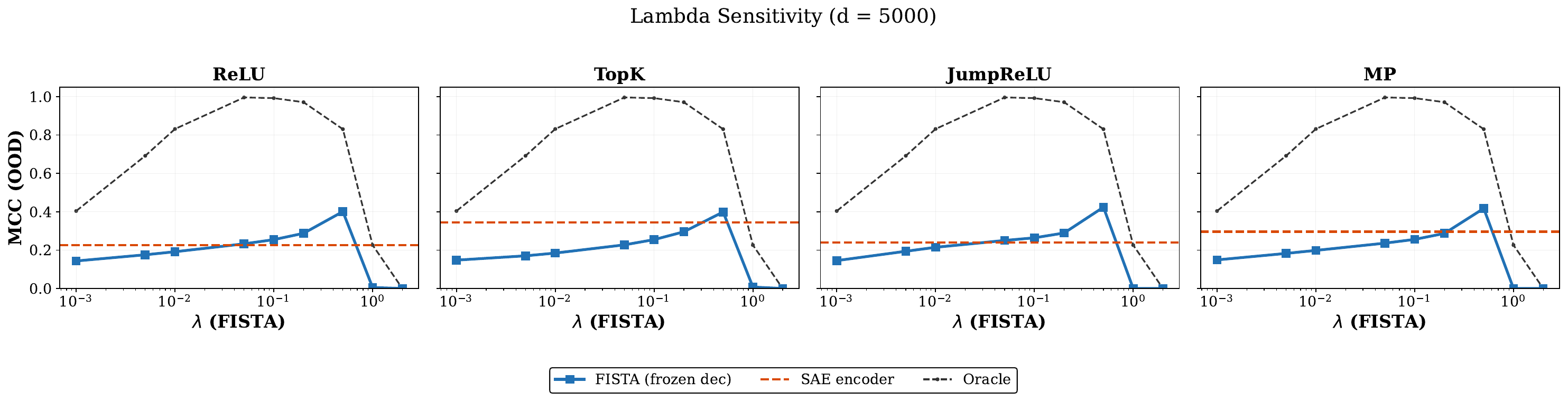}
    \caption{\textbf{Lambda sensitivity ($d_z=5000$).} Blue: FISTA on frozen SAE decoder across $\lambda$ values. Orange dashed: SAE encoder (lambda-independent). Gray dashed: oracle FISTA. The frozen decoder peaks modestly above the SAE at $\lambda \approx 0.5$, but the gap to oracle persists at every $\lambda$.}
    \label{fig:lambda-sensitivity}
\end{figure}

The frozen decoder experiments use $\lambda=0.1$ for FISTA, while SAE training uses $\gamma_\text{reg}=10^{-4}$---a $1000\times$ mismatch. To verify that the dictionary quality conclusion is not an artefact of this mismatch, we sweep $\lambda \in \{10^{-3}, \ldots, 2.0\}$ for both the frozen decoder and oracle conditions (\cref{fig:lambda-sensitivity}).

The oracle achieves peak MCC ${\sim}0.95$ at $\lambda \approx 0.1$--$0.5$, confirming the correct operating regime. The frozen decoder peaks at $\lambda \approx 0.5$ for most SAE types, modestly exceeding the SAE encoder baseline for ReLU and JumpReLU (e.g., $0.4$ vs $0.25$ at $d_z=5000$). However, the gap between frozen FISTA and oracle remains large at every $\lambda$, confirming that the dictionary---not the regularisation strength---is the bottleneck.

\begin{figure}
    \centering
    \includegraphics[width=\linewidth]{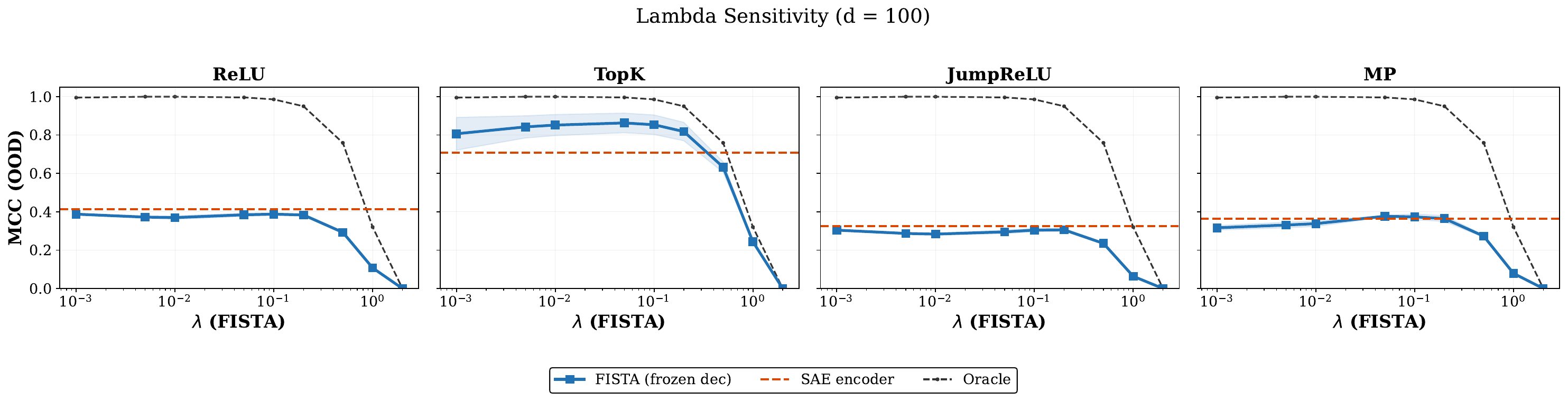}
    \caption{Lambda sensitivity, $d_z=100$.}
\end{figure}

\subsection{Learning dynamics at other latent dimensions}

\begin{figure}
    \centering
    \includegraphics[width=\linewidth]{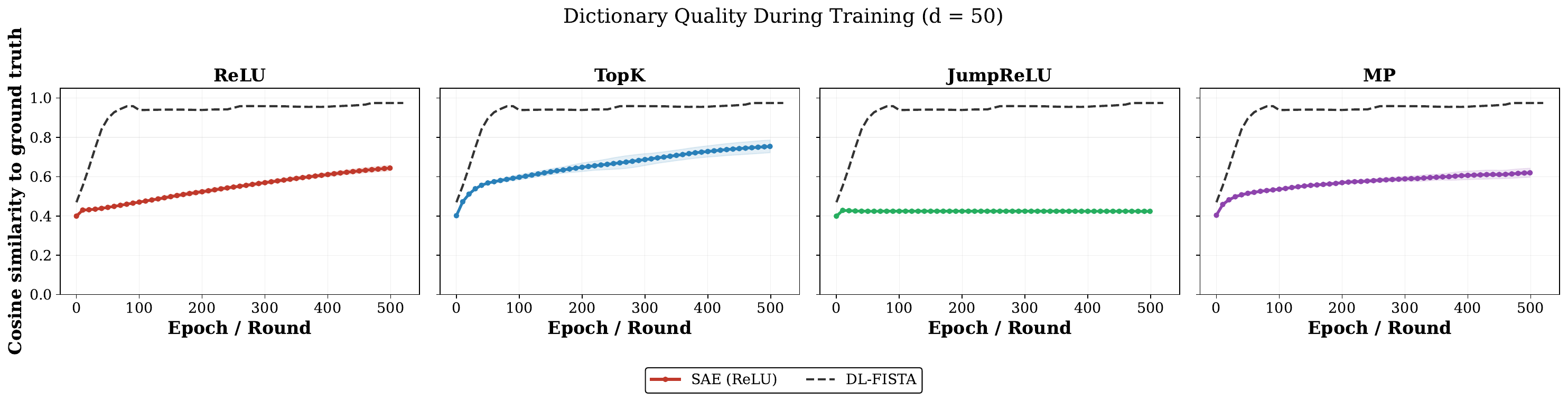}
    \caption{Dictionary quality during training, $d_z=50$.}
\end{figure}

\begin{figure}
    \centering
    \includegraphics[width=\linewidth]{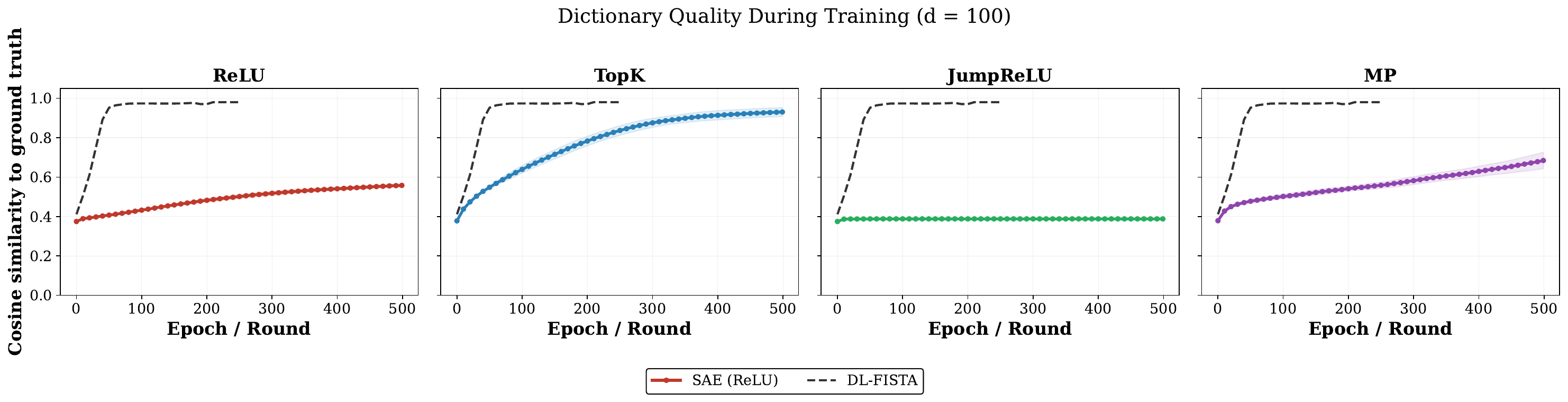}
    \caption{Dictionary quality during training, $d_z=100$. At this scale, DL-FISTA converges to high cosine while SAEs plateau at lower values.}
\end{figure}

\begin{figure}
    \centering
    \includegraphics[width=\linewidth]{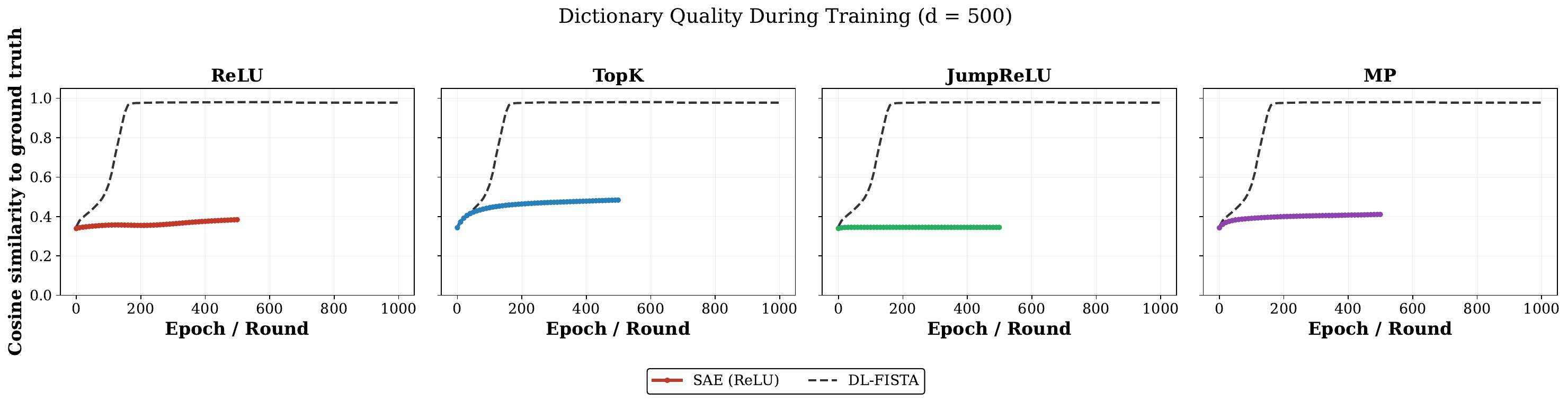}
    \caption{Dictionary quality during training, $d_z=500$.}
\end{figure}

\begin{figure}
    \centering
    \includegraphics[width=\linewidth]{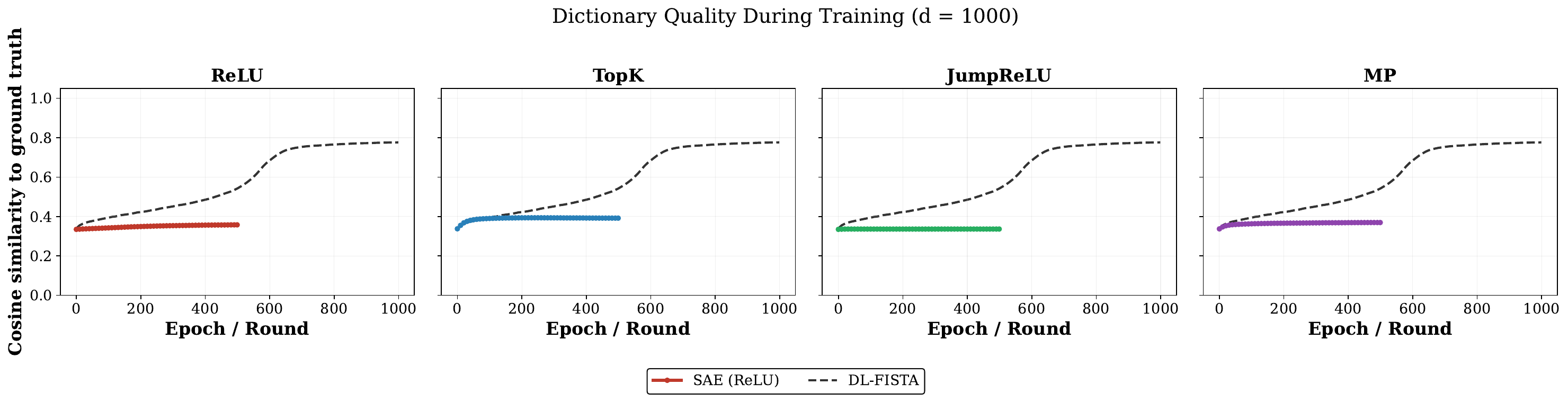}
    \caption{Dictionary quality during training, $d_z=1000$.}
\end{figure}

\begin{figure}
    \centering
    \includegraphics[width=\linewidth]{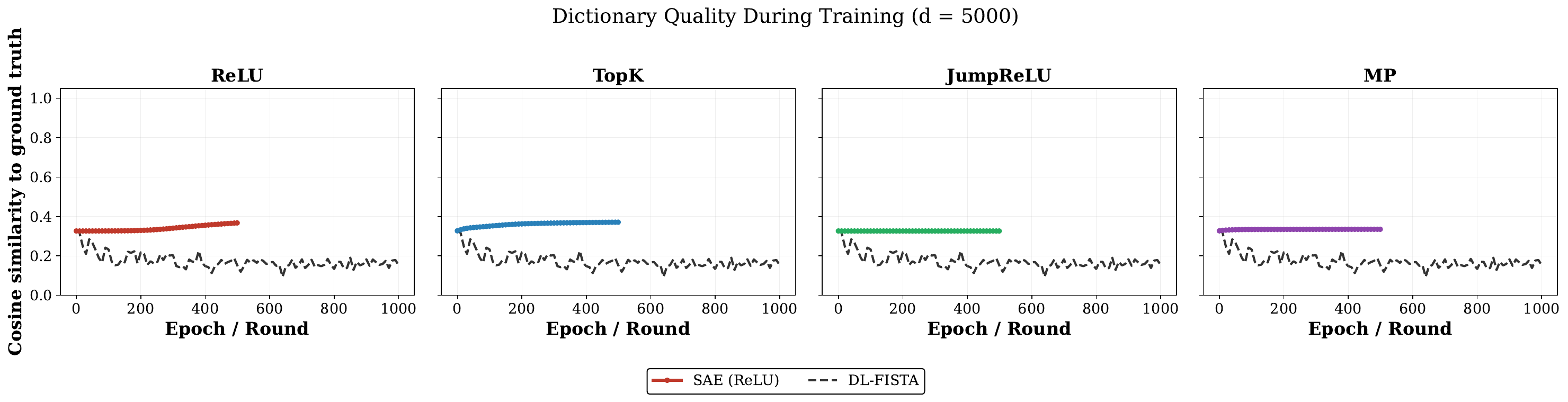}
    \caption{Dictionary quality during training, $d_z=5000$. Neither SAEs nor DL-FISTA converge to high cosine at this scale.}
    \label{fig:learning-dynamics-n5000}
\end{figure}

\begin{figure}
    \centering
    \includegraphics[width=\linewidth]{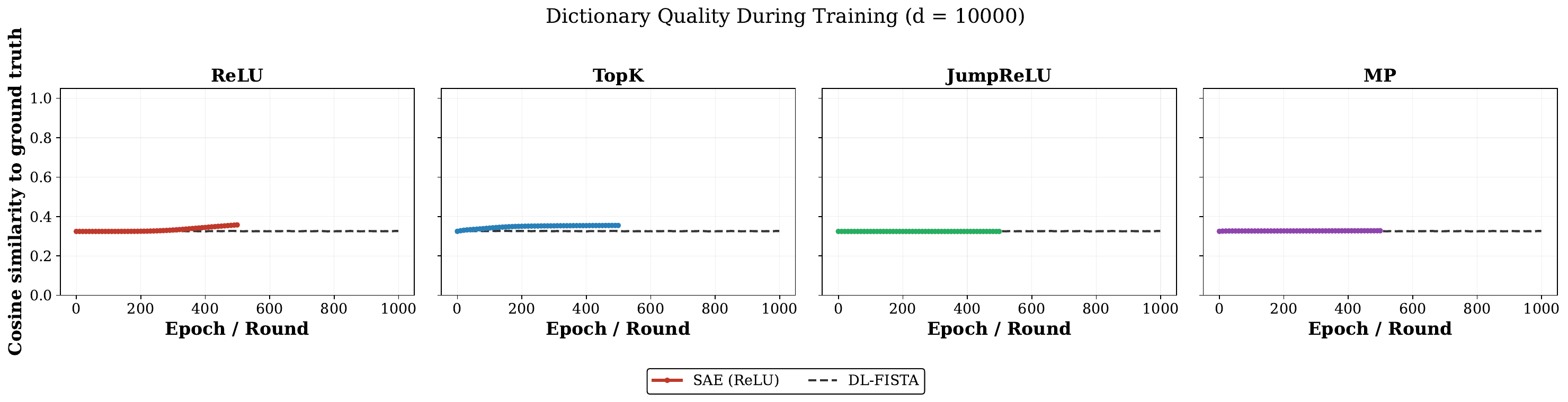}
    \caption{Dictionary quality during training, $d_z=10000$.}
\end{figure}

\subsubsection{Controlled experiments at other latent dimensions}

The main text reports controlled experiments at $d_z=100$. Below we show the same experiments at $d_z \in \{50, 500, 1000, 5000\}$ to confirm the findings hold across scales.

\parless{Decoder warm-start convergence.}
The SAE decoder provides a consistent head start for DL-FISTA across all $d_z$. The advantage is largest at small $d_z$ (where the SAE decoder is closer to the true dictionary) and diminishes at large $d_z$ (where both initialisations require many rounds to converge).

\begin{figure}[h]
    \centering
    \includegraphics[width=\linewidth]{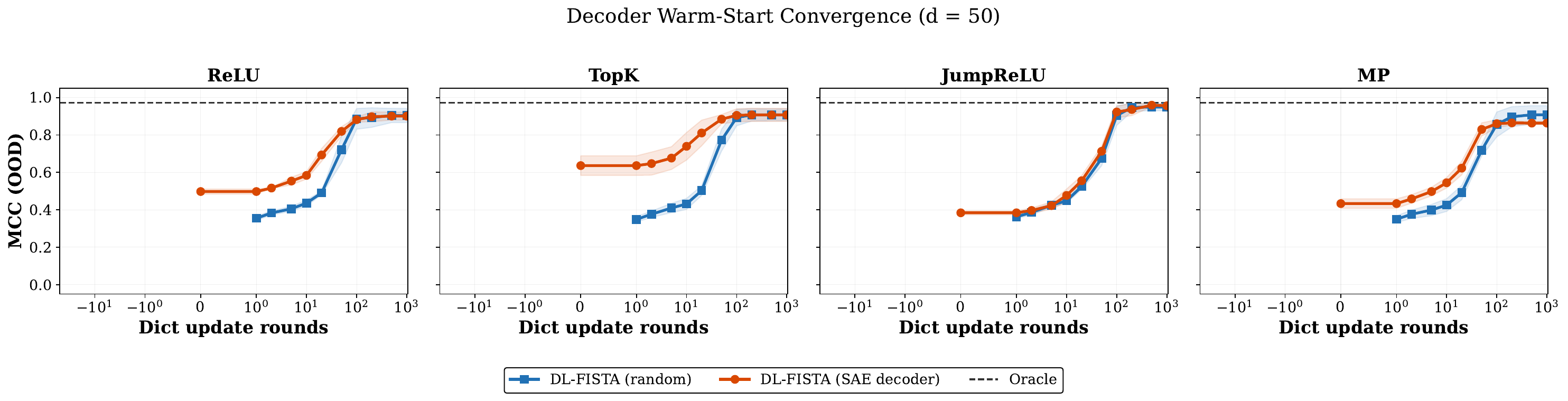}
    \caption{Decoder warm-start convergence, $d_z=50$.}
\end{figure}
\begin{figure}[h]
    \centering
    \includegraphics[width=\linewidth]{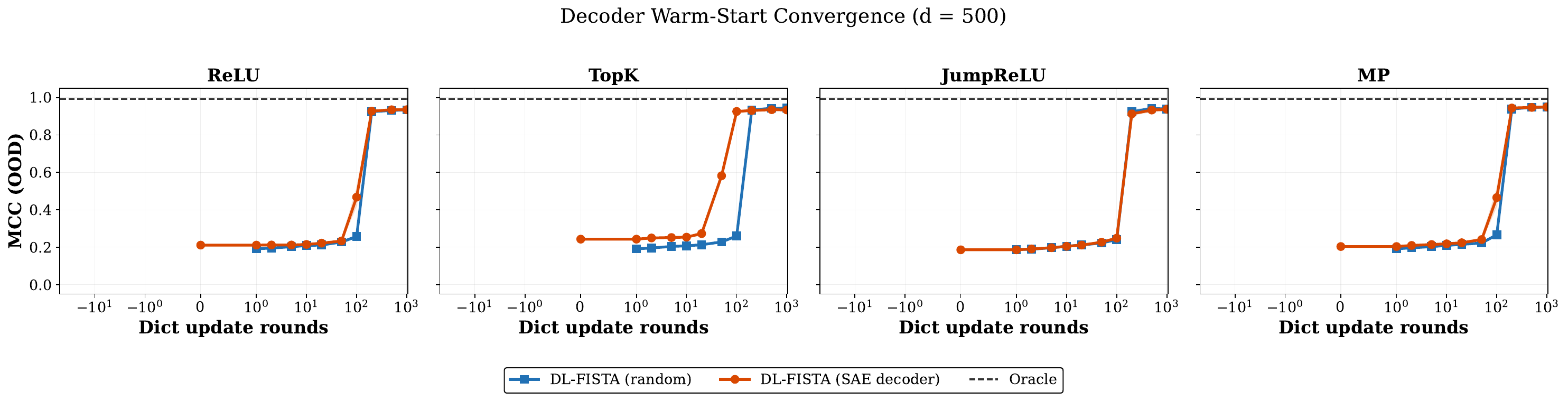}
    \caption{Decoder warm-start convergence, $d_z=500$.}
\end{figure}
\begin{figure}[h]
    \centering
    \includegraphics[width=\linewidth]{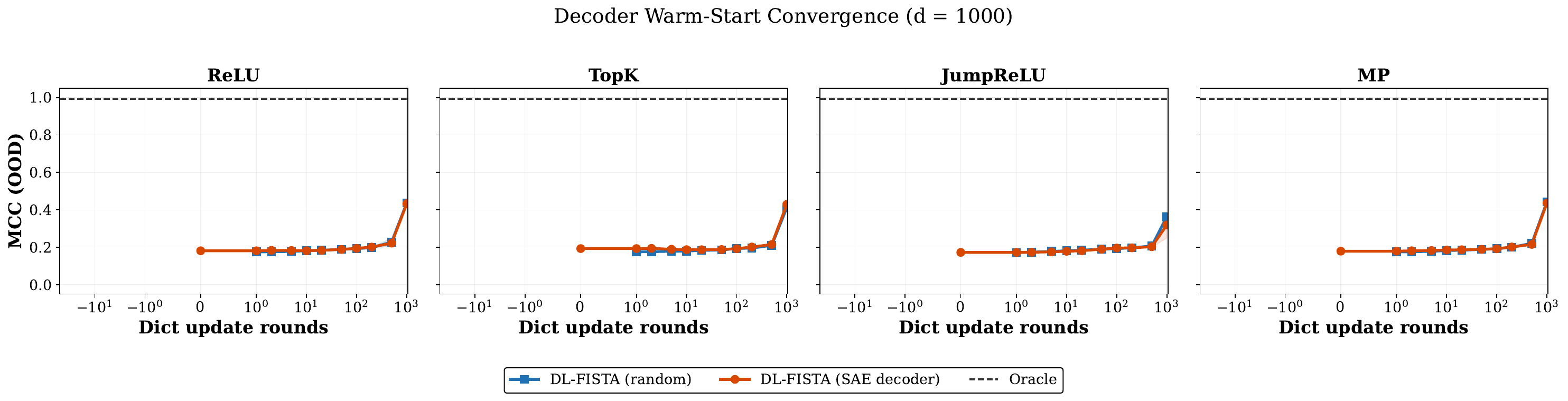}
    \caption{Decoder warm-start convergence, $d_z=1000$.}
\end{figure}
\begin{figure}[h]
    \centering
    \includegraphics[width=\linewidth]{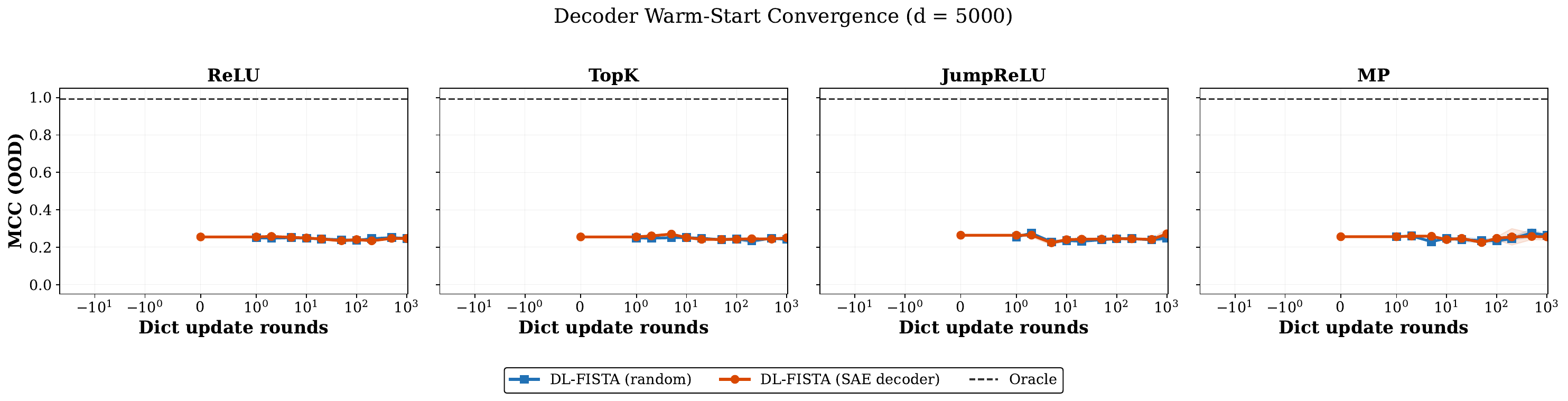}
    \caption{Decoder warm-start convergence, $d_z=5000$.}
\end{figure}
\begin{figure}[h]
    \centering
    \includegraphics[width=\linewidth]{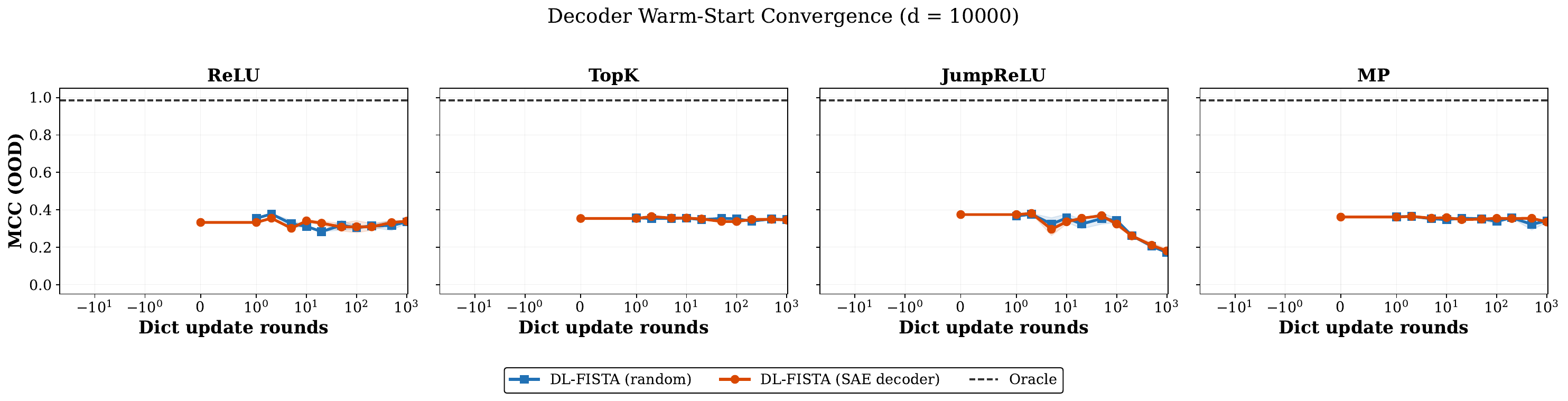}
    \caption{Decoder warm-start convergence, $d_z=10000$.}
\end{figure}

\parless{Encoder warm-start convergence.}
The convex-convergence pattern holds across all $d_z$: cold-start and warm-start reach the same optimum, with warm-starting providing a modest advantage only at very low iteration counts.

\begin{figure}[h]
    \centering
    \includegraphics[width=\linewidth]{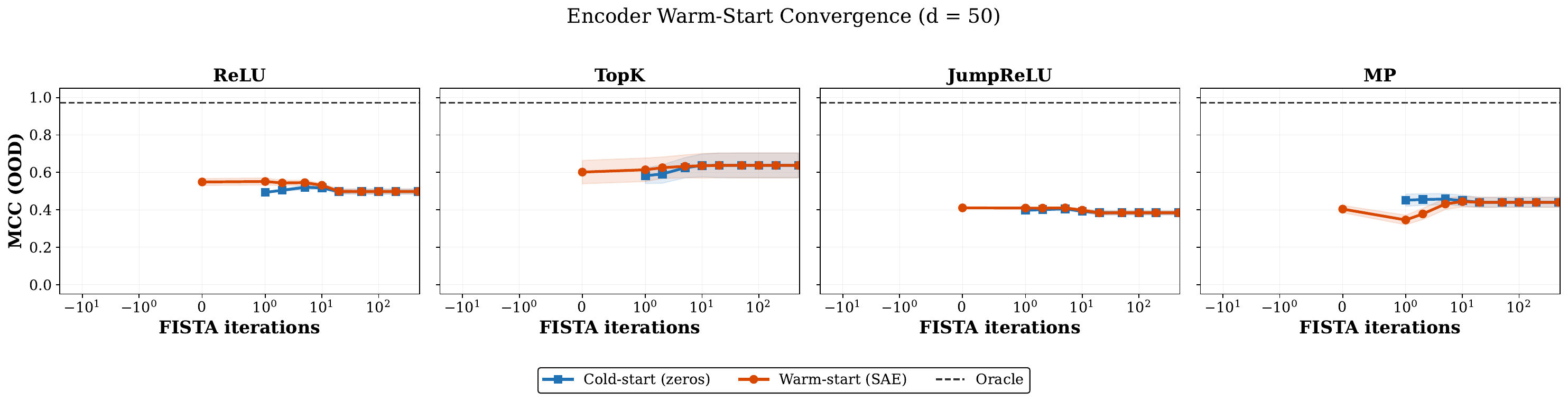}
    \caption{Encoder warm-start convergence, $d_z=50$.}
\end{figure}
\begin{figure}[h]
    \centering
    \includegraphics[width=\linewidth]{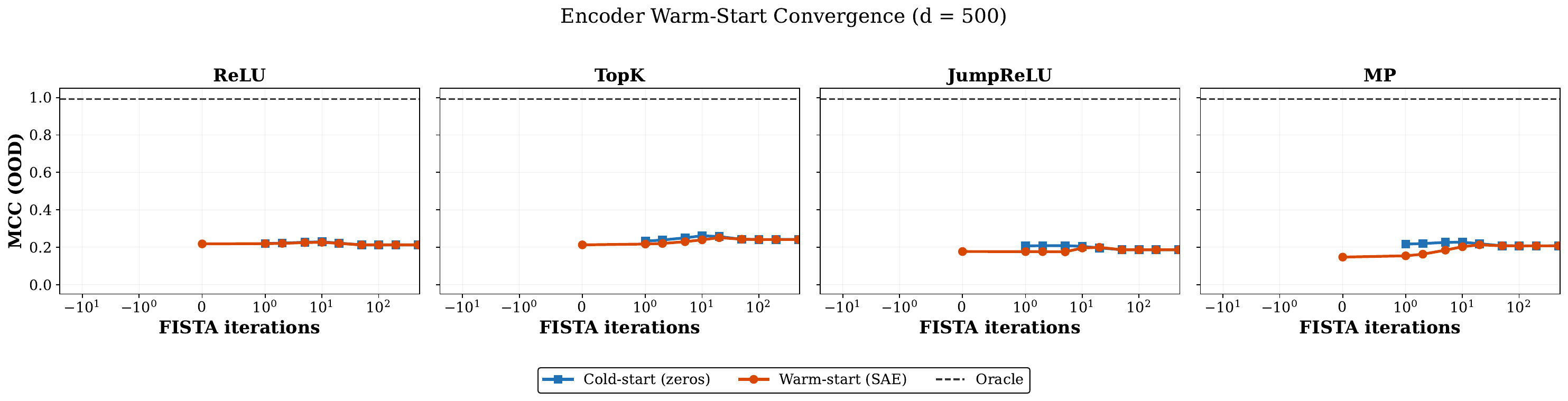}
    \caption{Encoder warm-start convergence, $d_z=500$.}
\end{figure}
\begin{figure}[h]
    \centering
    \includegraphics[width=\linewidth]{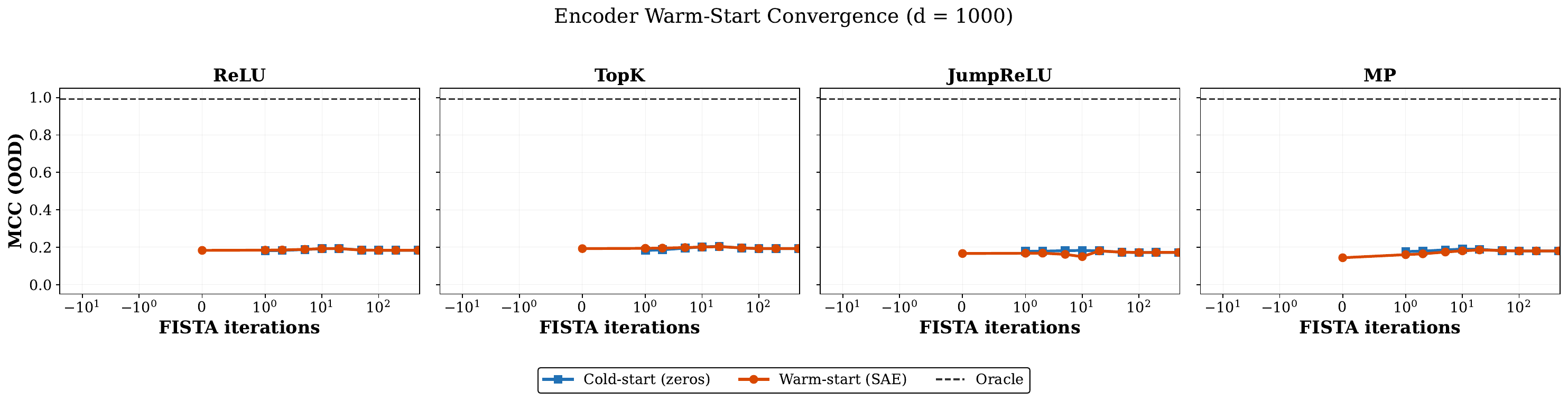}
    \caption{Encoder warm-start convergence, $d_z=1000$.}
\end{figure}
\begin{figure}[h]
    \centering
    \includegraphics[width=\linewidth]{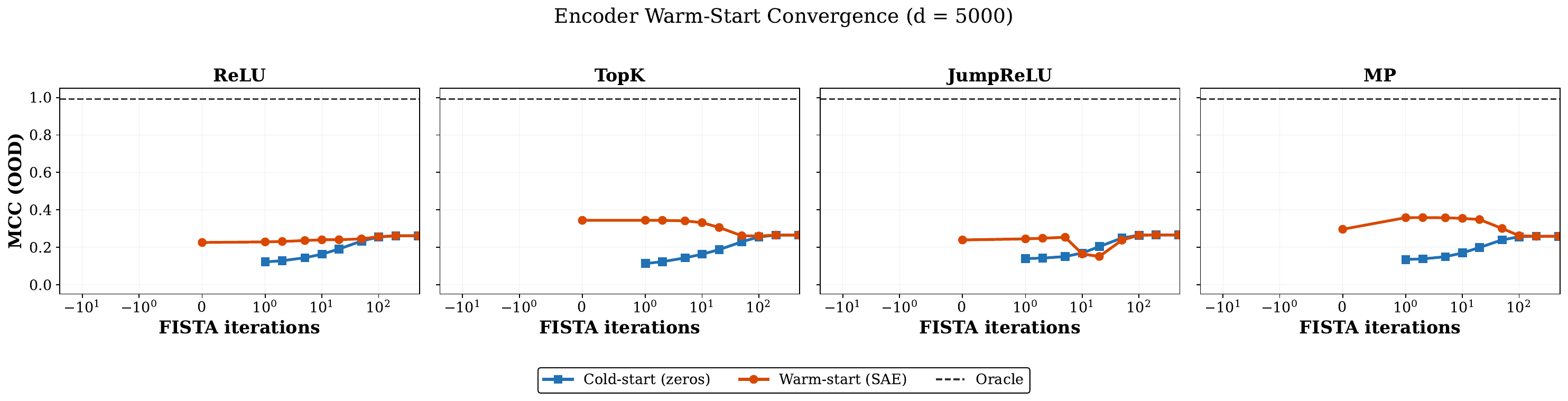}
    \caption{Encoder warm-start convergence, $d_z=5000$.}
\end{figure}
\begin{figure}[h]
    \centering
    \includegraphics[width=\linewidth]{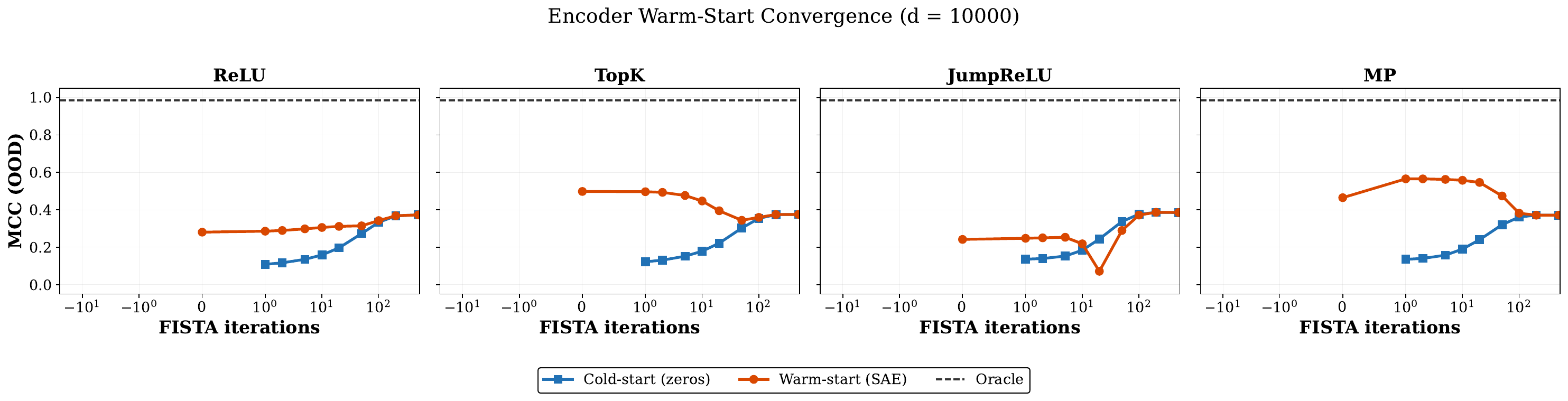}
    \caption{Encoder warm-start convergence, $d_z=10000$.}
\end{figure}

\parless{Dictionary quality and support recovery.}
The dictionary quality decomposition and support recovery patterns are consistent across $d_z$: re-normalising columns never helps, and ReLU/JumpReLU support precision remains catastrophically low.

\begin{figure}[h]
    \centering
    \includegraphics[width=\linewidth]{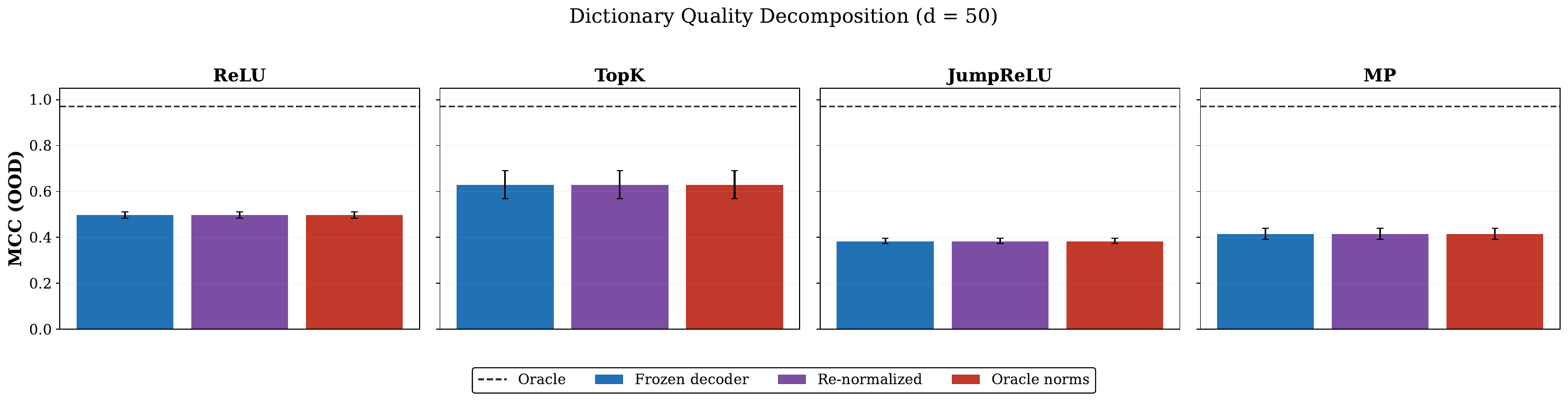}
    \caption{Dictionary quality decomposition, $d_z=50$.}
\end{figure}
\begin{figure}[h]
    \centering
    \includegraphics[width=\linewidth]{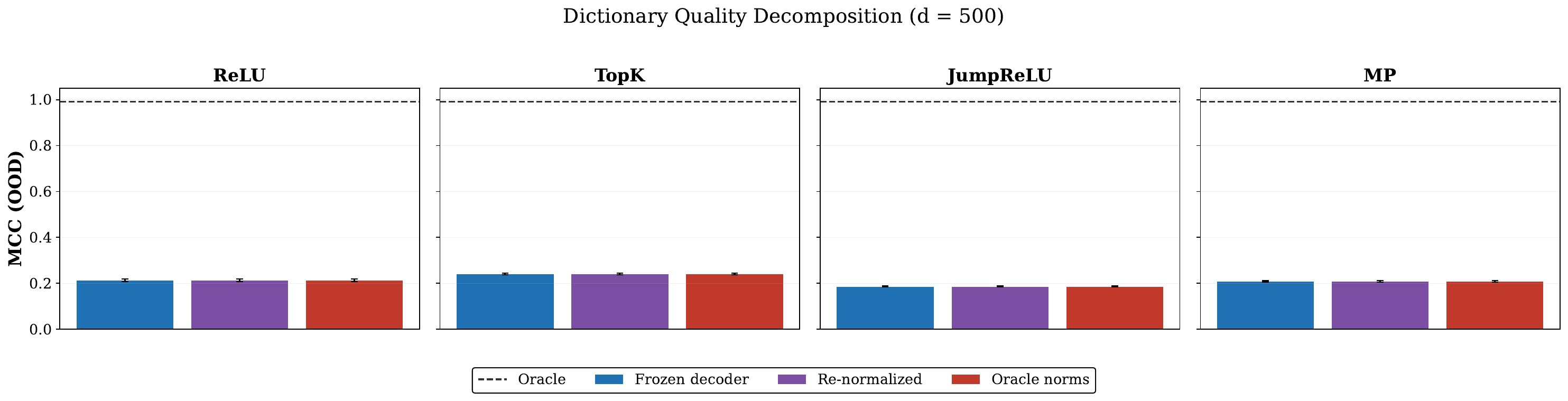}
    \caption{Dictionary quality decomposition, $d_z=500$.}
\end{figure}
\begin{figure}[h]
    \centering
    \includegraphics[width=\linewidth]{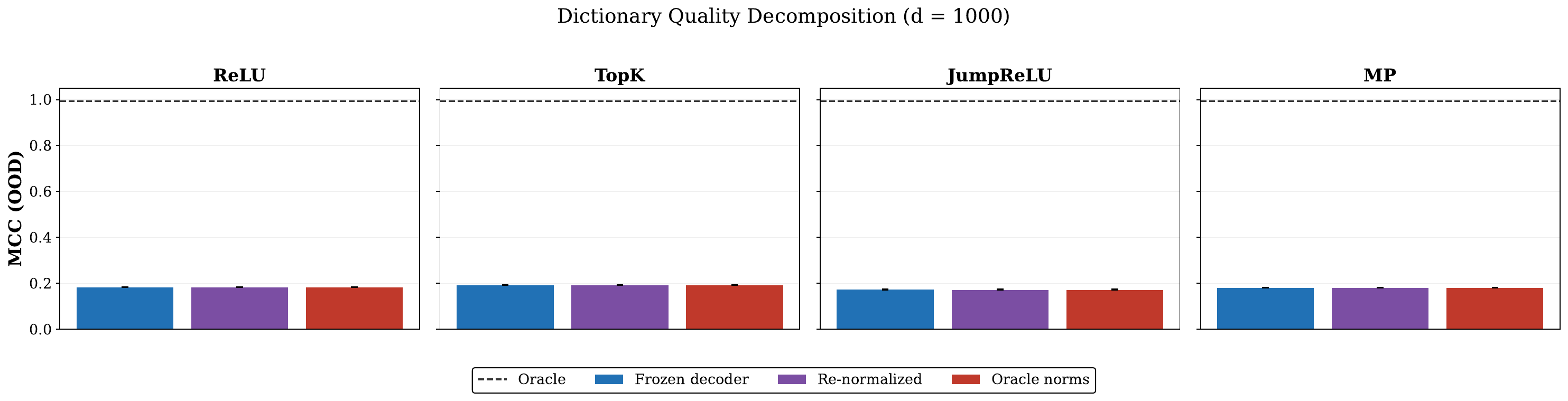}
    \caption{Dictionary quality decomposition, $d_z=1000$.}
\end{figure}
\begin{figure}[h]
    \centering
    \includegraphics[width=\linewidth]{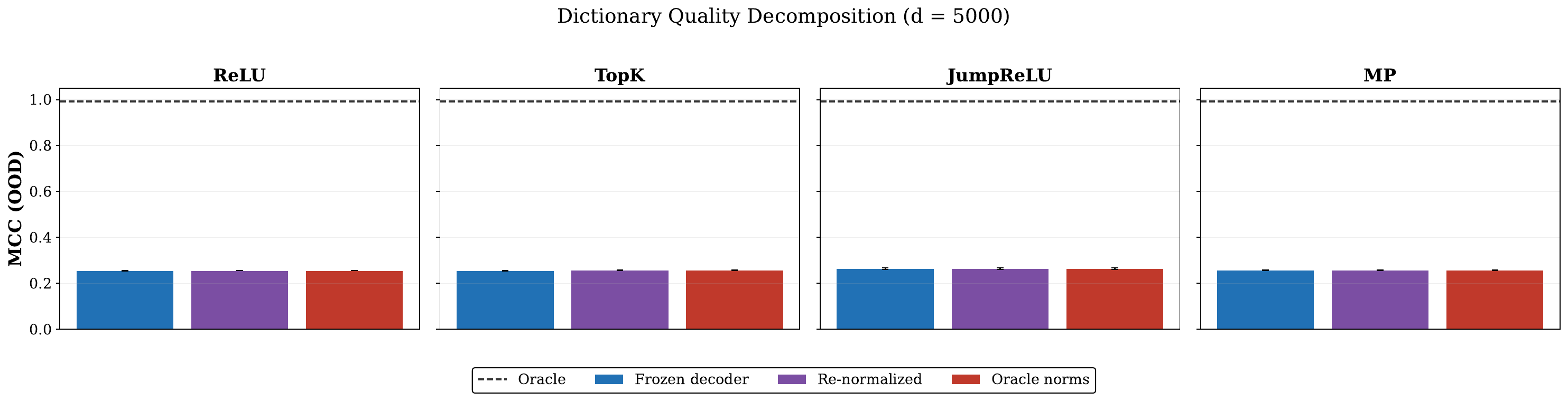}
    \caption{Dictionary quality decomposition, $d_z=5000$.}
\end{figure}
\begin{figure}[h]
    \centering
    \includegraphics[width=\linewidth]{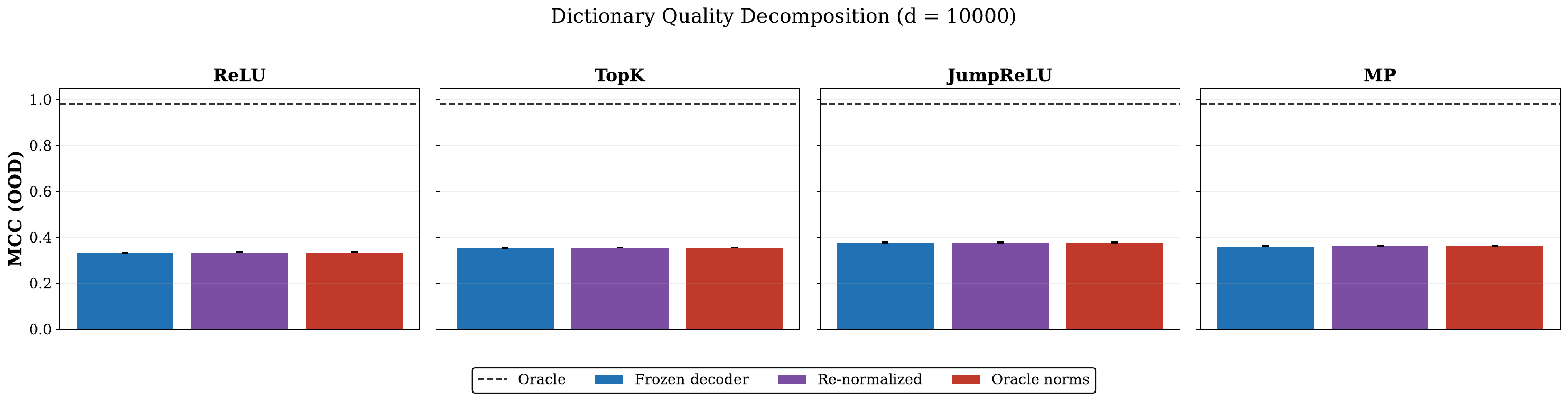}
    \caption{Dictionary quality decomposition, $d_z=10000$.}
\end{figure}

\begin{figure}[h]
    \centering
    \includegraphics[width=\linewidth]{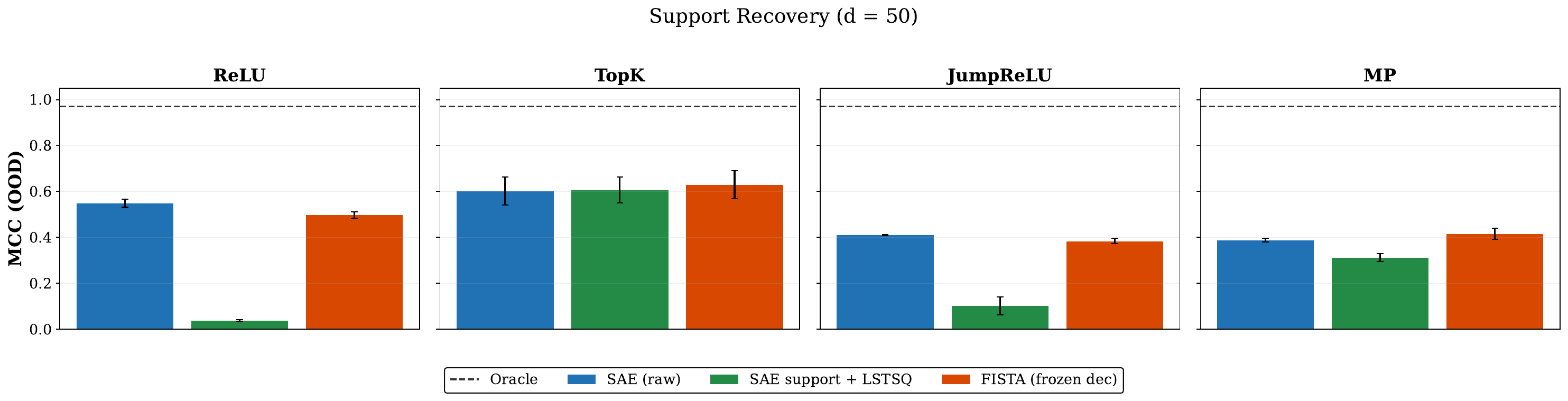}
    \caption{Support recovery, $d_z=50$.}
\end{figure}
\begin{figure}[h]
    \centering
    \includegraphics[width=\linewidth]{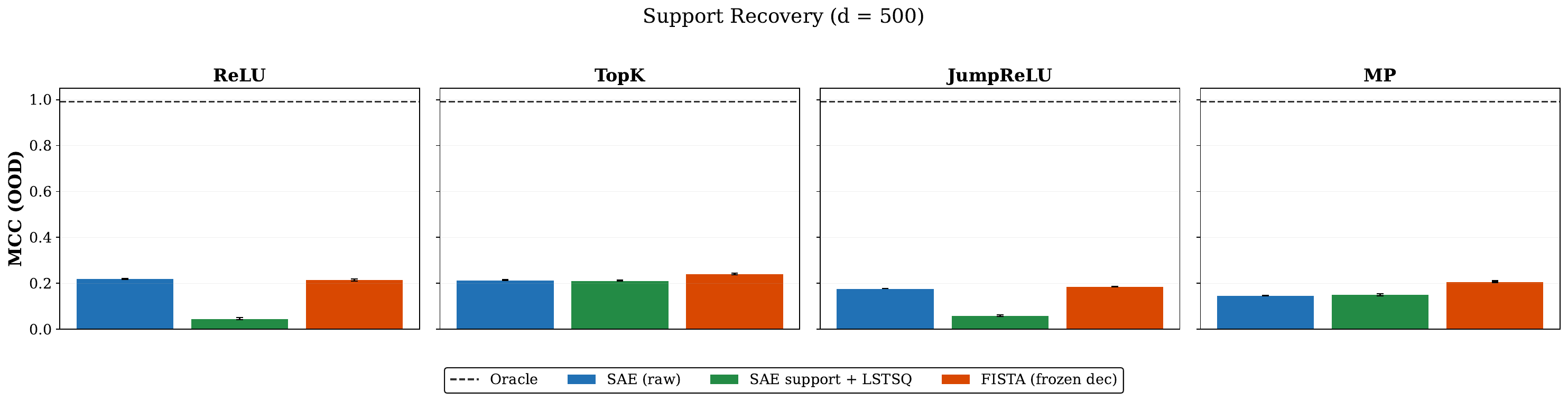}
    \caption{Support recovery, $d_z=500$.}
\end{figure}
\begin{figure}[h]
    \centering
    \includegraphics[width=\linewidth]{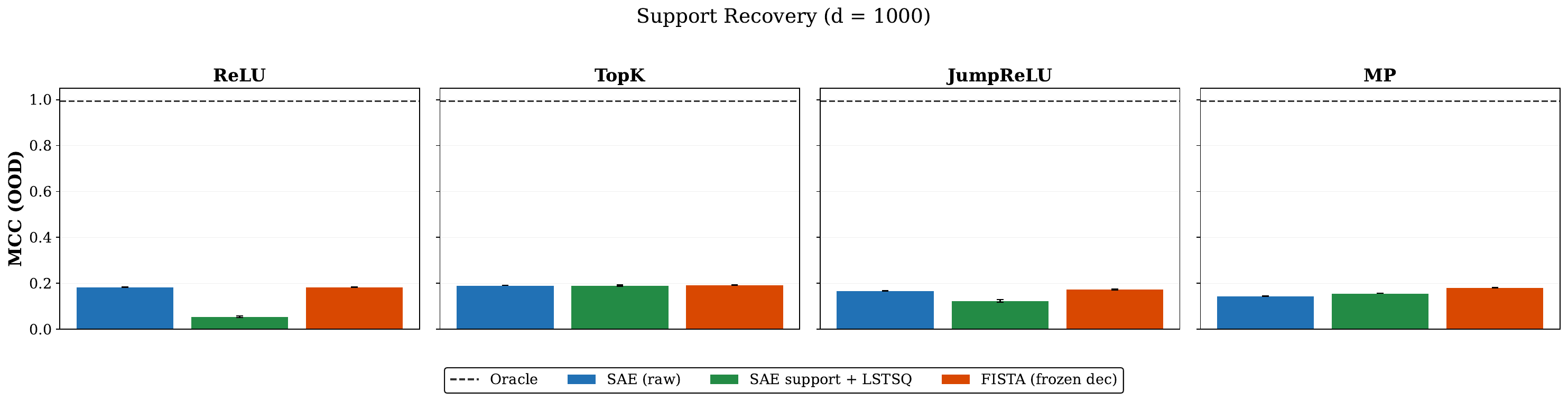}
    \caption{Support recovery, $d_z=1000$.}
\end{figure}
\begin{figure}[h]
    \centering
    \includegraphics[width=\linewidth]{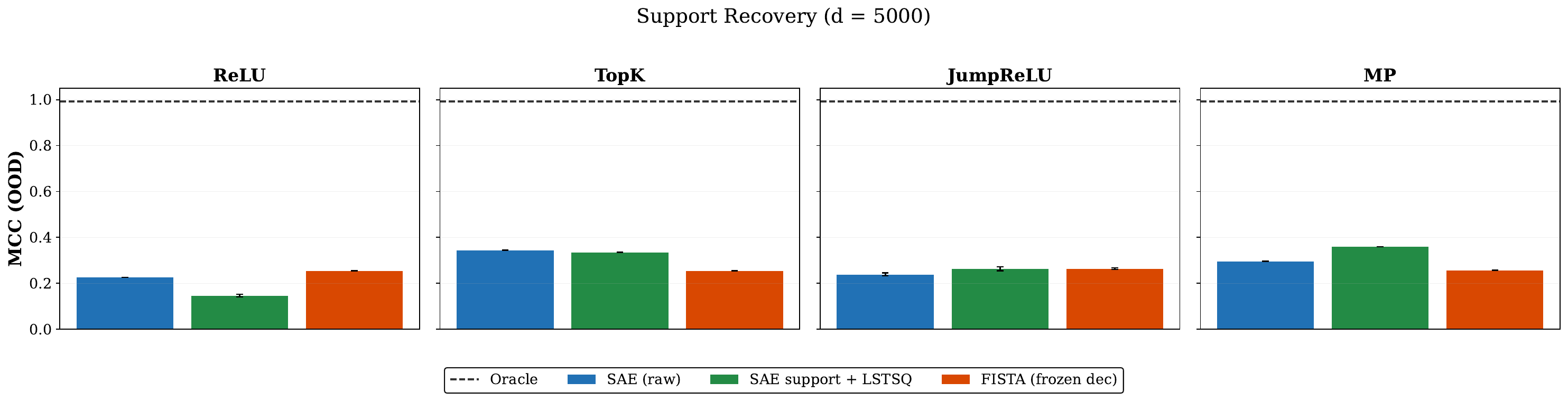}
    \caption{\textbf{Re-estimating magnitudes on the SAE's incorrect support degrades MCC.} Blue: raw SAE codes. Green: SAE support with least-squares magnitude re-estimation. Orange: FISTA on frozen decoder. Dashed: oracle. Only TopK's support is useful. $d_z=5000$, $k=10$.}
    \label{fig:support-recovery}
\end{figure}
\begin{figure}[h]
    \centering
    \includegraphics[width=\linewidth]{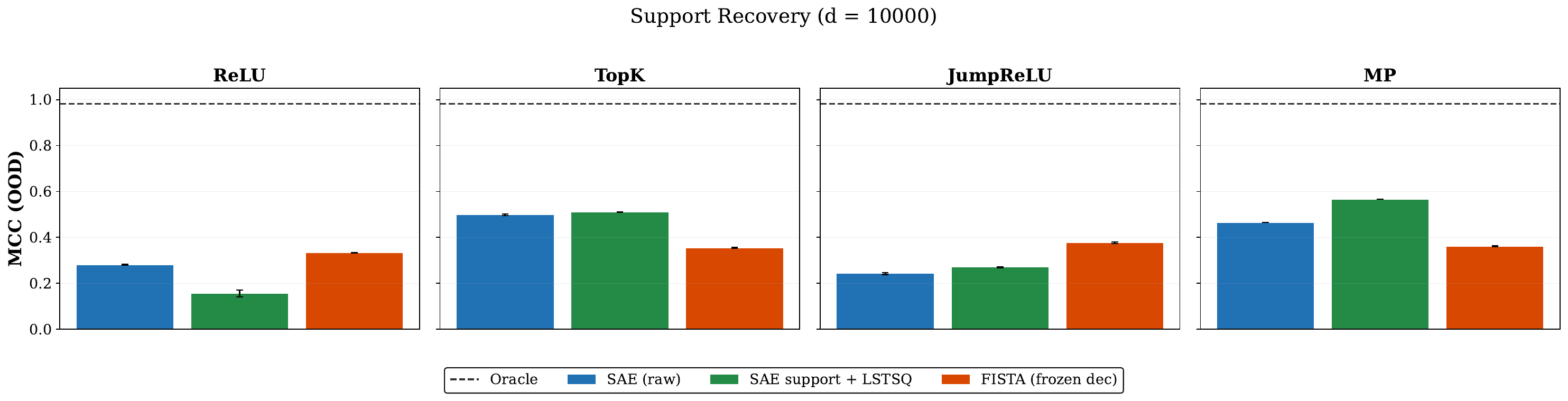}
    \caption{Support recovery, $d_z=10000$.}
\end{figure}

\subsection{Phase transition ablations}
\begin{figure}
    \centering
    \includegraphics[width=\linewidth]{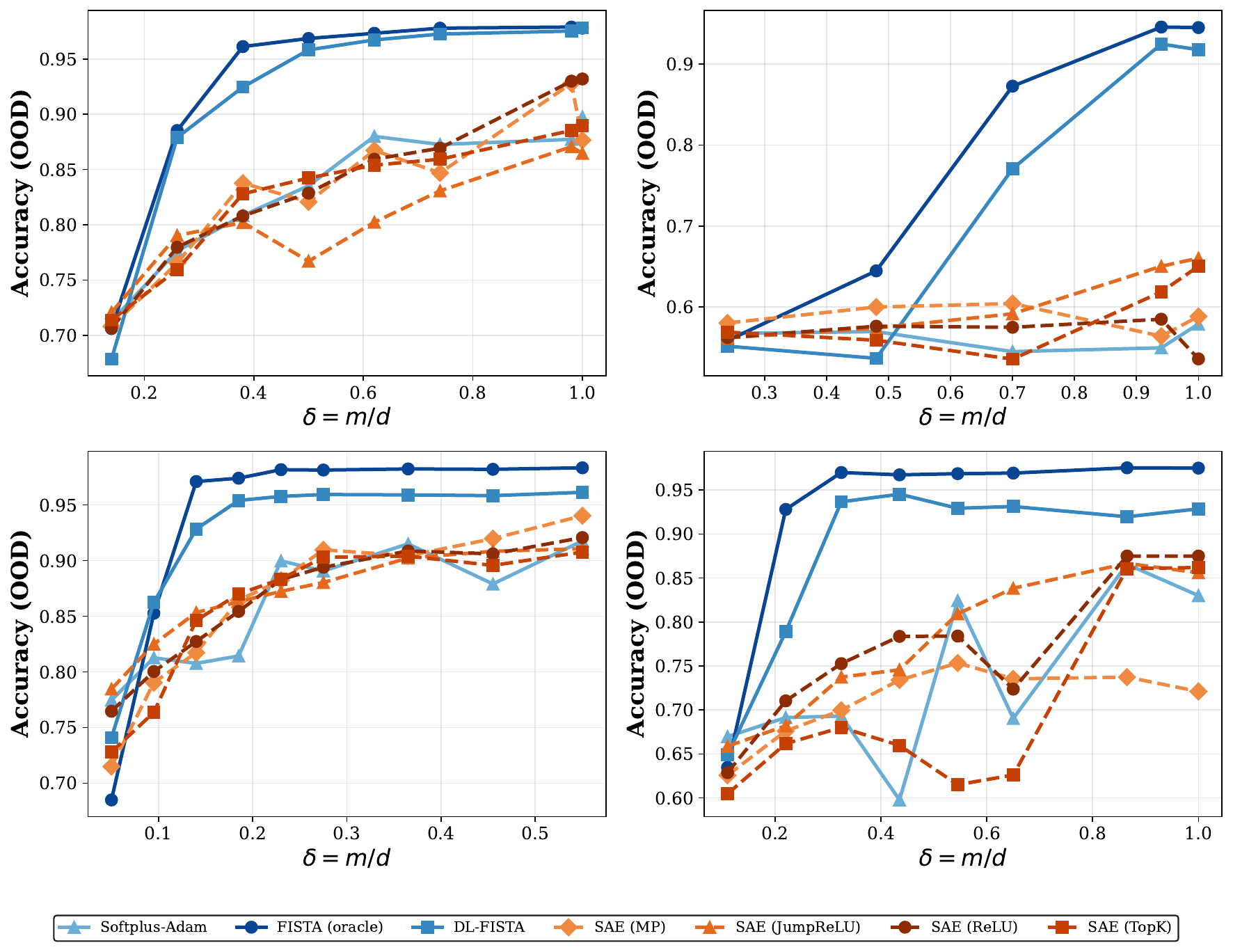}
    \caption{\textbf{Phase transition: Accuracy (OOD).} SAE OOD accuracy plateaus while per-sample methods transition sharply. The gap is most severe at moderate $\delta$ where compressed sensing succeeds but amortised inference fails.}
    \label{fig:phase_faceted_acc_ood}
\end{figure}

\begin{figure}
    \centering
    \includegraphics[width=\linewidth]{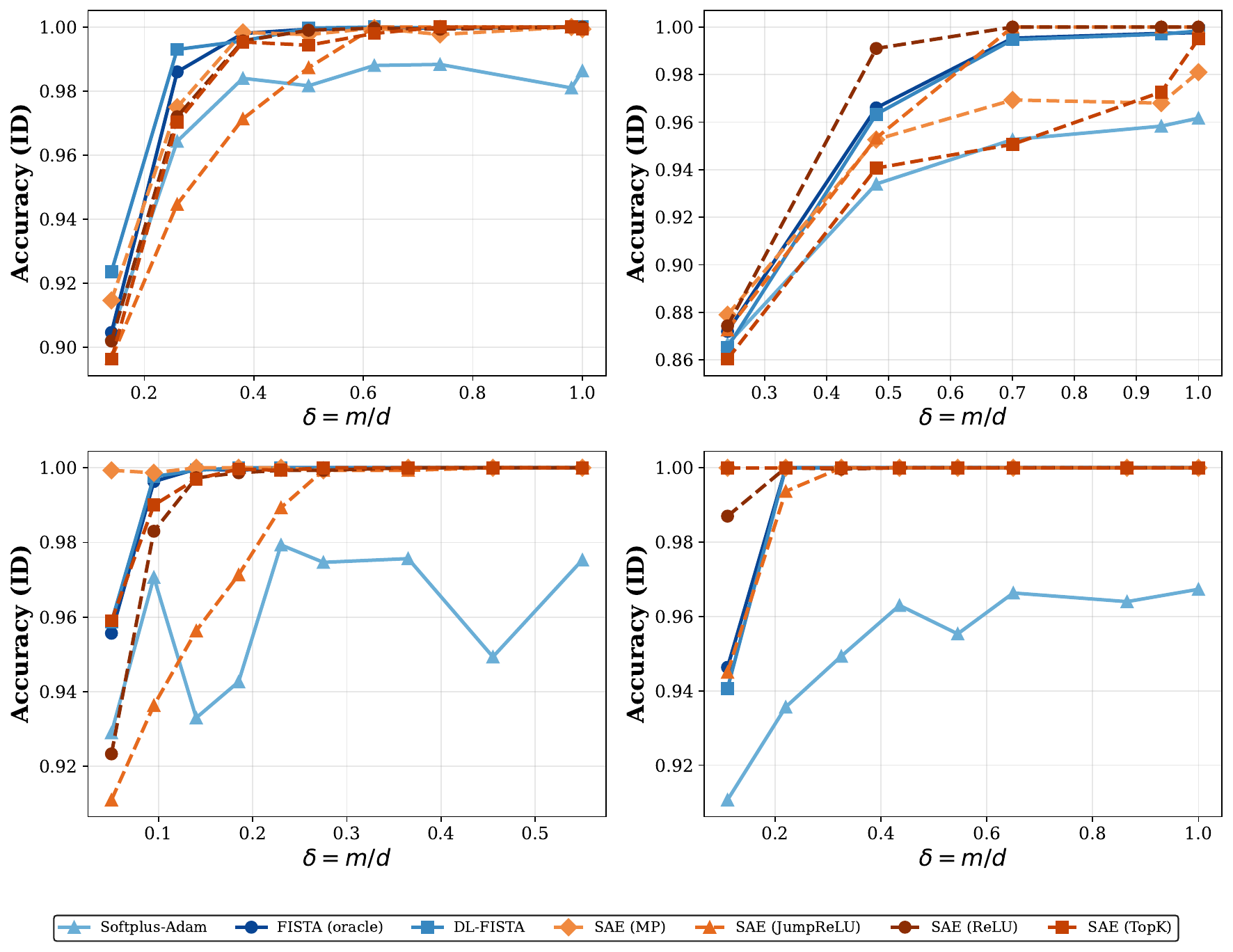}
    \caption{\textbf{Phase transition: Accuracy (ID).} ID accuracy is high for most methods once $\delta$ is sufficient, but SAEs show more variance and lower peak accuracy than per-sample methods.}
    \label{fig:phase_faceted_acc_id}
\end{figure}

\begin{figure}
    \centering
    \includegraphics[width=\linewidth]{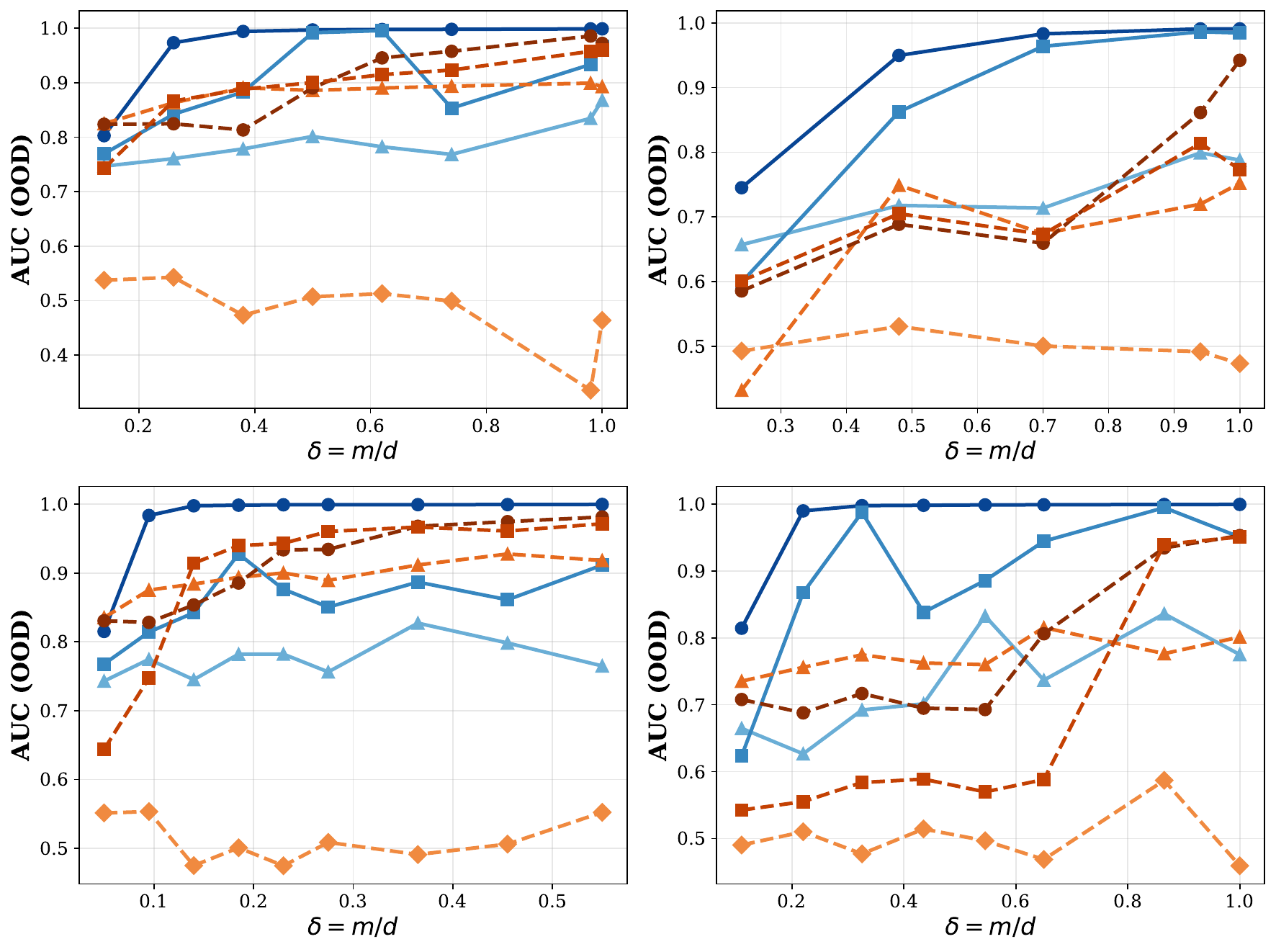}
    \caption{\textbf{Phase transition: AUC (OOD).} Per-feature AUC on OOD data confirms the same pattern: per-sample methods achieve near-perfect AUC while SAE features fail to isolate the label under novel compositions.}
    \label{fig:phase_faceted_auc_ood}
\end{figure}

\begin{figure}
    \centering
    \includegraphics[width=\linewidth]{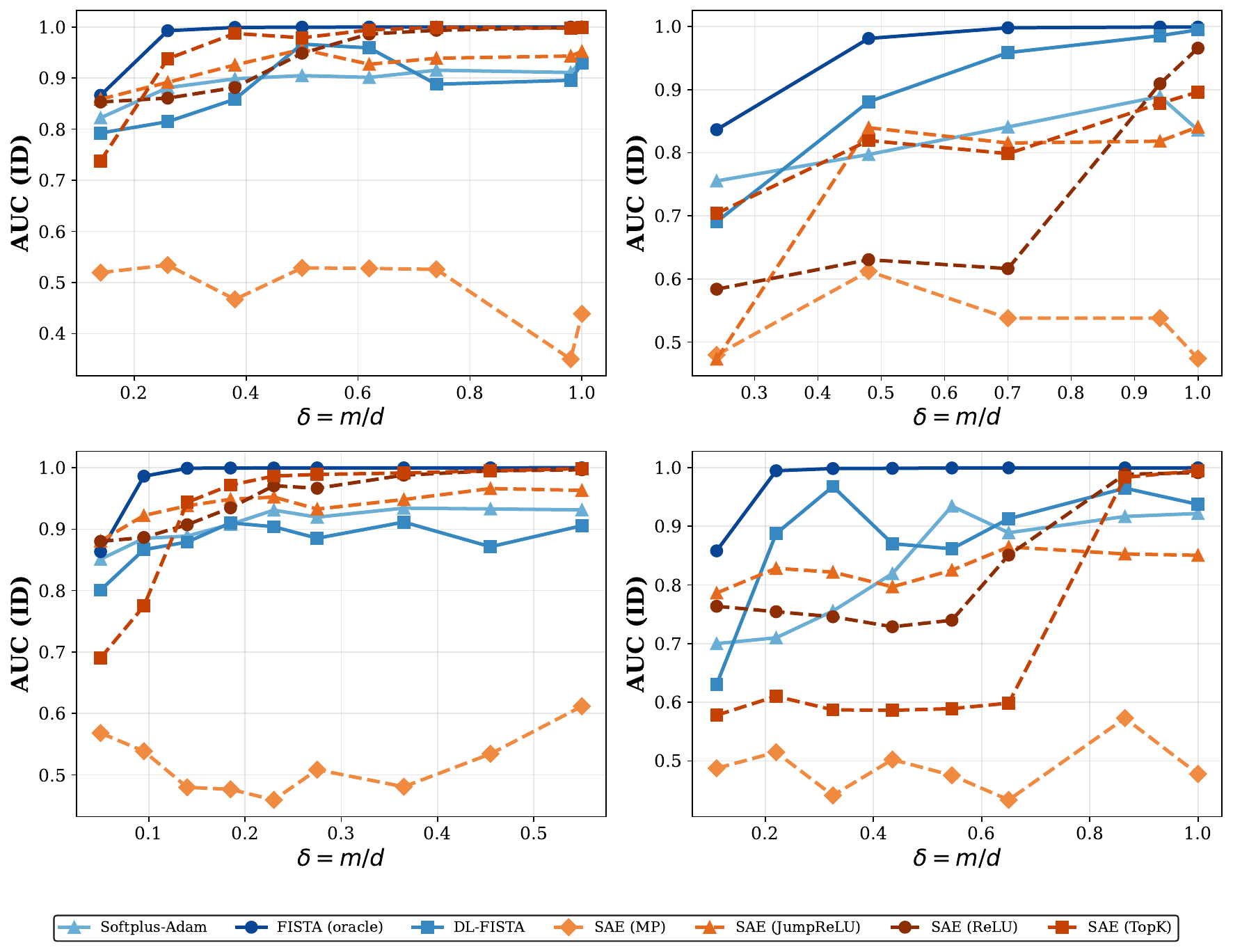}
    \caption{\textbf{Phase transition: AUC (ID).} ID AUC is high for most methods, but the gap between per-sample and amortised methods remains visible even in-distribution.}
    \label{fig:phase_faceted_auc_id}
\end{figure}

\begin{figure}
    \centering
    \includegraphics[width=\linewidth]{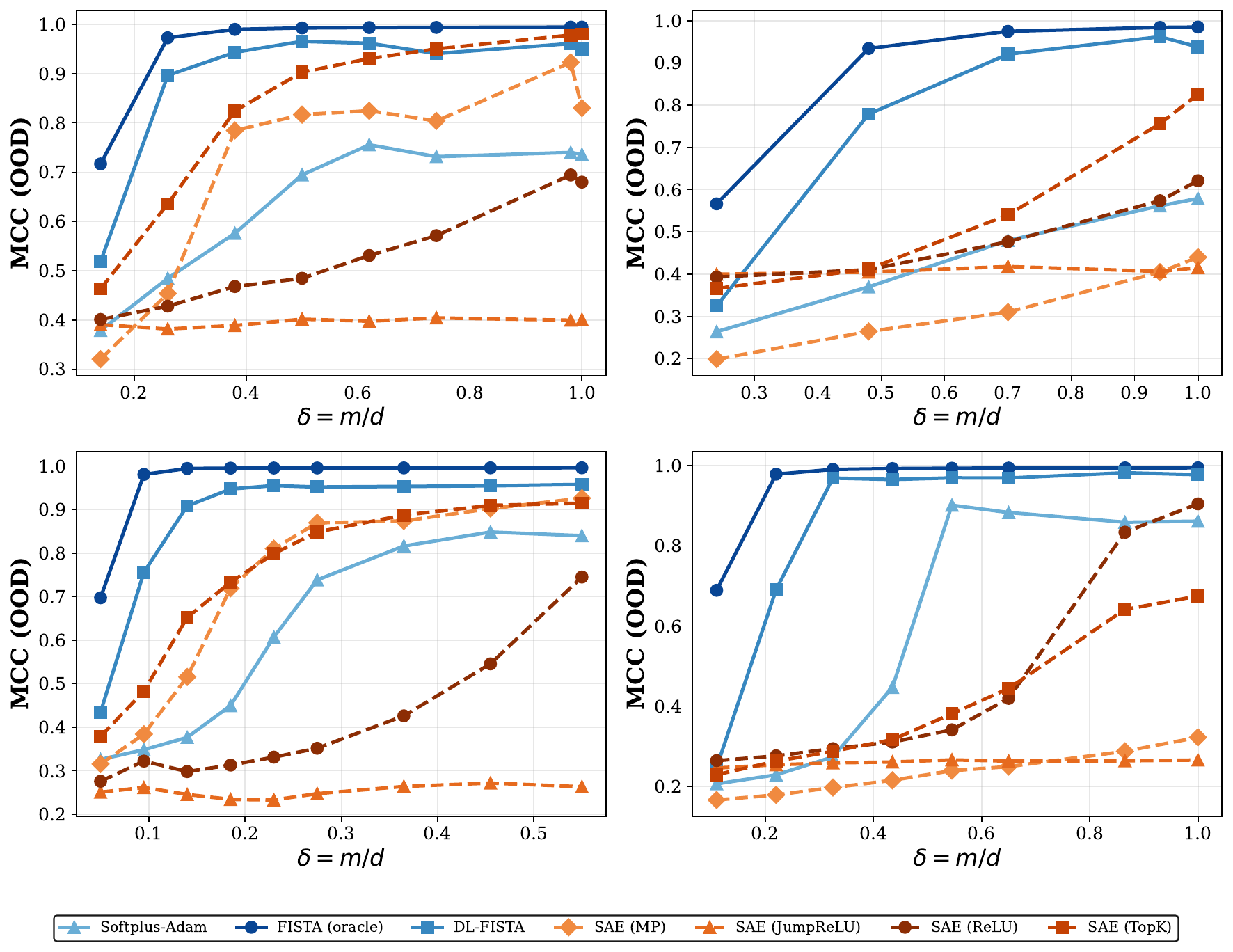}
    \caption{\textbf{Phase transition: MCC (OOD).} OOD MCC shows the clearest separation: per-sample methods recover the full latent structure under novel compositions while SAEs fail to identify latents OOD.}
    \label{fig:phase_faceted_mcc_ood}
\end{figure}

\cref{fig:phase_faceted_acc_ood,fig:phase_faceted_acc_id,fig:phase_faceted_auc_ood,fig:phase_faceted_auc_id,fig:phase_faceted_mcc_ood} extend the main-text phase transition (\cref{fig:phase}) with all remaining metrics. The phase transition pattern is consistent: per-sample methods exhibit a sharp transition to near-perfect performance once $\delta$ exceeds the compressed-sensing threshold, while SAEs plateau well below across all metrics.

\subsection{Studying the amortization gap}

\begin{figure}
    \centering
    \includegraphics[width=0.5\linewidth]{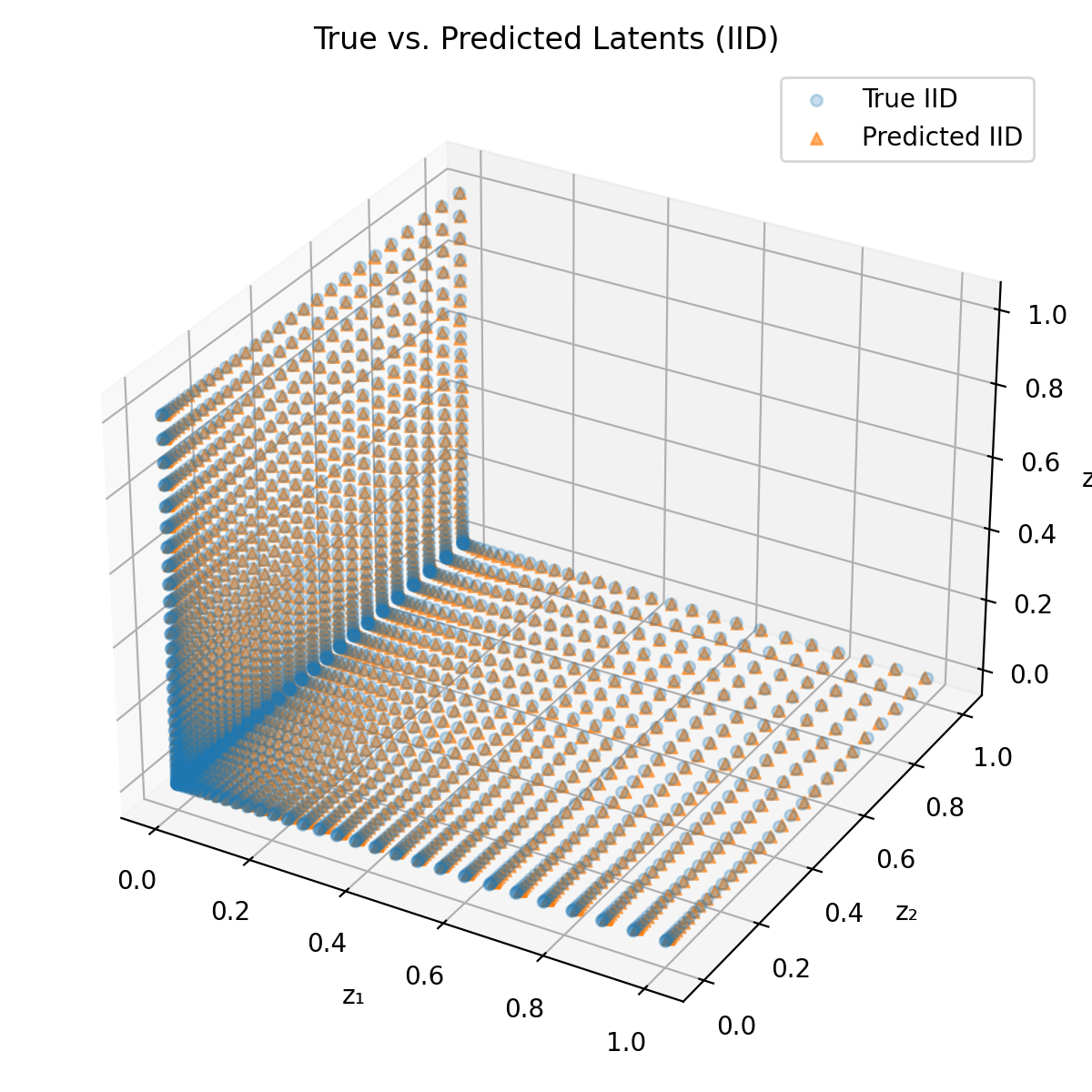}
    \caption{Latent reconstruction ID for SAE with increased capacity.}
    \label{fig:mlp_sae_iid}
\end{figure}

\begin{figure}
    \centering
    \includegraphics[width=0.5\linewidth]{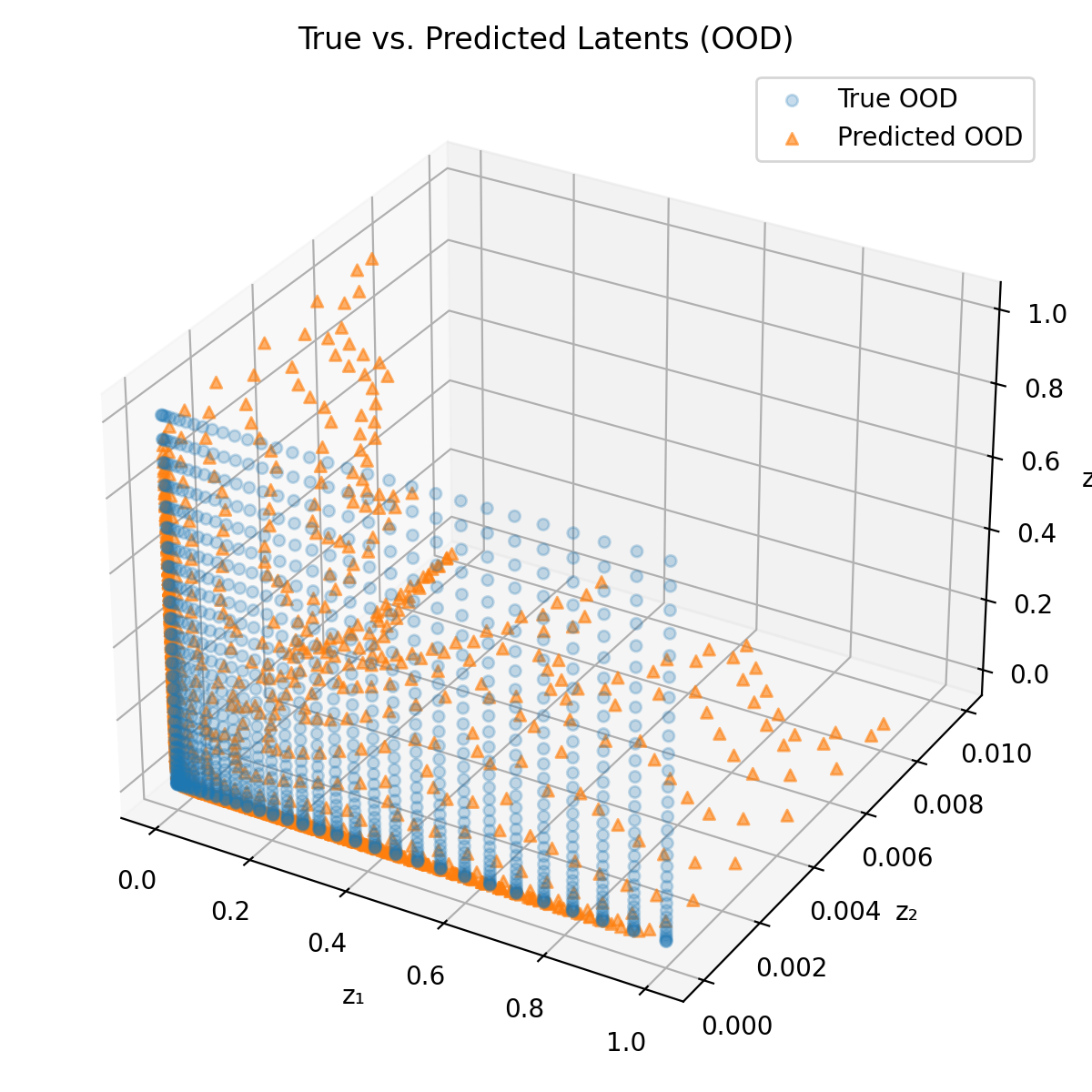}
    \caption{Latent reconstruction OOD for SAE with increased capacity.}
    \label{fig:mlp_sae_ood}
\end{figure}

To test the effect of the amortization of the inference model, we increase the capacity of standard SAEs in the toy setting described in Figure \ref{fig:toy3d}. We observe that when increasing the capacity of the encoder (and reducing the amortization error), the reconstruction of the latents in-distribution is nearly perfect (Figure \ref{fig:mlp_sae_iid}). In contrast, the OOD setting (Fig. \ref{fig:mlp_sae_ood}) is completely wrong. The true plane with a new combination of latent variables is completely mismatched: the SAE maps all the data points to the support of the variables seen in the training data. This visualization confirms that reducing the amortization error is not sufficient to improve OOD generalization when a particular combination of latent variables is not observed in the training set.

\paragraph{Experimental details:} Capacity is increased by adding an extra layer to the encoder, with dimension 256, followed by nonlinearity.

\section{Theoretical model for toy setting}

\begin{figure}
    \centering
    \includegraphics[width=0.7\linewidth]{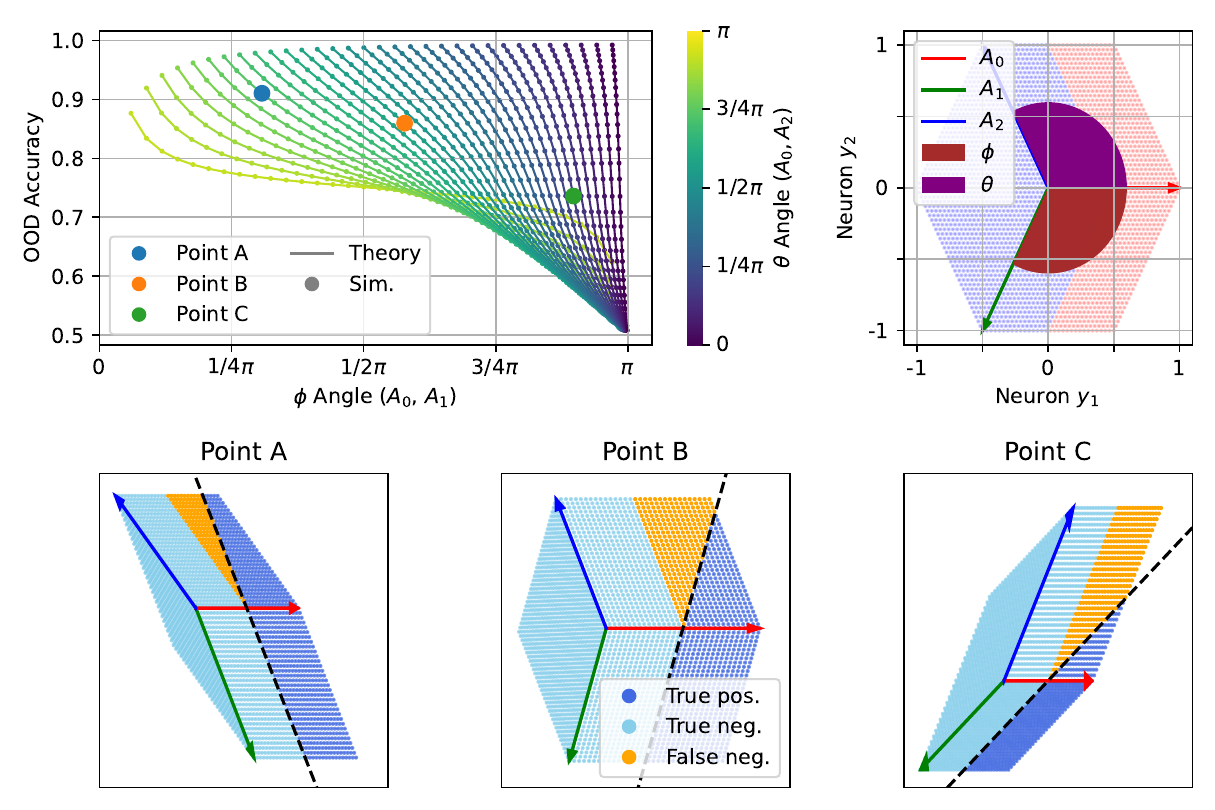}
    \caption{
    \textbf{Theory.}
    \textbf{Top left,} shows the theoretically predicted accuracy and simulations of a perfect linear classifier, trained and tested (OOD) on distinct latent combinations (see \cref{fig:overview_boundary}).
    \textbf{Top right,} illustrates the geometry of the classification problem (red and blue classes) with the directions of the decoder \(A\) columns for each latent and the angles (\(\phi, \theta\)) between them.
    \textbf{Bottom,} shows the resulting geometry for three sample points from the first plot.
    }
    \label{fig:theory}
\end{figure}

We study the geometry of a system where a sparse source vector $z \in [0, 1]^3$ with at most two non-zero elements ($\lVert z\rVert_0 \leq 2$) is linearly projected to an observation $y \in \mathbb{R}^2$ (see Fig.~\ref{fig:overview_boundary}):
\begin{equation}
    y = Az.
\end{equation}
The sparsity constraint implies that any observation is a combination of at most two active source components.
Whenever active, we assume that each source follows a uniform distribution $z_i \,|\, i \text{ active} \sim \text{Uniform}(0, 1)$.
The training data is considered \textit{independent and identically distributed} (IID) and is generated from combinations of sources $(z_1, z_2)$ or $(z_2, z_3)$. The test data is considered \textit{out-of-distribution} (OOD) and is generated from the novel combination $(z_1, z_3)$. Our goal is to determine whether the first variable $z_1$ is above a certain, safety-relevant, threshold $z_1=\frac{1}{2}$.

To analyze the geometry, we examine the columns $A_i \in \mathbb{R}^2$ of the projection matrix. We define the angles $\phi := \angle(A_1, A_2)$ and $\theta := \angle(A_1, A_3)$, which fully determine the system. To simplify the analysis, we make two assumptions:
\begin{enumerate}
    \item We align our coordinate system and fix the magnitude of the first basis vector relative to our threshold, such that $A_1=(2, 0)$ and $\lVert A_2 \rVert = \lVert A_3 \rVert = 1$.
    \item To ensure the cones spanned by the vectors do not overlap, we require that $0 < \phi, \theta < \pi$ and $\phi + \theta > \pi$. This is an illustrative way of understanding why and when compressed sensing is possible in this system.
\end{enumerate}

A perfect linear classifier trained on the IID data must separate the space based on the condition $z_1 = \frac{1}{2}$. In the observation space, this corresponds to a line parallel to $A_2$ and passing through the point $\frac{1}{2} A_1$. This decision boundary is the line parameterized by:
\begin{equation}
    y(\beta) = \frac{1}{2} A_1 + \beta A_2, \quad \beta \in \mathbb{R}.
\end{equation}
The question we are interested in is: \textit{what is the accuracy of this classifier on the OOD data?}
We derive the analytically predicted OOD accuracy for this perfect linear IID classifier, separating two cases, in Appendix \ref{sec:app_theory}.

The simulations and analytical prediction are tested and illustrated in Fig.~\ref{fig:theory} confirming the validity of the theory.

\subsection{Derivation}\label{sec:app_theory}

We study the geometry of a system where a sparse source vector $z \in [0, 1]^3$ with at most two non-zero elements ($\lVert z\rVert_0 \leq 2$) is linearly projected to an observation $y \in \mathbb{R}^2$ (see Fig.~\ref{fig:overview_boundary}):
\begin{equation}
    y = Az.
\end{equation}
The sparsity constraint implies that any observation is a combination of at most two active source components.
Whenever active, we assume that each source follows a uniform distribution $z_i \,|\, i \text{ active} \sim \text{Uniform}(0, 1)$.
The training data is considered \textit{independent and identically distributed} (IID) and is generated from combinations of sources $(z_1, z_2)$ or $(z_2, z_3)$. The test data is considered \textit{out-of-distribution} (OOD) and is generated from the novel combination $(z_1, z_3)$. Our goal is to determine whether the first variable $z_1$ is above a certain, safety-relevant, threshold $z_1=\frac{1}{2}$.

To analyze the geometry, we examine the columns $A_i \in \mathbb{R}^2$ of the projection matrix. We define the angles $\phi := \angle(A_1, A_2)$ and $\theta := \angle(A_1, A_3)$, which fully determine the system. To simplify the analysis, we make two assumptions:
\begin{enumerate}
    \item We align our coordinate system and fix the magnitude of the first basis vector relative to our threshold, such that $A_1=(2, 0)$ and $\lVert A_2 \rVert = \lVert A_3 \rVert = 1$.
    \item To ensure the cones spanned by the vectors do not overlap, we require that $0 < \phi, \theta < \pi$ and $\phi + \theta > \pi$. This is an illustrative way of understanding why and when compressed sensing is possible in this system.
\end{enumerate}

A perfect linear classifier trained on the IID data must separate the space based on the condition $z_1 = \frac{1}{2}$. In the observation space, this corresponds to a line parallel to $A_2$ and passing through the point $\frac{1}{2} A_1$. This decision boundary is the line parameterized by:
\begin{equation}
    y(\beta) = \frac{1}{2} A_1 + \beta A_2, \quad \beta \in \mathbb{R}.
\end{equation}
The question we are interested in is: \textit{what is the accuracy of this classifier on the OOD data?}
Clearly, the perfect linear classifier for the OOD data would have a decision boundary that is parallel to $A_3$ and shifted by $\frac{1}{2} A_1$, i.e., the line:
\begin{equation}
    y_{\text{OOD}}(\beta) = \frac{1}{2} A_1 + \beta A_3, \quad \beta \in \mathbb{R}.
\end{equation}
Since $\phi + \theta > \pi$, we know that the IID classifier's boundary cannot be aligned with the ideal OOD classifier's boundary, so there must be some OOD error. Moreover, we know that the IID classifier can never `under-shoot' on the OOD data (that would require $\phi + \theta < \pi$). Consequently, we will only observe \textit{false negatives}---that is, test points with $z_1 > \frac{1}{2}$ that are erroneously classified as safe.

We now have to distinguish:
\textbf{Case 1}, where the classifier passes right from the top right corner $(A_1 + A_3)$ (Fig.~\ref{fig:theory} Point C),
and \textbf{Case 2}, where the classifier passes left from the top right corner $(A_1 + A_3)$ (Fig.~\ref{fig:theory} Point A).

The separation happens when the classifier passes through the top right corner.
In that case it will form a triangle through the points $(\frac{1}{2}A_1, A_1, A_1 + A_3)$, with associated angles $(a, b, c) := (\pi - \phi, \pi - \theta, \phi + \theta - \pi)$.
By assumption, the base of this triangle has length $1$.
Consequently, trigonometry tells us that the first angle must have a fixed relation to the second angle $a = \frac{\pi - b}{2}$.
From this it follows that the condition for Case 1 is
\begin{equation}
    \frac{\pi - (\pi - \theta)}{2} < \pi - \phi \quad \Rightarrow \quad \phi + \frac{\theta}{2} < \pi.
\end{equation}

The total area of the right parallelogram $(\frac{1}{2}A_1, A_1, A_1 + A_3, \frac{1}{2}A_1 + A_3)$ is $\alpha = \sin(\theta)$.
To compute the area of a triangle within this diagram, we use the fact that the area of a triangle can be computed from one side and the adjacent angles.
We will always pick a side with length \(1\), so that if the adjacent angles are \((a, b)\), the area equals
\begin{equation}\label{eq:triangle_area}
    \alpha(a,b) = \frac{\sin(a) \sin(b)}{2 \sin(a + b)}.
\end{equation}

In Case 1, we compute the area ($\alpha_1$) of the triangle between the classifier and $A_1$.
The base between \(\frac{1}{2}A_1\) and \(A_1\) has length \(1\).
The angle on the left is \(a_1=\pi - \phi\) and the angle on the right is \(b_1=\pi - \theta\).
Thus, using \eqref{eq:triangle_area}, the area is
\begin{equation}
     \alpha_1 = \frac{\sin(\pi - \phi) \sin(\pi - \theta)}{2 \sin(\phi + \theta - \pi)} = \frac{\sin(\phi) \sin(\theta)}{2 \sin(\phi + \theta - \pi)}
\end{equation}
The OOD accuracy will be 50\% for the true negatives, plus 50\% times the proportion that the area occupies in the right parallelogram \((\alpha)\)
\begin{equation}
    acc_{1}(\text{OOD}) = \frac{1}{2} + \frac{\alpha_1}{2\alpha}.
\end{equation}

In Case 2, we compute the area ($\alpha_2$) of the triangle between the classifier and the correct OOD decision boundary.
The base between \(\frac{1}{2}A_1\) and \(\frac{1}{2}A_1 + A_3\) has length \(1\).
The angle on top is \(a_2=\pi - \theta\) and the angle below is \(b_2=\phi + \theta - \pi\).
Thus, using \eqref{eq:triangle_area}, the area is
\begin{equation}
     \alpha_2 = \frac{\sin(\pi - \theta) \sin(\phi + \theta - \pi)}{2 \sin(\phi)} = \frac{\sin(\theta) \sin(\phi + \theta - \pi)}{2 \sin(\phi)}
\end{equation}
The OOD accuracy will be 100\% minus the proportion that the area occupies in the left plus right parallelogram \((2\alpha)\)
\begin{equation}
    acc_{2}(\text{OOD}) = 1 - \frac{\alpha_2}{2\alpha}.
\end{equation}

\end{document}